\documentclass[11pt]{article}

\usepackage{amsmath,amssymb,amsfonts}
\usepackage{graphicx}
\usepackage{subfigure}
\usepackage{latexsym}
\usepackage{graphicx}
\usepackage{times}
\usepackage{fullpage}

\usepackage{epsfig}
\usepackage{color}

\usepackage{clrscode}

\newcommand{\mr}[1]{}
\newcommand{\mrx}[1]{}
\newcommand{\mb}[1]{}
\newcommand{\mbx}[1]{#1}
\newcommand{\mw}[1]{}
\newcommand{\xmr}[1]{}
\newcommand{\xmb}[1]{}

\newcommand{\cc}{2\eps}
\def\T{\mathcal{S}}
\def\t{t}
\newcommand{\CostP}{\mathbf{cost}}
\newcommand{\costP}{\mathbf{cost}}
\newcommand{\Pbad}{P_{\text{bad}}}
\newcommand{\s}{\overline{s}}
\newcommand{\f}{\overline{f}}
\newcommand{\HH}{\mathbf{H}}

\newcommand{\M}{Y}
\newcommand{\MM}{\mathcal{M}}
\newcommand{\B}{B}
\newcommand{\ind}{|}
\newcommand{\dimred}{\textsc{Metric-B-Coreset}}

\newcommand{\gm}{Y}
\newcommand{\coresetA}{\textsc{B-Coreset}}

\newcommand{\ZZ}{\mathbf{X}}
\newcommand{\DD}{D}

\newcommand{\rkmedian}{\textsc{$k$-Median-Coreset}}
\newcommand{\prr}{\mathrm{pr}}

\newcommand{\function}[2]{\DeclareMathOperator{#1}{#2}} % used for user functions to
\function{\flt}{flat}

\newcommand{\cent}{\mathbf{cent}}

\newcommand{\ftime}{\mathrm{t}}

\newcommand{\image}{[0,\infty)}

\newcommand{\sample}{{approximation}}

\newenvironment{proof}{\noindent {\bf Proof}.\ }{\qed \par\vskip 4mm\par}
\renewenvironment{proof}{\noindent {\bf Proof}.\ }{\qed \par\vskip 4mm\par}

\newtheorem{theorem}{Theorem}[section]
\newtheorem{corollary}[theorem]{Corollary}
\newtheorem{lemma}[theorem]{Lemma}

\newtheorem{definition}[theorem]{Definition}

\function{\columns}{columns}

\newcommand{\pr}{\mathbf{Pr}}                            %
\renewcommand{\labelenumi}{(\roman{enumi})}

\newcommand{\proj}{\mathrm{proj}}                                 %
\newcommand{\bicriteria}{\textsc{Bicriteria}}                                 % Algorithm Project-P-On-F
\newcommand{\median}{\textsc{Median}}                                 %

\newcommand{\satime}{\mathbf{SlowMedian}}

\newcommand{\ptastime}{\mathbf{SlowEpsApprox}}
\newcommand{\onemedtime}{\mathbf{SlowOneEpsMedian}}

\newcommand{\sq}{\hbox{\rlap{$\sqcap$}$\sqcup$}}
\newcommand{\qed}{\hspace*{\fill}\sq}
\newcommand{\COMMENTED}[1]{{}}                                                 % Commented text
\newcommand{\REAL}{\ensuremath{\mathbb{R}}}                                    % REAL
\newcommand{\eps}{\ensuremath{\varepsilon}}                                       % Epsilon
\newcommand{\e}{\ensuremath{\varepsilon}}
\newcommand{\dist}{\mathrm{dist}\hspace{-1pt}}                                 % Distance Function
\newcommand{\cost}{\mathrm{cost}}
\newcommand{\Cost}{\mathrm{Cost}}

\newcommand{\poly}{\mathrm{poly}}
\newcommand{\br}[1]{\left\{#1\right\}}                                         % {#}
\newcommand{\norm}[1]{\left\lVert#1\right\rVert}                               % Norm
\newcommand{\Span}[1]{\mathrm{span}\left({#1}\right)}                                  %Span of some point
\newcommand{\danny}[1]{}
\newcommand{\mikel}[1]{}

\newcommand{\range}{\mathbf{range}}
\newcommand{\ranges}{\mathbf{ranges}}
\newcommand{\RANGES}{\mathbf{Ranges}}
\newcommand{\XX}{\mathcal{X}}

\newcommand{\robusttime}{\mathbf{Median}}
\newcommand{\bicriteriatime}{\mathbf{Bicriteria}}
\newcommand{\zz}{z}
\begin{document}

%\conferenceinfo{STOC'11,} {June 6--8, 2011, San Jose, California, USA.}
%\CopyrightYear{2011}
%\crdata{978-1-4503-0691-1/11/06}
%\clubpenalty=10000
%\widowpenalty = 10000

\begin{titlepage}
\title{A Unified Framework for Approximating and Clustering Data
%\danny{I suggest: ``a unified framework for coresets" or ``for sub-linear time algorithms"}
}

\author{D. Feldman
\thanks{California Institute of Technology, Pasadena CA
91125. Email:  {\tt dannyf@caltech.edu}}
\and
M. Langberg
\thanks{Computer Science Division, Open University of Israel,
108 Ravutski St., Raanana 43107, Israel.
Email: {\tt mikel@openu.ac.il}.
Work supported in part by The Open University of
Israel's Research Fund (grant no. 46109), Cisco Collaborative
Research Initiative (CCRI), and ISF grant 480/08.}}
\date{}
\end{titlepage}

\maketitle

\begin{abstract}
Given a set $F$ of $n$ positive functions over a ground set $X$, we consider
the problem of computing $x^*$ that minimizes the expression $\sum_{f\in F}f(x)$, over $x\in X$. A typical application is \emph{shape fitting}, where we wish to approximate a set $P$ of $n$ elements (say, points) by a shape $x$ from a (possibly infinite) family $X$ of shapes. Here, each point $p\in P$ corresponds to a function $f$ such that $f(x)$ is the distance from $p$ to $x$, and we seek a shape $x$ that minimizes the sum of distances from each point in $P$. In the
$k$-clustering variant, each $x\in X$ is a tuple of $k$ shapes, and $f(x)$ is the distance from $p$ to its closest shape in $x$.

Our main result is a unified framework for constructing  {\em coresets} and {\em approximate clustering} for such general sets of functions.
To achieve our results, we forge a link between the classic and well defined notion of $\eps$-approximations from the theory of PAC Learning and VC dimension, to the relatively new (and not so consistent) paradigm of coresets, which are some kind of ``compressed representation" of the input set $F$.
Using traditional techniques, a coreset usually implies an LTAS (linear time approximation scheme) for the corresponding optimization problem, which can be computed in parallel, via one pass over the data, and using only polylogarithmic space (i.e, in the streaming model).
%The reduction from coresets to $\eps$-approximations allows our framework to rely only on the {\em combinatorial complexity} of the input family $F$ of functions  (i.e., the combinatorial complexity of the clustering problem at hand), and to use the vast literature on $\eps$-approximation to obtain improved results.

For several function families $F$ for which coresets are known not to exist, or the corresponding (approximate) optimization problems are hard, our framework yields \emph{bicriteria} approximations, or coresets that are large, but contained in a low-dimensional space.

We demonstrate our unified framework by applying it on projective clustering problems. We obtain new coreset constructions and significantly smaller coresets, over the ones that appeared in the literature during the past years, for problems such as:
\begin{itemize}
\item $k$-Median [Har-Peled and Mazumdar,STOC'04], [Chen, SODA'06], [Langberg and Schulman, SODA'10];
\item $k$-Line median [Feldman, Fiat and Sharir, FOCS'06], [Deshpande and Varadarajan, STOC'07];
\item Projective clustering [Deshpande et al., SODA'06] [Deshpande and Varadarajan, STOC'07];
\item Linear $\ell_p$ regression [Clarkson, Woodruff, STOC'09 ];
\item Low-rank approximation [Sarl{\'o}s, FOCS'06];
\item Subspace approximation [Shyamalkumar and Varadarajan, SODA'07],  [Feldman, Monemizadeh, Sohler and Woodruff, SODA'10],
[Deshpande,  Tulsiani, and Vishnoi, SODA'11].
\end{itemize}

The running times of the corresponding optimization problems are also significantly improved.
We show how to generalize the results of our framework for squared distances (as in $k$-mean), distances to the $q$th power, and deterministic constructions.
\end{abstract}

%A category including the fourth, optional field follows...
%\category{I.5.3}{Computing Methodologies}{Pattern Recognition}[Clustering, Algorithms]

%\terms{Algorithms, Performance, Theory}

%\keywords{coresets, epsilon-approximations, epsilon-nets, PAC-learning, clustering, approximating, SVD, PCA, CUR, regression, k-means, k-median
%} % NOT required for Proceedings

\section{Introduction}
Over the last couple of decades, much effort has been put in understanding the combinatorial and computational complexity of a wide range of clustering and shape fitting problems.
Given a set of $n$ data elements $P$, one of the powerful techniques used in this context is that of {\em coresets}, i.e., a small set $D$ of representative data elements which approximately represent $P$, in terms of various objective measures.
More precisely, for a set of candidate queries $X$, and a measure function $\cost(P,x)$, the set $D$ is an $\eps$-coreset for $P$ if $\cost(D,x)$ approximates $\cost(P,x)$ for every $x\in X$, up to a multiplicative factor of $1\pm \eps$.
See e.g. \cite{sur} for a nice (but not updated) survey.

Succinct coresets that lead to efficient algorithms appear in a variety of shape fitting and clustering problems. However, their proof of existence and efficient construction is usually tailor made to fit the properties of the problem at hand.
%Thus, coresets constructions and their (usually very technical) proofs of correctness are quite different.
%As a result, the need for a unified framework that will capture and extend the existing constructions is an important research direction.
Moreover, there are several natural clustering problems for which it is proven that no coresets of size $o(n)$ exist. These include, for example, approximating points in $\REAL^3$ by a pair of \emph{planes} \cite{nocry}, the clustering of weighted points in $\REAL^2$ by a set of $2$ {\em lines} \cite{integral}, and approximating a point set by $k$-lines~\cite{integral}, where $k\geq \log n$.
%(in both cases one is to in order to minimize the maximum distance from a point to its nearest plane; see
These kind of clustering problems are usually referred to as projective clustering.

\subsection{This work}

Let $F$ be a set of $n$ functions from $X$ to $\image$.
Throughout this work, each function $f \in F$ will correspond to a data {\em element}, and $x\in X$ will correspond to a \emph{center} (or a set of centers). For a center $x \in X$, the value $f(x)$ corresponds to the cost of evaluating $f$ with the center $x$. The cost of evaluating $F$ with $x\in X$ is defined as $\cost(F, x)=\sum_{f\in F}f(x)$.

Intuitively, the cost function should be interpreted in the context of shape fitting, where $X$ represents a set of shapes, and $f(x)$ represents the cost of fitting an element represented by $f$ to the shape $x$.
For a given query shape $x\in X$, the value $\cost(F,x)$ represents how well $x$ approximates $F$.
%The corresponding optimization problem is to compute a shape $x\in X$ that minimizes $\cost(F,x)$.
In the context of $k$-clustering, the ``center'' $x$ represents a tuple of $k$ centers, and $f(x)$ represents the distance from an element $f$ to its closest center in $x$.
For example, in the well known $k$-median problem in $\REAL^d$, the corresponding set $X$ is $(\REAL^{d})^k$.
For a data element $p \in \REAL^d$, and a {\em center tuple} $x=(x_1,\dots,x_k) \in (\REAL^{d})^k$, the corresponding function $f_p$ is defined as $f_p(x) = \min_i\dist(p,x_i)$.

In this work, we present a unified framework for the efficient construction of coresets for  clustering problems corresponding to a given function set $F$. Our  coresets are obtained via a new and natural reduction to the well studied notion of $\eps$-approximation from the theory of VC dimension \cite{Vap71a}. The reduction from coresets to $\eps$-approximations allows our framework to rely only on the {\em combinatorial complexity} of the input family $F$ of functions  (i.e., the combinatorial complexity of the clustering problem at hand), and to use the vast literature on $\eps$-approximation to obtain improved results (that are at times deterministic).
For several function families $F$ for which coresets are known not to exist, or the corresponding (approximate) optimization problems are hard, our framework yields \emph{bicriteria} approximation, or coresets that are large, but contained in a low-dimensional space.

In the body of the paper, we give an overview of the contributions of our work.
We start by presenting, in Section~\ref{sec:app}, several concrete results that follow from our algorithmic paradigm, including a detailed comparison with corresponding previous work.
We then present the main proof techniques and conceptual novelties in our approach in Section~\ref{sec:tech_intro}.
Finally, in Section~\ref{sec:over}, we present a detailed overview of our algorithms for the construction of corestes and bicriteria approximation.
The above discussion will take up the body of this extended abstract.
All of the technical details of our results appear in the (self contained) appendix.
%full version of this work \cite{FL11full}.
A first application of our framework (for HD-image processing) already appeared in~\cite{ffs11}.
%, which The appendix is designed to be read in a self contained manner, keeping in mind the discussions and outlines presented shortly.

\section{Concrete Contributions}
\label{sec:app}
\subsection{Projective clustering}\label{uni}
Our concrete results are taking from the broad family of projective clustering problems.
In the task of projective clustering we are given a set $P \subset \REAL^d$ of $n \geq d$ data elements, a positive integer $k \leq n$, and a non-negative integer $j \leq d$. A center $x \in X$ is a $k$ tuple $(x_1,\dots,x_k)$ where each $x_i$ is a $j$-dimensional affine subspace (flat) in $\REAL^d$.
The objective is to find a center $x^*$ that minimizes the $\cost(P,x) = \sum_{p \in P}\dist(p,x)$ over $x\in X$.
Here, $\dist(p,x)$ denotes the Euclidean distance from a point $p$ to its nearest subspace $x_i$ in $x = (x_1,\dots,x_k)$.
More generally, for a given $z\geq 1$, we wish to minimize the sum of distances to the power of $z$, i.e, $\sum_{p \in P}\big(\dist(p,x)\big)^z$ .
In this section we define three types of coresets for projective clustering: \vspace{2mm}

%\newpage
\noindent\textbf{Strong coresets:} A weighted set of points $D$ in $\REAL^d$ that approximate the distances to \emph{every} possible $k$-tuple of $j$-flats in $\REAL^d$, up to a multiplicative factor of $(1+\eps)$.
\medskip
\\\noindent\textbf{Weak coresets:} A weighted set of points $D$ in $\REAL^d$, such that a $(1+\eps)$-approximation for the optimal solution of $D$ yields a $(1+\eps)$-approximation for the optimal solution of the full data set $P$. That is, \emph{any} black box algorithm or heuristic that computes a $(1+\eps)$-approximation for the coreset would yield a $(1+\eps)$-approximation for the original set. Hence, a weak coreset can be viewed as a {\em reduction} from the clustering problem with input $P$ to the same problem with input $D$. We note that in previous papers~(e.g., \cite{weak, FMSW08}) the only way to get a PTAS for the original set is to run exhaustive search on the coreset.
\medskip
\\\noindent\textbf{Streaming coresets:} A weak coreset $D$ that is updated online during one pass over the $n$ points of $P$, while using only $O(d\cdot |D|)$-space in memory. Streaming coresets can thus be used online to compute a $(1+\eps)$-approximation for the optimal solution of the points in $P$ viewed so far.

\medskip
All the algorithms that are described in this section are randomized, and succeed with probability at least $1/2$ (or any other constant approaching $1$).

Roughly speaking, the results given in this section are specific applications of our framework which, for general values of $j$, yields a bicriteria approximation $B$ for the projective clustering problem followed by a so called $B$-coreset: $D=\proj(P,B)\cup \T$.
Here, a bicriteria approximation is a set of possibly more than $k$ centers, that approximates the cost of the optimal solution $x^*$ up to some constant factor approximation. The set $\proj(P,B)$ denotes the projection of the data set $P$ onto the bicriteria centers $B$, and $\T$ is a set of $\t$ points. Our sets $D$ have the qualitative properties of coresets.
Namely, for $\t=O(djk/\eps^2)$ the set $D$ we obtain is a strong coreset, for $\t=O(kj^2\log(1/\eps)/\eps^3)$ we obtain weak coresets, and for $\t=O(kj^2\log(1/\eps)\log^4 n/\eps^3)$ streaming coresets.
%Our results generalize to powers $z\geq 1$, at the price of multiplying $\t$ by a factor of $1/\eps^{2z}$, and the construction time of $D$ is $O(ndjk+\t\log n)$.

Our $B$-coresets are constructed by the union of the two sets $\T$ and $\proj(P,B)$.
While $\T$ is of small size $t$, the set $\proj(P,B)$ may be large in size.
Nevertheless, our coresets are of substantial interest as they imply a {\em dimension reduction} from the set $P$ to the set $\proj(P,B)$.
Indeed, when our centers are points (i.e., $j=0$), we are able find a set $B$ of size $k$, so $\proj(P,B)$ is also of size $k$.
When our centers are lines (i.e., $j=1$), the set $\proj(P,B)$ is contained in a small set of lines and we use~\cite{FFS06} to reduce the size of $\proj(P,B)$ to $(\eps^{-1}\log n)^{O(k)}$.
We discuss these cases and others (derived from our framework) in the subsections to come.

The construction time of the strong and weak coresets is $O(ndjk+\t\log n)$.
All our coresets and running times below are generalized to sum of distances to the power of $z>1$, after replacing the term $\eps$ in the corresponding results by $1/\eps^{2z}$.

\subsection{$k$-Median and its generalizations}
We start by discussing the setting in which the centers $X$ are $k$-tuples of points in $\REAL^d$ (i.e., $j=0$).
\medskip
\\\noindent
{\bf Strong coresets:}
For the case $j=0$ and $z=1$, which is the standard $k$-median problem, we present a {\em strong} coreset of size $\t=O(dk/\eps^2)$.
This improves on previous results in \cite{sarielb,KeChen06,LS10}, where the construction of $\e$-coresets of size $O(k^3\e^{-d-1})$, $O(k^2d\e^{-2}\log{n})$, and $\tilde O(d^2k^3\e^{-2})$, is respectively presented. The term $\tilde O(x)$ hide factors that are poly-logarithmic in $x$.

\medskip
For general metric spaces (e.g., $\dist(p,x)$ is defined as the distance between $p$ and $x$ in the given metric), the dimension $d$ is to be replaced by $\log n$, implying strong coresets of size $\t=O(k\log({n})/\eps^2)$.
This improves on the result of Ke Chen~\cite{KeChen06}, which gives a coreset of size $O(k^2\log{(n)}/\eps^2)$ for this problem.
Both our results and those of \cite{LS10} are generalized to cost functions which use a power $\zz$ of the distance, namely $\cost(P,x) = \sum_{p \in P}(\dist(p,x))^z$.
\medskip
\\\noindent\textbf{Weak coresets.}
For the $k$-median problem, our framework yields a weak coreset $D$ of size $O(k\log(1/\eps)/\eps^3)$.
By computing a $(1+\eps)$-approximation to the $k$-median of $D$, we are able to compute a set of $k$ centers that gives a $(1+\e)$ approximation to the optimal centers for $P$ in time $O(ndk+2^{\poly(1/\eps,k)})$. Our results generalize to any integer $z>1$ by replacing $\eps$ with $\eps^{2z}$ in the corresponding time and space term.

For the case of $z=1,2$ (median and mean problems), Ke-Chen~\cite{KeChen06} suggested an $O(ndk)+\poly(d,\log n)\cdot 2^{\poly(k/\eps)}$ PTAS.
For the $k$-mean case ($z=2$), Feldman, Monemizadeh and Sohler \cite{weak} improved this result using a weak coreset of size \\$O(k\log^2k\log(1/\eps)/\eps^5)$, that yields a PTAS that takes time $O(ndk)+d\cdot \poly(k/\eps)+2^{\tilde O(k/\eps)}$.
\medskip
\\\noindent\textbf{Streaming coresets.}
Our framework yields streaming coresets of size $\t=O(k\log(1/\eps)\log^4 (n)/\eps^3)$ for $k$-median and its generalizations for $z>1$. This improves on the result of Ke Chen~\cite{KeChen06} which suggests a streaming coreset of size $O(dk^2\eps^{-2}\log^8n)$ for $z=1,2$.
We note that Feldman, Monemizadeh and Sohler \cite{weak} present a streaming coreset of size $\poly(k\log n/\eps)$ for the special case of $k$-mean ($z=2$).
To the best of our knowledge, no streaming coresets of size independent of $d$ were known for the case $z>2$.

\subsection{$k$-Line median and its generalizations}
In this case, we seek to cluster the points in $P$ by $k$ lines in $\REAL^d$ (i.e., we take $j=1$). Very little is known about this problem in high dimensional space.
\medskip
\\\noindent\textbf{Strong coresets.}
Combining our results with techniques presented in \cite{FFS06}, we obtain strong coresets for this problem of size $(\log(n)/\eps)^{O(k)} + O(dk/\eps^2)$. This improves on the previous work of \cite{FFS06} that for $z=1,2$ introduces coresets of size $\log^{O(k)}n/\eps^{O(d\log d+k)}$.
\medskip
\\\noindent\textbf{Weak coresets.}
The best PTAS (prior to our work) for this problem takes time $dn\cdot \poly(k/\eps)+n(\log n)^{\poly(k/\eps)}$; see~\cite{DesVar07}.
We suggest a weak coreset for this problem of size $(\log(n)/\eps)^{O(k)}$
which improves the running time of this result to $O(ndk)+(\log n)^{\poly(k/\eps)}$.
\medskip
\\\noindent\textbf{Streaming coresets.} We construct the first streaming coreset for this problem. Its size is $(\log (n)/\eps)^{O(k)}$.

\subsection{Subspace approximation}
In the problem of subspace approximation one seeks a single $j$-flat that approximates the data set $P$ (i.e., in our notation $k=1$).
\medskip
\\
\textbf{Strong coresets.}
We suggest a strong coreset of size  $\t=O(dj/\eps^2)$ for any $j\geq 1$.
This is the first strong coreset of size polynomial in $d$ for approximating the sum of distances to any $j$-dimensional subspace.
In~\cite{FFS06} a strong coreset of size $(1/\eps)^{\poly(j,d)}\cdot \log^{O(j^2)} n$ is constructed in $nd\cdot j^{O(j)}$ time.

For the case $z=2$ and $j=d-1$ (sum of squared distances to a hyperplane) Baston, Speilman and Srivastava~\cite{BSS09} recently proved that there is a coreset of size $O(d/\eps^2)$ which is a also a weighted subset of $P$. Many applications of this construction were suggested in~\cite{naor11}. Such a coreset can be constructed directly from Theorem~\ref{mainmain} below in time $O(nd^2+d/\eps^2)$, with high probability, while~\cite{BSS09} provide a deterministic construction in $O(nd^4/\eps^2)$ time.
Unlike the above constructions, our results can be generalized for any $z\geq 1$ and $j\leq d-1$ where $\eps$ is replaced by $\eps^{2z}$ in the running time and coreset's size. Deterministic constructions of such coresets can  can be computed in time $n\cdot(1/\eps)^d$ using the de-randomization technique of~\cite{Matousek95}.
\medskip
\\\noindent\textbf{Weak coresets.}
We obtain a weak coreset of size $O(j^2\log(1/\eps)/\eps^3)$ for the subspace approximation problem that yields an $O(dnj)+ 2^{\poly(j,1/\eps^2)})$ time PTAS.
A result of Shyamalkumar and Varadarajan~\cite{ShyVar07} and subsequent work by Deshpande and Varadarajan~\cite{DesVar07} gave a $(1+\eps)$-approximation algorithm for the case $z\geq 1$, with running time $dn\exp(j,1/\eps^z)$. For the case $z=1$, the running time was recently improved to $O(dn\poly(j,1/\eps)+O(d+n)\exp(j,1/\eps)$ by Feldman, Monemizadeh, Sohler and Woodruff~\cite{FMSW08}.
\medskip
\\\noindent\textbf{Streaming coresets.}
Our streaming coresets for subspace approximation are of size $\t=O(j^2\log(1/\eps)\log^4 n/\eps^3)$, and thus use $O(d\cdot \t)$ space.
Sarlos~\cite{sarlos} provides a streaming algorithm that requires two passes over the data and uses space $O(n)(k/\eps+k\log k)^2$.

For the case of non constant $j$, Deshpande, Tulsiani, and Vishnoi recently showed that computing a PTAS for this
problem is ``hard"~\cite{DTV11}. However, they suggested a constant factor approximation using a relaxation to convex programming, which takes time $d\cdot \poly(n)$. Applying this algorithm on the output coresets of our framework would thus yield a constant factor approximation in $O(dn+d\cdot\poly(j))$ time together with a strong, and streaming coreset.
\medskip
\\\noindent\textbf{CUR Decomposition.} Given $j\geq 1$ and an $n\times d$ matrix $A$, the CUR decomposition $\tilde{A}=CUR$ consists of an $n\times m$ matrix $C$, $m\times j$ matrix $U$, and $j\times d$ matrix $R$, such that:  (i) The columns of $C$ are subset of columns from $A$, and the rows of $R$ are a subset of rows from $A$. (ii) $\tilde{A}$ minimizes $\sum_{i=1}^n\norm{a_i-\tilde{a}_i}_2^z$ over every $\tilde{A}$ of rank $j$, up to a multiplicative factor of $(1+\eps)$. Here, $a_i$ and $\tilde{a}_i$ are the $i$th row of $A$ and $\tilde{A}$, respectively.

For the case $z=2$, Boutsidis et al.~\cite{BDM11} provide $(2+\eps)$ randomized and deterministic CUR decompositions using $m=O(j/\eps)$ columns. They also provide an updated reference for this long line of research.
Mahoney and Drineas suggested a randomized algorithm that yields a $(1+\eps)$-approximation for the case $z=2$~\cite{mahoney2009cur}.

To the best of our knowledge, the CUR a decomposition is not discussed for $z\neq 2$ or for the streaming model. Since all the approximated $j$-subspaces that are described in this paper are spanned by $\poly(j/\eps)$ input points, our coresets yields corresponding $(1+\eps)$-approximation for the CUR decomposition in these cases using the observations from~\cite{mahoney2009cur}.
\medskip
\\\noindent\textbf{Linear regression.}
In the $\ell_1$ regression problem, the input is an $n\times (d-1)$ matrix $A$ and a
vector $b\in\REAL^{n}$. The the goal is to minimize $||Ay-b||_1$ over all $y\in\REAL^{d-1}$.
By defining a set $P$ of $n$ points in $\REAL^d$ that correspond to the rows of the matrix $[A|b]$, and mapping any vector $y\in \REAL^{d-1}$ to the hyperplane $x$ that is orthogonal to the vector $[y^T,-1]^T$, it is easy to verify that a strong coreset for the subspace approximation of $P$ with $j=d-1$ would yield a strong coreset for the corresponding linear regression problem for $A$, $b$.
%We have that \[\norm{[y^T,-1]^T}_2\cdot \cost(P,x)=\norm{[A|b]\cdot [y^T, -1]^T}_1=\norm{Ay-b}_1.\] Hence,

In particular, our strong coresets for subspace approximation with $j=d-1$ yield a strong coreset for the linear regression problem of size $\t=O(d^2/\eps^2)$. The construction time is $O(nd^2+d^2\eps^{-2}\log n)$. Computing the $\ell_1$ regression on the strong coreset would thus take $O(nd^2+\poly(d/\eps))$ time (e.g., using~\cite{DasguptaDHKM08}). Maintaining these strong coresets in the streaming model will yield a streaming algorithm that takes space $\t=O(d^2\log ^2n/\eps^2)$. As mentioned in the beginning of Section~\ref{sec:app}, the results are generalized for any $z\geq 1$ where $\eps$ is replaced by $\eps^{2z}$ in our running time and size of coresets.

\medskip
Efficient approximation algorithms for the regression problem are given by Clarkson~\cite{clarkson} for $z = 1$, Drineas, Mahoney, and Muthukrishnan~\cite{DrineasMM06} for $z = 2$, and Dasgupta et
al.~\cite{DasguptaDHKM08} for $z\geq 1$ in time $O(nd^5\log n+\poly(d/\eps))$. All these results are obtained by constructing weak coresets for the corresponding problem.
Some small space streaming algorithms are available in the turnstile model (where the points are constrained to be on an integer grid of size $n^{O(1)}$) for $l_z$ regression where $1\leq z\leq 2$ by~\cite{FMSW08} and~\cite{ClarksonW09} for $z=2$.
However, we are not aware of previous strong or streaming coresets for the original (unconstrained) problem.

\subsection{Projective clustering}
We now discuss the broad setting in which both $j$ and $k$ may be arbitrary.
When $j \geq 2$ and  $k$ is taken to be general, there are no strong coresets (of size $o(n)$) for these problems, even for $j=k=2$ and $d=3$; this can be proven using a simple generalization of the results of~\cite{nocry}. Also, for $k>\log n$, the optimization problem cannot be approximated in polynomial time, for any approximation factor, unless {\sl P=NP}~\cite{tamir}.
However, the problem does allow
one of the following bicriteria approximations (where one allows some leeway in both the number
or dimension of flats and the quality of the objective function).
In what follows, an $(\alpha,\beta)$ bicriteria solution is a set $B$ of $\beta$ flats such that clustering the points $P$ via $B$ can be done at a cost at most $\alpha$ times the optimal $k$ clustering.
We now present our results in this context.
\medskip
\\\noindent
{\bf Bicriteria Approximations.}
Giving a set of points in $\REAL^d$, whose minimum enclosing ball is of radius $r^*$, suppose we want to compute a set of $O(\log n)$ balls of radius at most $r^*$ that covers $P$. There is a generic and simple greedy algorithm that compute such a set in $O(nd)$ time using the theory of VC-dimension~\cite{bron95}. This algorithm works for any family of shapes of small VC-dimension. In this paper we generalize this algorithm for the case of non-covering problems. In general, our bicriteria algorithm has many advantages over previous work~(e.g.,~\cite{ind99, CzuSoh07a}), both in the fact that it is widely applicable (for a general families of functions, not necessarily metric spaces), more efficient (in terms of the approximation factors and running time), and implies deterministic constructions.

In the context of projective clustering, in~\cite{FFSS07}, an $(\alpha,\beta)$-bicriteria approximation algorithm was suggested, which produces, with high probability,
at most $\beta(k,j,n) = \log n\cdot(jk \log \log n)^{O(j)}$ flats of dimension $j$, which exceed the optimal
objective value for any $k$ $j$-dimensional flats  by a factor of $\alpha(j)= 2^{O(j)}$. The running time is $dn\log n\cdot (2k)^{\poly(j)}$.
Our framework improves (the running time, $\alpha$ and $\beta$) upon this result and yields several bicriteria approximations algorithms.
For small values of $j$ and $k$, we present a bicriteria algorithm that yields an $\alpha= 1 + \eps$ approximation. It returns $\beta = k\log n$ flats in time
$O(dnjk)+d\cdot \poly(j,k,1/\eps)+2^{\poly(j,k,1/\eps)}\log^2 n$.
For large values of $k$, we suggest a $(1+\eps, \beta)$-approximation that returns $\beta = \log n\cdot k^{\poly(j,1/\eps)}$ flats of dimension $j$, and the running time is $O(dn\beta) + d\cdot \poly(j,k,1/\eps)\cdot \log^2 n$.
\medskip
\\\noindent
{\bf Low-Dimensional $B$-Coresets for large $j$.}
Deshpande and Varadarajan~\cite{DesVar07} describe an algorithm that
returns a subspace $V$ spanned by $\poly(jk/\eps)$ points that is guaranteed, with probability at least
$1/2$, to contain $k$ $j$-subspaces whose union is a $(1+\eps)$-approximation to the optimum solution.
Using the volume sampling technique their algorithm runs in $dnj^3k^3(jk/\eps)^z$ time for any $z\geq 1$.

Note that this result does not have the reduction property of weak coresets as defined in the beginning of this section. That is, even if we have an algorithm that computes the optimal set $x^*$ of $k$ $j$-subspaces for any given set of points, it is not clear how to use it with $V$ in order to have a more efficient solution for the original problem. Similarly, it seems that this result can not be generalized for the streaming model when the subspace $V$ needs to be computed for a stream of $n$ points $P$ using less than $O(nd)$ space.

For these problems (where $k,j>1$), we suggest strong, weak, and streaming coresets contained in low-dimensional subspaces,
and therefore take sub-linear space. Our coresets, referred to as $B$-coresets, were described in Section~\ref{uni}, and are used as the first step for the construction of all the coresets presented in this section  (including when $j=1$ or $k=1$).

\section{Novelties in proof techniques}
\label{sec:tech_intro}

As specified in Section~\ref{sec:app}, our unified framework yields a number of improved results in the context of approximate clustering and shape fitting.
In what follows, we briefly touch on the major new ideas used in our algorithms allowing theses improved results.
\medskip
\\\noindent
{\bf Reduction to $\eps$-approximation:}
The main reason that our framework is able to address a spectrum of clustering and approximation problems lies in our reduction from the inconsistent definition of coresets to the notion of  $\eps$-approximation.
Using this reduction we can:
\textbf{(i)} use a common ground in our analysis, thus removing the specialized (and sometimes tedious) analysis of the required sampling sizes used in many of the related works mentioned in Section~\ref{sec:app}.
\textbf{(ii)}  use smaller sample sizes that improve on those obtained in previous works, due to recent results taken from the context of Machine Learning~\cite{li00improved}.
\textbf{(iii)} apply numerous results from the field of Computational Geometry, dated back to~\cite{HauWel86},  regarding the study of VC-dimension and $\eps$-approximations. For example: deterministic constructions~\cite{Matousek95}, for convex shapes (which have unbounded VC-dimension)~\cite{ChazelleEtAl95b}, and in the streaming model~\cite{Bagchi:2007:DSR}.

Our reduction includes multiple stages and uses the new notions of {\em robust approximation} and {\em robust corests} as intermediate points.
We elaborate on our reduction to $\eps$-approximation (including our new notions) in the upcoming Section~\ref{sec:over} which addresses a detailed overview of our framework.
\medskip
\\\noindent
{\bf Functional representation of data elements and coresets:}
To study coresets over a wide range of objectives, we present an abstract framework in which the data points are considered as functions. Namely, for a center $x$, the value $f(x)$ represents the cost of clustering the data element corresponding to $f$ with $x$.
This representation is not superficial, and is in a sense crucial, as in our setting the coresets we construct are no longer ``data elements'' (as is common in the literature) but rather functions as well.
Indeed, in some cases, our coresets will correspond to a subset of data elements, and thus their representation by functions will have no special meaning. However, in several cases the coreset consists of a small set of functions, that are closely related to the original data functions, however differ in certain behaviors.

For example, several of our coresets use functions $g$ corresponding to the data functions $f$ such that $g(x) = f(x)$ only if $f(x)$ is smaller than a certain threshold; otherwise $g(x)$ will be {\em neglected} and equal to zero. Another example includes the use of functions $g$ that correspond fully to data elements $f$, but appear in the coreset as having {\em negative} weight. We extend and generalize results from~\cite{FMSW08} that had such properties. However, unlike in~\cite{FMSW08}, a PTAS for the optimization problem can be computed from the coresets without using the original data.

One may argue that this skewed succinct representation of the original data violates the traditional line of thought in which a coreset consists of a subset of ``real'' data elements, and thus in many cases we make an effort in finding such ``standard'' coresets. However, when considering the computational objective in the construction of coresets, namely a tool to allow the efficient approximation of clustering problems, our notion of coresets plays a role equivalent to that of standard coresets. The flexibility in allowing our coresets to deviate from standard conception is a key point in our ability to obtain improved results.
\medskip
\\\noindent
{\bf Generalized range spaces:}
In the vast literature on clustering, the notion of coresets is defined in several ways. Two common definitions include strong and weak coresets, which roughly speaking, address the combinatorial and computational aspects of clustering respectively. Namely, strong coresets require a similar behavior when compared to the data set for {\em every} set of centers, while weak coresets require ``just enough'' so that the coreset can be used in the design of efficient algorithms for approximate clustering.

In this work we unify the study of weak coresets that was used recently in~\cite{sur, weak, FMSW08} with older results related to $\eps$-approximation~\cite{ChaFri90}, called \emph{$\eps$-frames}.
As our work reduces the study of coresets to that of $\eps$-approximation in certain range spaces, this unification is captured by the development of a new notion:  a {\em generalized range space} and a corresponding {\em generalized dimension}.

More specifically, in the standard study of range spaces, an $\eps$-approximation captures the propertied of the original space with respect to {\em any} range in the space. This intuitively corresponds to the study of strong coresets. For the (more delicate) study of weak coresets, we enhance the standard definition of a range space, to obtain a generalized definition and theory.
In our generalized view, an $\eps$-approximation captures the propertied of the original space with respect to a {\em subset} of predetermined ranges in the space (and not necessarily all of the ranges).
Choosing the predefined subsets carefully, one may capture the essence of weak coresets.
The study of generalized range spaces enables us to use the same algorithms in our constructions of coresets, whether weak or strong, where the difference in the obtained results (in size and running time) is now easily traced back to the notion of the generalized dimension of the range space at hand.

\section{Framework overview}
\label{sec:over}

We now review the concept of $\eps$-approximations and $\eps$-coresets followed by a detailed overview of our general framework.
\subsection{$\eps$-Approximations and coresets}
\noindent%\textbf{$\eps$-approximations.}
For a multi-set $F$ of non-negative functions on a set $X$, we say that $S\subseteq F$ is an $\eps$-approximation for $F$, if for every every $x\in X$ and $r\geq 0$ we have
\[
  \left|\frac{\range(F,x,r)}{|F|}- \frac{\range(S,x,r)}{|S|}\right|\leq \eps.
\]
where $\range(S,x,r)=\br{f\in S\mid f(x)\leq r}$.
%For the above example of query balls, we have $X=\REAL^d$, every $f_p\in F$ represents a point $p\in P$, and $f_p(x)=\dist(p,x)$ is the Euclidean distance from $p$ to $x\in\REAL^d$. Hence, $\range(F,x,r)$ is the intersection of $P$ with the ball of radius $r$ that is centered at $x$.
%
\medskip
\\\noindent
%\textbf{$\eps$-coresets.}
%Given a set $P$ of $n$ points in $\REAL^d$, suppose that we wish to compute the sum of Euclidean distances from $P$ to a query point $x\in\REAL^d$, i.e, $\sum_{p\in P}\dist(p,x)$. For an error parameter $\eps>0$, an $\eps$-coreset for this problem is a set $D$ such that, for every $x\in\REAL^d$, we have that $\sum_{p\in D}\dist(p,x)$ approximates $\sum_{p\in P}\dist(p,x)$ up to a multiplicative error of $\eps$.
For a set $F$ of non-negative functions on a set $X$, we say that $D$ is an $\eps$-coreset for $F$, if for every $x\in X$ we have
\[
  (1-\eps)\cost(F,x) \leq \cost(D,x)\leq (1+\eps)\cost(F,x),
\]
where $\cost(F,x)=\sum_{f\in F}f(x)$ and $\cost(D,x)=\sum_{f\in D}f(x)$.
\medskip
%\\\noindent\textbf{From $\eps$-approximation to $\eps$-coresets.}
In this paper we forge a link between $\eps$-approximations and $\eps$-coresets for general families of queries.
As a warm-up, we present the following theorem which is a special case of our main theorem (Theorem~\ref{the:coresetA_intro}). It relates to the notion of \emph{sensitivity} that was introduced in~\cite{LS10} for $k$-median type problems.
\newcommand{\N}{\mathbb{N}}
\begin{theorem}\label{mainmain}
Let $F$ be a set of functions from $X$ to $[0,\infty)$ and $0<\eps<1/4$.
Let $m:F\rightarrow \N\setminus \br{0}$ be a function on $F$ such that
\begin{equation}\label{cc}
m(f)\geq n\cdot \max_{x\in X}\frac{f(x)}{\cost(F,x)}.
\end{equation}

For each $f\in F$, let $g_f:X\rightarrow [0,\infty)$ be defined as $g_f(x)=f(x)/m(f)$.
Let $G_f$ consists of $m_f$ copies of $g_f$, and let $S$ be an $(\eps \cdot n/\sum_{f\in F}m(f))$-approximation of the set $G=\bigcup_{f\in F}G_f$.
Then $D=\br{g_f\cdot |G|/|S|\mid g_f\in S }$ is an $\eps$-coreset for $F$. That is, for every $x\in X$,
\[
 |\cost(F,x)-\cost(D,x)|\leq \eps\cost(F,x).
\]
\end{theorem}

For example, suppose that we are given a set $P$ of $n$ points in $\REAL^d$, and we wish to compute a small set of functions $D$ such that, for every $x\in\REAL^d$, we will have that $\cost(D,x)$ is a $(1+\eps)$-approximation to the sum of Euclidean distances $\sum_{p\in P}\norm{p-x}_2$.
For every $p\in P$ and $x\in X=\REAL^d$, let $f_p(x)=\norm{p-x}_2$ and $F=\br{f_p\mid p\in P}$. Let $x^*$ denote the point that minimizes the sum of distances to $P$, and define
\[
m(f_p)=\left\lceil \frac{n\cdot f_p(x^*)}{\cost(F,x^*)}\right\rceil +2.
\]
It is not hard to verify that~\eqref{cc} holds for this definition of $m(f_p)$ and $\sum_{f\in F}m(f)=O(n)$; see~\cite{LS10}. By the PAC-learning theory, a random sample $S\subseteq G$ of size $O(d/\eps^2)$ is an $\eps$-approximation of the set $G$ that is defined in Theorem~\ref{mainmain}, with high probability; see~\cite{LiLonSri01a}. By Theorem~\ref{mainmain} we conclude that there exists a set $D$, $|D|=O(d/\eps^2)$, such that $|\cost(F,x)-\cost(D,x)|\leq \eps\cost(F,x)$ as desired. In the next sections we present tools that allow us to compute such a small coreset $D$ efficiently, deal with high dimensional spaces (say, when $d=n$), and with $k$-clustering problems (for example, when $x=(x_1,\cdots, x_k)$ and $f_p(x)=\min_{i}\norm{p-x_i}$).

\subsection{Bicriteria approximation}

%\begin{figure}[t]
%\begin{center}

\begin{figure}[!t]
\centering
\subfigure{\fbox{\begin{minipage}{10cm}
\begin{tabbing}
\noindent\textbf{Algorithm} {\bicriteria}$(F, \eps,\alpha,\beta)$\\
\end{tabbing}
\vspace{-2.5em}
\begin{codebox}
\li $i\gets 1$; $F_1\gets F$
\label{definitions_intro}
\li \While $\displaystyle |F_i|\geq 10/\eps$ \Do \label{boundq_intro}
%\li $\mincost \gets\infty$
\li $\M_i\gets$ A $(3/4,\eps,\alpha,\beta)$-median of $F_i$
\label{lineAlgone_intro}
\li $G_i\gets$ The set of the $\big\lceil (1-5\eps)\cdot3|F_i|/4\big\rceil$ functions \\
\ \ \ \ \ \ \ \ \ \ \ \ \ \  \ \ \ $f\in F_i$
  with the smallest value $f(\M_i)$.\label{linefour_intro}
\li $F_{i+1}\leftarrow F_i\setminus G_i$\label{addf_intro}
\li $i\gets i+1$ \;
\End
\li $\M_i\gets$ An $(\alpha,\beta)$ bicriteria to $F_i$
%A subset $X'$ of $X$ of size $\beta$)$-median of $F_i$
\label{lastptas_intro}
%\li $G_i\gets F_i$\label{Gi_last}
\li \Return $\cup Y_i$
\end{codebox}
\end{minipage}}}\caption{The algorithm~{\bicriteria}.\label{fig:alg_bi_intro}}
\end{figure}

As common in several studies of geometrical clustering, our starting point is that of bicriteria approximation.
Given the function family $F$, and a set of potential centers $X$, an $(\alpha,\beta)$ bicriteria solution to the clustering problem $(F,X)$ is a subset $B$ of $X$ of size $\beta$ such that
$\cost(F,B) \leq \alpha \min_{x \in X}\cost(F,x).$
Here, for a set $B$, the term $\cost(F,B)$ is equal to $\sum_{f \in F} f(B)$, where $f(B)$ is a slight abuse of notation which represents the expression $\min_{x \in B}f(x)$.
Efficient bicriteria approximation algorithms for constant values of $\alpha$ and $\beta$ have been extensively studied over the last decade for a number of function families $F$. For example, in \cite{sariela,KeChen06, FFS06,weak, FFKN08, FMSW08,LS10} the starting point for the efficient construction of small $\e$-coresets for $k$-median is an efficient bicriteria algorithm for $k$-median. Bicriteria approximation was also used as a starting point for computing clustering in the setting of outliers and penalties; see \cite{Cha01, Chen08}.

%During the last few years, several connections between bicriteria approximation, coresets, and approximate clustering have been presented.
The first part of our framework yields a general paradigm for bicriteria approximations, that essentially reduces the task at hand to that of $\eps$-approximations from the theory of Machine/PAC Learning and VC dimension \cite{Vap71a, HauWel86}.
Roughly speaking our reduction includes three steps.
In the first step, we determine the {\em combinatorial complexity} of the clustering problem at hand by defining a corresponding {\em generalized range space} and studying its {\em generalized VC-dimension} (we elaborate on these notions shortly).
We then show that an $\eps$-approximation to the corresponding range space, yields a relaxed notion of bicriteria clustering we refer to as a {\em robust median}.
Finally, we show how to use these robust medians in able to obtain a bicriteria solution.
An outline of our framework follows.
\medskip
\\\noindent
{\bf Generalized VC dimension:}
Given the clustering problem at hand (i.e., the function family $F$), one starts by defining a corresponding range space and by studying its combinatorial complexity (i.e., {\em dimension}).
%We start by defining our notion of a generalized range space, and then turn to define the range space corresponding to $F$.
%The below definition also appears, e.g., in \cite{li00improved}.

\begin{definition}[e.g., \cite{li00improved}]\label{dimfunctions_intro}
Let $F$ be a finite set of functions from a set $X$ to $\image$.
The \emph{dimension} $\dim(F)$ of $F$ is the dimension of the range space $\big(F,\ranges(F)\big)$, where $\ranges(F)$ is the \emph{range space of $F$}, that is defined as follows. For every $x\in X$ and $r\geq 0$, let $\range(x,r)=\br{f\in F\mid f(x)\leq r}$. Let the set $\ranges(F)$ be defined as $\br{\range(x, r)\mid x\in X, r\geq 0}$.
The dimension of $(F,\ranges)$ is the minimum $d$ such that
$$
\forall S \subseteq F: \ \left|S \cap \ranges(F)\right| \leq |S|^d
$$
%An $\eps$-approximation for the range space $(F,\ranges(F))$ is called an $\eps$-approximation for $F$.
\end{definition}

%As common in the study of VC-dimension, an $\eps$-approximation to $(F,\ranges(F))$ is a subset $S \subseteq F$ %that approximates $F$ with respect to each $\range$ in $\ranges(F)$. Namely,
%$$
%\forall \range \in \ranges(F):\ \left| \frac{|\range|}{|F|}-\frac{|\range \cap S|}{|S|}\right| \leq \eps
%$$

To allow the unified study of both strong and weak coresets, we enhance the definition above to that of a generalized range space.
In a generalized range space corresponding to $F$, for every subset $S$ of functions one defines a corresponding subset of {\em important} ranges $\ranges(S) \subset \ranges(F)$.
In our context of clustering, the set $\ranges(S)$ will be defined by a subset $\XX(S)$ of centers $x \in X$ that are guaranteed to include a {\em good} center to be used in the clustering of $S$.
%Accordingly, $S$ is an $\eps$-approximation if
%$$
%\forall \range \in \ranges(S):\ \left| \frac{|\range|}{|F|}-\frac{|\range \cap S|}{|S|}\right| \leq \eps
%$$
More precisely:

\begin{definition}\label{dimfunctionsHigh2_intro}
Let $F$ be a finite set of functions from a set $X$ to $\image$. Let $\XX$ be a function that maps every subset $S\subseteq F$ to a set of items $\XX(S)\subseteq X$.
The pair $(F, \XX)$ is called a \emph{generalized function space}, if for any $S \subseteq S'$ it holds that $\XX(S) \subseteq \XX(S')$.
The dimension of $(F, \XX)$ is the smallest integer $d$, such that
\[
\forall S\subseteq F:\Big|\br{S\cap\range \mid \range\in\ranges(S)}\Big|\leq |S|^{d}\enspace.
\]
where $\ranges(S)=\br{\range(x, r)\mid x\in \XX(S), r\geq 0}$.
\end{definition}

For a generalized function space $(F,\XX)$, we now seek small subsets $S \subseteq F$ that are $\eps$-approximations to the range space $(F,\ranges(S))$. Loosely speaking, such sets will approximate the function set $F$ with respect to the centers in $\XX(S)$ that are (by definition) of ``importance'' to the approximation of $S$. Combining this with a proof that centers that approximate $S$ also approximate $F$, will yield the weak coresets we desire.
Notice that in the above definition we have required the function $\XX$ to be monotone.
This allows us to obtain the following (immediate) connection between random sampling and $\eps$-approximation (e.g., via \cite{LiLonSri01a}).

\begin{theorem}\label{mycor2_int}
Let $(F,\XX)$ be a function space of dimension $d$ from $X$ to $[0,\infty)$. Let $\eps,\delta>0$. Let $S$ be a sample of
$|S|=\frac{c}{\eps^2}\left(d+\log\frac{1}{\delta}\right)$
i.i.d functions from $F$, where $c$ is a sufficiently large constant. Then, with probability at least $1-\delta$, $S$ is an $\eps$-approximation of the range space $(F,\ranges(S))$.
%$F_{\ind \XX(S)}$.
\end{theorem}
\medskip
To illustrate our definitions, consider the standard problem of $k$-median in $\REAL^d$.
Here, the range space corresponding to $F$ in Definition~\ref{dimfunctions_intro} has dimension $O(dk)$.
Thus, using this range space in our work would imply weak coresets and algorithms with running time that depends in an undesired fashion on $d$.
As all our algorithms at their core are based on the notion of $\eps$-approximation, to avoid this dependence on $d$, it suffices to define a generalized function space of dimension that is independent of $d$.

Indeed, using the results of \cite{ShyVar07} it can be shown that every subset $S$ of $F$ has a {\em low dimensional} corresponding set of centers (set of $k$-tuples) $\XX(S)$ such that $\min_{x\in \XX(S)}\cost(S,x) \leq (1+\eps)\min_{x \in (\REAL^d)^k}\cost(S,x)$.
Specifically, $\XX(S)$ will consist of all $k$-tuples $x$ in the subspaces spanned by $\eps^{-1}\log(\eps^{-1})$ points in $S$.
It is not hard to verify that the dimension of $(F,\XX)$ is now $O(k \eps^{-1}\log(\eps^{-1}))$, and thus independent of $d$.
Which finally yields a succinct $\eps$-approximation $S$ via Theorem~\ref{mycor2_int} that approximates $F$ on all centers in $\XX(S)$.
%As $\XX(S)$ includes good centers for $S$, our framework will finally yield good centers for $F$.
\medskip
\\\noindent
{\bf From $\eps$-approximation to robust medians:}
In what follows we define the {\em robust median} problem, which is a relaxed version of bicriteria clustering which strongly resembles the problem of clustering with outliers.
In a nutshell, a robust median for a set of data elements (functions) $S$, is a set of centers $Y \subset X$ that cluster all but a small fraction of the elements in $S$ very efficiently.
In the below definition, the parameter $\alpha$ represents to the quality of clustering, the parameter $\beta$ refers to the size of $Y$, the parameter $\gamma$ refers to the amount of outliers, and $\eps$ is a slackness parameter.

\begin{definition}\label{robustapprox_int}
Let $F$ be a set of $n$ functions from a set $X$ to $\image$. Let $0<\eps,\gamma<1$, and $\alpha>0$. For every $x\in X$, let $F_x$ denote the $\big\lceil \gamma n\big\rceil$ functions $f\in F$ with the smallest value $f(x)$. Let $Y\subseteq X$, and let $G$ be the set of the $\lceil(1-\eps)\gamma n\rceil$ functions $f\in F$ with smallest value $f(Y)=\min_{y\in Y}f(y)$. The set $Y$ is called a \emph{$(\gamma,\eps,\alpha,\beta)$-median} of $F$, if $|Y|=\beta$ and
\[
\sum_{f\in G}\min_{y\in Y}f(y)\leq \alpha\min_{x\in X}\cost(F_x, x)\enspace.
\]
\end{definition}

Notice that a set of centers $Y$ which are a $(1,0,\alpha,\beta)$-median are (by definition) an $(\alpha,\beta)$ bicriteria approximation.
Thus, one is interested in finding good robust medians for $F$.
We show that this is possible via $\eps$-approximations $S$ to the function space $(F,\XX)$.
In the lemma below we use $\beta=1$.
We note that a similar lemma, for general $\beta$, also holds, and appears in the appendix.

\begin{lemma}\label{functions_intro}
Let $(F,\XX)$ be a function space of dimension $d$.
Let $\gamma\in (0,1]$, $\eps \in (0,1/10)$, $\delta\in (0,1/10)$, $\alpha>0$.
Let $S$ be a random sample of
$
s=\frac{c}{\eps^4\gamma^2}\left(d+\log\frac{1}{\delta}\right),
$
i.i.d functions from $F$, where $c$ is a sufficiently large constant. Suppose that $x\in\XX(S)$ is a $((1-\eps)\gamma,\eps,\alpha,1)$-median of $S$, and that $|F|\geq s$. Then, with probability at least $1-\delta$, $x$ is a $(\gamma,4\eps,\alpha,1)$-median of $F$.
\end{lemma}
Once the connection between $\eps$-approximation and robust medians is established, one can find robust medians for $F$ via an exhaustive (or sometimes more efficient) algorithm that addresses the $\eps$-approximation $S$.
\medskip
\\\noindent
{\bf From robust medians to bicriteria.}
We are now ready to present our algorithm for bicriteria approximation.
Before presenting our algorithm, we note that although an $(\alpha,\beta)$-bicriteria approximation is precisely a $(1,0,\alpha,\beta)$-median, we cannot use Lemma \ref{functions_intro} above to obtain a bicriteria solution (as in Lemma~\ref{functions_intro}, $\eps>0$ and there is a slackness in the reduction w.r.t. $\gamma$).

Our algorithm \bicriteria$(F, \eps,\alpha,\beta)$ for bicriteria approximation appears in Figure~\ref{fig:alg_bi_intro}.
The algorithm receives the function family $F$ and parameters $\alpha,\beta, \eps$ and outputs a subset of centers of size logarithmic (in $|F|$) that act as a bicriteria approximation to the median problem on $F$.
The main recursive call for ``$(3/4,\eps,\alpha,\beta)$-median'' in \bicriteria\ is to the computation of a $(3/4,\eps,\alpha,\beta)$-median for $F$ which is essentially done via the connection to $\eps$-approximation specified above.
Namely, to compute a $(3/4,\eps,\alpha,\beta)$-median for the function set $F_i$ (defined in the algorithm), we take a random sample $S$ of $F_i$, find a corresponding robust median for $S$, and return it as a robust median for $F_i$.
Our main theorem in the context of bicriteria approximation follows.

\begin{theorem}\label{the:costfx_intro}
Let $F$ be a set of $n$ functions from a set $X$ to $\image$, and let $\alpha,\beta\geq 0$, $\eps \in [0,1]$. Let $\B$ be the set that is returned by the algorithm $\bicriteria(F, \eps/100, \alpha,\beta)$; see Fig.~\ref{fig:alg_bi_intro}.
Then $B$ is a $((1+\eps)\alpha,\beta\log n)$-approximation for $F$. That is, $|B|\leq \beta\log_2 n$ and
$
\sum_{f\in F}\min_{x\in B}f(x)\leq (1+\eps)\alpha\cdot \min_{x\in X}\cost(F, x).
$
This takes time $$\bicriteriatime = O(n\ftime+\log^2 n\cdot\mbox{\bf RobustMedian}+\mbox{\bf ExhaustiveBicriteria}),$$ where:\vspace{-2mm}
\begin{itemize}
\item $\ftime$ is an upper bound on the time it takes to compute $f(Y)$ for a pair $f\in F$ and $Y\subseteq X$ such that $|Y|\leq \beta$.\vspace{-2mm}
\item $O(\mbox{\bf RobustMedian})$ is the time it takes to compute a $(3/4,\eps,\alpha,\beta)$-median for a set $F'\subseteq F$.\vspace{-2mm}
\item $O(\mbox{\bf ExahstiveBicriteria})$ is the time it takes to compute an  $(\alpha,\beta)$ bicriteria for a set $F'\subseteq F$ of size $|F'|=O(1/\eps)$.
\end{itemize}
\end{theorem}
\looseness -1

The size and running time are specified in Theorem~\ref{the:costfx_intro} in an abstract manner as a function of $\alpha$, $\beta$, $\eps$, {\bf RobustMedian}, {\bf ExhaustiveBicriteria},
and implicitly $d$ - the generalized VC dimension of the function space $(F,\XX)$.
In Section~\ref{sec:app}, we presented some concrete examples in which the size and running time specified in Theorem~\ref{the:costfx_intro} are computed for specific well studied clustering problems.
More examples appear in the appendix of this work.
%full version of this work \cite{FL11full}.
As we show, our framework improves upon previously best known results.

\subsection{From bicriteria to coresets}
Once one has established an $(\alpha,\beta)$ bicriteria approximation for the clustering problem at hand, we present a paradigm for obtaining coresets (both strong and weak as defined in Section~\ref{sec:app}).

We start the description of our results in the special case that the function set $F$ corresponds to the classical $k$-median problem in $\REAL^d$.
We then turn to present our framework when the function set $F$ corresponds to the problem of clustering points onto $k$ lines in $\REAL^d$ (i.e., {\em projective clustering}).
Finally we present our framework in its most abstract form, addressing general function families $F$.
The algorithms presented in the case study above (presented in Figures~\ref{fig:kmedian_intro} and \ref{fig:alg_intro}) are all derived from the general algorithm presented in Figure~\ref{fig:algA_intro}.

%points in a metric space $(\MM,\dist)$, and each element in the set $X$ consists of $k$ points from $M$. We refer to this problem as the metric $k$-median problem.
%The metric $k$-median problem clearly corresponds to the classical $k$-median problem in $\REAL^d$.
%%As we show shortly, for metric $k$-median our results closely resemble recent works on the study of coresets ({\em e.g.}, \cite{}).
%We then present our framework when the function set $F$ corresponds to points in a metric space $(\MM,\dist)$, and each element in the set $X$ consists of $k$ {\em subsets} of $M$ (instead of $k$ points in $M$). We refer to this problem as the metric $k$-{\em set} median problem. The metric ``$k$-set median'' problem, generalizes the previous ``metric $k$-median'' problem and corresponds for example to projective clustering tasks in which we cluster a set of points in $\REAL^d$ by $k$ lines or $k$ subspaces.
%Finally we present our framework in is most abstract form, addressing general function families $F$.
\medskip
\noindent
{\bf The $k$-median problem in $\REAL^d$:}
Let $P$ be a set of data elements in $\REAL^d$.
Let the centers $X$ consist of all $k$-tuples of $\REAL^d$.
(In this context, there is a function $f_p\in F$ corresponding to each point $p \in P$ defined as $f_p(x)=\dist(p,x)$.)
Our coreset construction in this case is very simple in nature and consist of two major steps.
In the first step, using a bicriteria approximation $B$, we assign a {\em weight} $m_p$ to each data element $p \in P$.
We then iteratively sample the point set $P$ according to the distribution implied by the weights $\{m_p\}$, to obtain a {\em small} sample $S \subset P$.
Our algorithm \rkmedian\ is presented in Figure~\ref{fig:kmedian_intro}.
%We note that this general algorithmic paradigm has been used successfully in the past, e.g. \cite{KeChen06,FMSW08,LS10}}.

\begin{figure}[!t]
\centering
\subfigure{\fbox{\begin{minipage}{15cm}
\begin{tabbing}
\noindent\textbf{Algorithm} {\rkmedian$(P,B,t,\eps)$}
\end{tabbing}
%\vspace{-0.8cm}
\begin{codebox}
\li \For each $b\in B$ \Do
\li     $P_b\gets$ the set of points in $P$ whose closest point in $B$ is $b$.
%\hspace{12mm}
Ties are broken arbitrarily.
\End
\li \For each $b\in B$ and $p\in P_b$ \Do \hspace{3cm}
\zi \hspace{1cm}$\displaystyle
m_{p}\gets \left\lceil \frac{|P|\dist(p,B)}{\cost(P,B)}\right\rceil+1.
$
\End
\li Pick a non-uniform random sample $\T$ of $t$
points from $P$,
where the probability \zi that a point in $\T$ equals $p\in P$, is $m_p/\sum_{q\in P}m_q$.
%where for every $q\in \T$ and $p\in P$, \zi we have $q=p$ with
%probability $m_p/\sum_{q\in P}m_q$.
\li     \For each $p\in \T$ \Do
\zi \hspace{1cm}$\displaystyle
w(p)\gets \frac{\sum_q m_q}{|\T|\cdot m_p}.
$
\End
\li     \For each $b\in B$ \Do
\li \hspace{1cm}$\displaystyle
w(b)\gets (1+10\eps)|P_b|-\sum_{p\in \T\cap P_b}w(p).
$
\End
\li $\DD\gets \T\cup B$
\li \Return $(\DD,\T,w)$
\end{codebox}
\end{minipage}}}\caption{The algorithm~{\rkmedian}.\label{fig:kmedian_intro}}
\end{figure}

This general algorithmic paradigm in itself is the basis of several coreset constructions that have been recently suggested, e.g.,  \cite{KeChen06,FMSW08,weak,LS10}.
However, the main novelty in our algorithm is in its second step, which essentially adds the bicriteria centers as additional elements in the coreset.
Adding the bicriteria  centers to the coreset, combined with a delicate weighting mechanism (that may assign negative weights), enables the proof of the following theorem.
In what follows, we assume $B$ is an $(O(1),O(k))$ bicriteria approximation. This can be obtained from previous works (e.g., \cite{KeChen06}) or by the use of our framework in an enhanced version of Theorem~\ref{the:costfx_intro} (details appear in the appendix).
%full version \cite{FL11full}).

\begin{theorem}
\label{the:kmedian1}
Let $P$ be a set of $n$ points in $\REAL^d$. Let $k\geq 1$ be an integer, $0<\eps,\delta<1/2$, and $t=\frac{c}{\eps^2}\cdot \big(dk+\log(1/\delta)\big)$,
where $c$ is a sufficiently large constant.
Then, with probability at least $1-\delta$, $\rkmedian(P,B,t,\eps)$ returns a weighted $\eps$-coreset $\DD\subseteq P$ of size $t$.
The running time needed to compute $\DD$ is $O(ndk+\log^2(1/\delta)\log^2 n+k^2+\t\log n)$.
\end{theorem}

\noindent
Replacing $\REAL^d$ by any metric space $(\MM,\dist)$ we obtain an analogous theorem in which the dimension $d$ of the corresponding function space (which effects the sample size $t$ in the theorem) is now $\log(n)$.

\begin{theorem}%\label{metrickmed}
\label{the:kmedian2}
Let $(P,\dist)$ be a metric space of $n$ points. Let $0<\eps,\delta<1/2$, and $t=\frac{c}{\eps^2}\cdot \big(k\log n+\log(1/\delta)\big),$
where $c$ is a sufficiently large constant.
Then, with probability at least $1-\delta$, $\rkmedian(P,B,t,\eps)$ returns a weighted $\eps$-coreset $\DD\subseteq P$ of size $t$.
The running time needed to compute $\DD$ is $O(nk + \log^2(1/\delta)\log^2 n+ k^2+ \t\log n)$.
\end{theorem}

The main idea governing the proofs of Theorems~\ref{the:kmedian1} and \ref{the:kmedian2} lies in the fact the the random sample $\T$ of algorithm \rkmedian\ is an $\eps$-approximation to (a slightly modified version of) the function family $F$ corresponding to $k$-median clustering of $P$.
To obtain our succinct setting for $t$, we perform a delicate analysis which determines the weights $\{m_p\}$, $\{w(p)\}$ and $\{w(b)\}$ specified in \rkmedian.
In the case of $k$-median clustering, our coresets consist of points in the data set $P$ (as common in the study of coresets for approximate clustering).
In the coresets to come, this will no longer be the case, and the functional representation of our data will be central.

\medskip
\noindent
{\bf Clustering onto $k$-lines:}
We now turn to address the more complicated case of clustering onto $k$ lines.
Namely, let $P$ be a set of data elements in $\REAL^d$.
Let the centers $X$ consist of all $k$-tuples $x$ of {\em lines} in $\REAL^d$.
%(In this context there is a function $f_p\in F$ corresponding to each point $p \in P$ defined as $f_p(x)=\dist(p,x)$.)
As in the $k$-median problem, our starting point is a bicriteria approximation $B$.
However, in this case, our algorithm will have three steps instated of two.
The first two steps are similar in nature to those of algorithm \rkmedian, however instead of returning a {\em standard} coreset, they will yield a so-called $B$-coreset (for {\bf B}icriteria) --- to be discussed in detail shortly.
Once a $B$-coreset is obtained, we take advantage of its structure to obtain a standard coreset.

We start by discussing the first two steps outlined in algorithm \dimred\ of Figure~\ref{fig:alg_intro}.
As before, our coreset $D$ is the union of two groups of points in $\REAL^d$: the subset $\T$ which is obtained by a (non-uniform) random sampling; and a second subset which is obtained via the bicriteria solution $B$.
However, in this case, the second group cannot consist of the $(\alpha,\beta)$ bicriteria $B$ itself as it is no longer a succinct set of points --- but rather a set of lines!
Thus, to proceed we {\em project} the points $P$ onto the bicriteria solution to obtain a new subset of points $P'$ of size identical to $|P|$.
Namely, for each point $p \in P$ we define a new point $p'$ on the closest line in $B$ to $p$ such that $\dist(p,B)=\|p-p'\|$.

Our $B$-coreset $D$ is now {\em in essence} the union of the sample $\T$ and the set $P'$ denoted by $\proj(P,B)$ and acts as a coreset to $P$.
To be more precise, the coreset $D$ is a function family which is a weighted and ``threshold'' defined version of $\dist(p,x)$ for points $p$ in $\T \cup P'$.
For a point $p \in \T$ and a center $x \in X$, the corresponding function in $D$ is proportional to $\dist(p,x)$ when $p'=\proj(p,B)$ is {\em close} to $x$ and zero otherwise (via the weight function $w(p,x)$).
In a complementary manner, for a point $p' \in P'$ and a center $x \in X$, the corresponding function in $D$ equals $\dist(p',x)$ when $p'$ is {\em far} from $x$ and zero otherwise (via the weight function $w(p',x)$).
Roughly speaking, the combination of functions corresponding to $\T$ and $P'$ in our coreset allows to prove the quality of $D$ using a case analysis that depends on the query point $x \in X$.
Namely, for some centers $x$ we will assign the cost of $\dist(p,x)$ to the function in $D$ corresponding to $p'$ and for others to the functions corresponding to $\T$.
This freedom will allow us to prove that indeed the cost of clustering $D$ is a good approximation to that of clustering $P$.

However, as the reader may have noticed, the size of our coreset is {\em larger} than the set we started with, so where is the gain?
The gain is in the structure of the coreset $D$ compared to the data set $P$: it is (essentially) the union of a small set $\T$ with a set $P'$ that lies in a low dimensional space.
Specifically, $P'$ can be partitioned to sets, each consisting of points on a single line (from $B$).
Thus, if $B$ is small (and using Theorem~\ref{the:costfx_intro} it is logarithmic), we have conceptually reduced the problem of finding a coreset for $P$ to that of finding a coreset for $D$, which can now be done via its specialized structure (e.g., via \cite{FFS06}).
The following theorem summarizes the quality of the resulting algorithm, which (a) first runs \dimred\ to obtain $D$ corresponding to $\T$ and $P'$, (b) then uses \cite{FFS06} and a few additional ideas to find a small set of points $\T'$ that are a good approximation to $P'$ (including a corresponding weight function), and (c) returns a succinct function set corresponding to $\T$ and $\T'$.

\begin{figure}[!t]
\centering
\subfigure{\fbox{\begin{minipage}{15cm}
\begin{tabbing}
\noindent\textbf{Algorithm} {\dimred}$(P, B,\t,\eps)$
\end{tabbing}
%\vspace{-2.5em}
\begin{codebox}
\li \For each $p\in P$\Do
\zi \label{mpdef_intro}\hspace{0.5cm}$\displaystyle
m_{p}\gets \left\lceil \frac{|P|\dist(p,B)}{\cost(P,B)}\right\rceil+1.
$\\[-0.2cm]
\End
\li Pick a non-uniform random sample $\T$ of $\t$ \label{scons2}
points from $P$, where for every $q\in \T$ and $p\in P$, \zi we have $q=p$ with
probability $m_p/\sum_{z\in P}m_z$.
\li For $p\in P$, let $p'= \proj(p,B)$.
\li \For every $p\in \T$ and set $x$ of points, define \Do
\zi \hspace{0.5cm}$\displaystyle
w(p,x)=\begin{cases} \label{w1_intro}
\frac{\sum_{z\in P}m_z}{m_p\cdot |\T|} & \dist(p',x)\leq \frac{\dist(p,B)}{\eps}\\
0 & \text{otherwise}.
\end{cases}\enspace
$\label{lineTdef_intro}
\End
\li     \For every $p\in P$ and a set $x$ of points, define
\Do
%\zi \hspace{3cm}$p'= \proj(p,B)$
\zi \hspace{0.5cm}$\displaystyle \label{w2_intro}
w(p',x)=\begin{cases}
0 & \dist(p',x)\le \frac{\dist(p,B)}{\eps}\\
1 & \text{otherwise}.
\end{cases}\enspace
$
\End
\li $D\gets \T\cup \proj(P,B)$
\li \Return $(D,\T,w)$
\end{codebox}
\end{minipage}}}\caption{The algorithm~{\dimred}.\label{fig:alg_intro}}
\end{figure}

\begin{figure}
\centering
\subfigure{\fbox{\begin{minipage}{15cm}
\begin{tabbing}
\noindent\textbf{Algorithm} {\coresetA}$(F,F',s,m,\eps)$
\end{tabbing}
%\vspace{-3mm}
\begin{codebox}
\li For each $f\in F$, let $t_f:X\rightarrow [0,\infty)$ be defined as:
\hspace{0.5cm}
$\displaystyle t_f(x)=\begin{cases}
f'(x) & f'(x)> s_f(x)\\
0& \text{otherwise}
\end{cases}\enspace
$
\li Let $T =\{t_f \mid f\in F\}$. \label{TTdef}
%\li For each $f\in F$ let $\displaystyle n_f= \left\lceil |F|\cdot\max_{\br{x\in X: f(x)\leq s(f,x)}} \frac{f(x)}{\cost(F,x)} \right\rceil$
\li For each $f\in F$ let $g_f:X\rightarrow [0,\infty)$ be defined as:
\hspace{0.5cm}
$
g_f(x)=\begin{cases}
0& f'(x)> s_f(x)\\
\frac{f(x)}{m_f} & \text{otherwise}
\end{cases}\enspace
$
\label{def:G_f_intro}
\li Let $G_f$ consist of the $m_f$ copies of $g_f$.
\li $G \leftarrow \bigcup_{f\in F}G_f$.\label{Gdef_intro}
\li $S \leftarrow$ An $\eps$-approximation of $G$.\label{Sdef}
\End
\li $U \leftarrow \br{g_f\cdot \frac{|G|}{|S|}\quad\Big|\, g_f\in S}$.
\label{def:S_intro}
%\vspace{-40mm}
\li \Return $D \leftarrow T\cup U$.
\end{codebox}
\end{minipage}}}\caption{The algorithm~{\coresetA}.\label{fig:algA_intro}}
\end{figure}

\begin{theorem}
Let $P\subseteq \REAL^d$, $k\geq 1$, $0<\eps,\delta \leq 1/2$,  $r=k+\log(1/\delta)$ and
$t\geq \frac{c}{\eps^2} \left(dk+\log\frac{1}{\delta}\right)$,
for a sufficiently large constant $c$. A set $D$ of $O(\t)+((1/\eps)\log n)^{O(k)}$ points and a weight function $w:D \times X \rightarrow [0,\infty)$ can be computed in
$O(ndk + dt^2)+ t^{O(k)}\log^2 n$ time, such that, with probability at least $1-\delta$, for every set $x$ of $k$ lines in $\REAL^d$,
\[
\left|\sum_{p\in P}\dist(p,x)-\sum_{p\in D}w(p,x)\dist(p,x)\right|\leq \eps \sum_{p\in P}\dist(p,x).
\]
\end{theorem}

\medskip
\noindent
{\bf The general setting:}
We now address the general setting in which we are given a general function family $F$.
As in the previous case, our algorithm first finds a $B$-coreset, and only then may try to utilize the nature of the $B$-coreset to obtain a standard coreset.
Our algorithm \coresetA\ for finding the $B$-coreset is presented in Figure~\ref{fig:algA_intro} and is phrased in an abstract manner that captures the previously defined coreset algorithms \dimred\ and \rkmedian.

Roughly speaking, as before, our $B$-coreset will consist of two subsets of functions, the subset $T$ which is defined by the ``projection'' of $F$ onto a given bicriteria $B$; and the function set $U$ which is a weighted random sample of the function set $F$.
However, for a general function set $F$, there is no natural notion of projection.
To address this difficulty, we {\em define} the projection of $F$ onto a bicriteria solution $B$, as an additional function set $F'$ given as input to \coresetA.
In our analysis, we will rely on certain properties of $F'$ that intuitively correspond to the standard notion of projection that arises in various applications.
Additional inputs to algorithm \coresetA\ include a threshold function $s_f:X \rightarrow [0,\infty)$ for every $f \in F$, and a weight function $m: F \rightarrow \mathbb{N}\setminus \br{0}$.
These will play the role of the threshold and weight functions defined in the previous algorithm \dimred.

We now turn to discuss the set $U$ returned as output by \coresetA.
Notice, that there is no use of random sampling in algorithm \coresetA.
Instead, to construct the set $U$ we use the more general notion of $\eps$-approximation, again on a weighted and threshold defined variant of $F$.
To be precise, we could have used the notion of $\eps$-approximation in the previously defined coreset algorithms as well, but instead represented them in terms of random sampling for ease of presentation.

All in all, algorithm \coresetA\ returns two sets, the function set $T$ that corresponds to a threshold version of $F'$ (which intuitively corresponds to a projected version of $F$ onto a given bicriteria solution), and the function set $U$ which corresponds to a small sized $\eps$-approximation to (a threshold and weighted version) of the family $F$.
Our main theorem in the this general setting is now:

\begin{theorem}
\label{the:coresetA_intro}
Let $F$ be a set of functions from $X$ to $[0,\infty]$, and $0<\eps<1/4$. Let $s:(F,X)\rightarrow [0,\infty)$, and $m:F\rightarrow \mathbb{N}\setminus \br{0}$.
For every $x\in X$, let $M(x)=\br{f\in F: f'(x)\leq s_f(x)}$.
For each $f \in F$ let $f'$ be a corresponding function associated with $f$, and let $F' = \{f' | f \in F\}$.
Then for $D = \coresetA(F,F',s,m,\eps)$ it holds that
{\small{
\begin{equation*}
%\label{toprove10}
\begin{split}
\forall x\in X:
&|\cost(F, x)-\cost(D, x)|\leq \\
& \sum_{f\in F\setminus M(x)}\big|f(x)-f'(x)\big|+\eps\max_{f\in M(x)}\frac{s_f(x)}{m_f}\sum_{f\in F}m_f.
\end{split}
\end{equation*}
}}
\end{theorem}

{\bf Some remarks} are in place.
Primarily, our presentation of Theorem~\ref{the:coresetA_intro} is very general and involves several parameters and function sets.
From this presentation, both the the size and quality of our coreset $D$ is hard to decipher.
The abstract nature of Theorem~\ref{the:coresetA_intro} allows us to apply it on several function families $F$.
In Section~\ref{sec:app} we have presented a number of concrete algorithmic applications.
These applications are proven in detail in the appendix.
%full version of this work \cite{FL11full}.

Secondly, as discussed in Section~\ref{sec:tech_intro}, the output of algorithm \coresetA\ is a new set of functions $D$ that may not be a subset of $F$.
Indeed, this is the case, however we stress that the set $U$ is essentially a subset of $F$ which differs only by our weights $m_f$ and threshold cut-off $s_f$.
Moreover, the function set $F'$ and thus the set $T$ will be a set of functions that are typically easy to compute from a bicriteria of $(F,X)$.
As we have shown, in certain cases, such as the $k$-median problem discussed previously, we are able to slightly modify our algorithm so that it returns a set of points $D \subset F$ as the desired coreset and not a function set that may have cut-off thresholds.

\section{Acknowledgment}
We wish to thank Christos Boutsidis, Michael Mahoney and Leonard Schulman for helpful discussions on this paper.

\bibliographystyle{alpha}
\bibliography{mybib}

\newpage
\appendix
\noindent
{\bf \Large{Appendix}}
\def\thesection       {\arabic{section}}
\setcounter{section}{4}

\section{Road map}
The body of this extended abstract holds a detailed discussion of our results, without elaborating on the rigorous technical content. In this self contained appendix, we present the complete definitions and proofs of all our claims discussed in the body of this work. The appendix is organized as follows.
\begin{itemize}
\item In Section~\ref{sec:a_pre}, we review the notion of $\eps$ approximation for range spaces and define and analyze the new notion of $\eps$-approximations for function families.
\item In Section~\ref{sec:a_hd}, we define and analyze the notion of generalized range spaces and generalized dimension, including the connection between these notions and the classical notions of Section~\ref{sec:a_pre}.
\item In Section~\ref{sec:step2}, we show a connection between $\eps$-approximations and a new relaxed notion of coresets we refer to as robust coresets.
\item In Section~\ref{sec:robustapprox}, we further study the notion of robust coresets and link them with the notion of a robust median discussed in the body of the paper. This connection ties the notion of robust medians with that of $\eps$-approximations.
\item In Section~\ref{sec:centaa} we define the notion of a {\em centroid set} to be used in the sections to come.
\item In Section~\ref{sec:b} we tie the notion of robust coresets with that of bi-criteria approximation, a connection discussed in the body of this work.
\item In Section~\ref{sec:a_app1}, we use the analysis of previous sections to obtain concrete results on the bicriteria approximation of several clustering problems, some of which were discusses in Section~\ref{sec:app} in the body of the paper.
\item In Section~\ref{sec:app_bcoreset} we use our bi-criteria approximation to obtain algorithms for $B$-coresets (specified in the body of this work). In Section~\ref{sec:metric_b_coresets} we study the special case in which our functions $F$ correspond to points in a metric space, in Section~\ref{sec:a:kmed} we focus on the $k$-median problem in metric spaces, and finally in Section~\ref{sec:kminrd} we study the $k$-median problem in $\REAL^d$. Many of the concrete results stated in Section~\ref{sec:app} are proven in detail in these sections.
\item In Section~\ref{sec:kline} we study the $k$-line median problem, and prove the results stated in Section~\ref{sec:app}.
\item In Section~\ref{sec:weak}, we show how to apply our framework in
order to construct (low-dimensional)  $B$-coresets and coresets for
subspace approximation. We apologize to the reader, and note that we
are currently still writing parts of this section, which will be
uploaded to a future version on arXiv.
%\item In Section~\ref{sec:weak}, we show how to apply our framework in order to construct (low-dimensional)  %$B$-coresets and coresets for subspace approximation.
%\item \mw{Add section on streaming if it will remain}.
\end{itemize}

\section{$\eps$-Approximations}
\label{sec:a_pre}

In this section we will discuss the basic definitions of $\eps$-approximation used throughout this work.

\begin{definition}[range space.]
\label{def:d}
A \emph{range space} is a pair $(F, \ranges)$ where $F$ is a set, and $\ranges$ is a set of subsets of $F$. The \emph{dimension} of the range space $(F, \ranges)$ is the smallest integer $d$, such that for every $G\subseteq F$ we have \[\Big|\br{G\cap\range \mid \range\in\ranges}\Big|\leq |G|^{d}\enspace.\]
\end{definition}
The dimension of a range space relates (but is not equivalent) to a term known as the VC-dimension of a range space.

\begin{definition}[$\eps$-\sample~of a range space.]\label{shatter}
\medskip
A set $S$ of functions is an $\eps$-{\sample} of the range space $(F,\ranges)$,
if for every $\range\in\ranges$ we have
\begin{equation*}\label{enet}
\left| \frac{|\range|}{|F|}-
\frac{|S\cap \range|}{|S|}\right|
\leq \eps.
\end{equation*}
\end{definition}
Usually $S\subseteq F$, otherwise $S$ is called in the literature a \emph{weak} $\eps$-\sample .

The following well known theorem states that a random sampling from a set is also an $\eps$-{\sample} of $F$. See discussion in~\cite{HP09}.
\begin{theorem}[\cite{li00improved, Vap71a}]
\label{enettheorem}
Let $(F,\ranges)$ be a range space of dimension $d$. Let $\eps,\delta>0$. Let $S$ be a sample of
\begin{equation*}%\label{sdef}
|S|=\frac{c}{\eps^2}\left(d+\log\frac{1}{\delta}\right)
\end{equation*}
i.i.d items from $F$, where $c$ is a sufficiently large constant. Then, with probability at least $1-\delta$, $S$ is an $\eps$-approximation of $(F,\ranges)$.
\end{theorem}

\begin{definition}[range space and dimension of $F$.~\cite{li00improved}]\label{dimfunctions}
Let $F$ be a finite set of functions from a set $X$ to $\image$.
The \emph{dimension} $\dim(F)$ of $F$ is the dimension of the range space $\big(F,\ranges(F)\big)$, where $\ranges(F)$ is the \emph{range space of $F$}, that is defined as follows. For every $x\in X$ and $r\geq 0$, let $\range(F,x,r)=\br{f\in F\mid f(x)\leq r}$. Let $\ranges(F)=\br{\range(F,x, r)\mid x\in X, r\geq 0}$.
%An $\eps$-approximation for the range space $(F,\ranges(F))$ is called an $\eps$-approximation for $F$.
\end{definition}

The following lemma follows directly from our definitions:
\begin{lemma}\label{clustering}
Let $F$ be a set of functions from $X$ to $\image$, and let $k\geq 1$. For every $f\in F$ define a corresponding function $f':X^k\rightarrow \image$ such that $f'(x_1, \cdots, x_k)=\min_{1\leq i\leq k}f(x_i)$,
for every $x_1,\cdots x_k\in X$. Let $F'=\br{f'\mid f\in F}$ be the union of these functions. Then $\dim(F') \leq k\cdot \dim(F).$
\end{lemma}

\begin{definition}[cost]
Let $F$ be a set of functions from $X$ to $[0,\infty)$. Let $x\in X$. We define $\cost(F,x)=\sum_{f\in F}f(x).$
\end{definition}

We now define the notion of an $\eps$-approximation for a function set $F$ and tie it to an $\eps$-approximation of the corresponding range space.
This notion plays a central part in our work.
Roughly speaking, an $\eps$-approximation for a function set $F$ is a subset $S$ that approximates the average {\em cost} of ranges in the range space corresponding to $F$.  To allow invariance by constant multiplication, the quality of the approximation defined below is necessarily related to the parameter $r$ bounding the value of our functions in the range being considered.

\begin{definition}[$\eps$-approximation of $F$]
Let $F$ be a set of functions from $X$ to $\image$, and let $\eps \in (0,1)$. An $\eps$-{\sample} of $F$ is a set $S\subseteq F$ that satisfies
\[
\forall x\in X, r\geq0:\left|\frac{\cost(\range(x,r),x)}{|F|}
-\frac{\cost(S\cap \range(x,r),x)}{|S|}\right|
\leq \eps r,
\]
where $\range(x,r)=\br{f\in F\mid f(x)\leq r}$.
\end{definition}

%The following theorem was proved in~\cite{li00improved} for the special case where $r=1=\max_{f\in F,\, x\in X}f(x)$ and $S$ is a random sample from $F$ as defined in~\ref{enettheorem}. Our generalization will be useful in the following sections, in order to deal with outliers, and to obtain deterministic results.

We now show the connection between $\eps$-approximations for range spaces and for function families.

\begin{theorem}\label{enet2}
Let $F$ be a set of functions from $X$ to $\image$, and let $\eps \in (0,1)$. Let $S$ be an $\eps$-{\sample} of the range space of $F$. Then
$S$ is an $\eps$-approximation of $F$.
\end{theorem}
\begin{proof}
Let $x\in X$ and $r\geq 0$. For every $b\geq 0$, let $\range(b)=\range(x,b)$.
Let $\range(r)=\br{f_1,\cdots, f_n}$ denote the $n$ functions in $\range(r)$, sorted by their $f(x)$ value.
Let $a_0=a_1=0$, and $m=n/\lceil \eps n\rceil$. For every $i$, $1\leq i\leq m$, let $a_{2i}=a_{2i+1}=f_{i \lceil \eps  n \rceil}(x)$.
We define the partition $\br{F_1,\cdots, F_{2m+1}}$ of $\range(r)$, where $F_1=\br{f\in F \mid f(x)=0}$ and, for $1\leq i\leq m$,
\begin{equation}\label{Fbound}
F_{2i}=\br{f\in F\mid a_{2i-1}<f(x)<a_{2i}},
\end{equation}
\[F_{2i+1}=\begin{cases}
\br{f\in F \mid f(x)=a_{2i}} & a_{2i}\neq a_{2i-1}\\
\emptyset & a_{2i}=a_{2i-1}.
\end{cases}\enspace
\]
%
%The partition $\br{F_1,\cdots, F_{2m}}$ of $\range(r)$ has the following two properties for every $1\leq i\leq m$:
%\begin{equation}\label{prop1}
%|F_{2i-1}|\leq \eps n\leq \eps |F|,
%\end{equation}
%and, for every $f\in F_{2i}$ we have
%\begin{equation}\label{prop2}
%f(x)=a_{2i}=a_{2i-1}.
%\end{equation}

Note that $\cost(F_1,x)=0$. For every $i$, $2\leq i\leq 2m+1$, and $S_i=F_i\cap S$, we have
\begin{equation}\label{sumcost}
\begin{split}
\cost(S_i,x)&=\sum_{f\in S_i}f(x)=\sum_{f\in S_i}(f(x)-a_{i-1})+|S_i|a_{i-1}\\
&=\sum_{f\in S_i}(f(x)-a_{i-1}) +|S_i|\sum_{j=1}^{i-1}(a_{j}-a_{j-1}).
\end{split}
\end{equation}
Let $r_j=F_{j+1}\cup \cdots \cup F_{2m+1}$ for every $1\leq j\leq 2m$. Summing the last term of~\eqref{sumcost} over $2\leq i\leq 2m+1$ yields
\[
\begin{split}
\sum_{i=2}^{2m+1} \sum_{j=1}^{i-1} |S_i| (a_{j}-a_{j-1})
 &=\sum_{j=1}^{2m}\sum_{i=j+1}^{2m+1}  |S_i|(a_{j}-a_{j-1})\\
&=\sum_{j=1}^{2m}(a_{j}-a_{j-1})\sum_{i=j+1}^{2m+1}  |S_i|=\sum_{j=1}^{2m}(a_{j}-a_{j-1})|S\cap r_j|.
\end{split}
 \]
Hence, summing~\eqref{sumcost} over $2\leq i\leq 2m+1$ yields
\begin{equation}\label{costSx}
\begin{split}
\cost(S\cap \range(r),x)
&=\sum_{i=2}^{2m+1}\cost(S_i,x)\\
&=\sum_{i=2}^{2m+1}\sum_{f\in S_i}(f(x)-a_{i-1})+\sum_{j=1}^{2m}(a_{j}-a_{j-1})|S\cap r_j|.
\end{split}
\end{equation}
Similarly,
\begin{equation}\label{costFx}
\cost(\range(r),x)=\sum_{i=2}^{2m+1}\sum_{f\in F_i}(f(x)-a_{i-1})+\sum_{j=1}^{2m}(a_{j}-a_{j-1})|r_j|.
\end{equation}

By the triangle inequality,
\begin{alignat}{4}
\left|\frac{\cost(\range(r),x)}{|F|}-\frac{\cost(S\cap\range(r),x)}{|S|}\right|
\label{allTriangle}&\leq \left|\frac{\cost(\range(r),x)}{|F|}-\sum_{j=1}^{2m}\frac{(a_{j}-a_{j-1})|r_j|}{|F|}\right|\\
\label{all2}&+\left|\sum_{j=1}^{2m}(a_{j}-a_{j-1})\cdot\left(\frac{|r_j|}{|F|}-\frac{|S\cap r_j|}{|S|}\right)\right|\\
\label{all3}&+\left|\sum_{j=1}^{2m}\frac{(a_{j}-a_{j-1})|S\cap r_j|}{|S|}
-\frac{\cost(S\cap\range(r),x)}{|S|}\right|.
\end{alignat}
We now bound each term in the right hand side of the last equation.
Using~\eqref{costFx}, we have
\begin{equation}\label{888}
\begin{split}
\left|\frac{\cost(\range(r),x)}{|F|}-\sum_{j=1}^{2m}\frac{(a_{j}-a_{j-1})|r_j|}{|F|}\right|
&=\sum_{i=2}^{2m+1}\sum_{f\in F_{i}}\frac{f(x)-a_{i-1}}{|F|}\\
&\leq \sum_{i=1}^{m}(a_{2i}-a_{2i-1})\cdot\frac{|F_{2i}|}{|F|}
\leq \frac{a_{2m}\eps n }{|F|}\leq \eps a_{2m},
\end{split}
\end{equation}
which bounds ~\eqref{allTriangle}. Similarly, using~\eqref{costSx},
\begin{equation}\label{sfu}
\begin{split}
\left|\sum_{j=1}^{2m}\frac{(a_{j}-a_{j-1})|S\cap r_j|}{|S|}-\frac{\cost(S\cap \range(r),x)}{|S|}\right|
&=\sum_{i=2}^{2m+1}\sum_{f\in S_i}\frac{f(x)-a_{i-1}}{|S|}\\
&\leq \sum_{i=1}^{m}(a_{2i}-a_{2i-1})\cdot \frac{|S_{2i}|}{|S|}.
\end{split}
\end{equation}

Since $S$ is an $\eps$-approximation for $(F,\ranges(F))$, we have
\begin{equation}\label{allx}
\forall b\geq 0: \left|\frac{|\range(b)|}{|F|}-\frac{|S\cap\range(b)|}{|S|}\right|\leq \eps.
\end{equation}
Put $1\leq i\leq m$, and \[
b_{2i}=\begin{cases}
\max_{f\in F_{2i}}f(x) &  F_{2i}\neq \emptyset \\
a_{2i-1}&F_{2i}= \emptyset
\end{cases}\enspace.
\]By~\eqref{Fbound} and~\eqref{allx}, we have
\[
\begin{split}
\frac{|S_{2i}|}{|S|}&=\frac{|S\cap \range(b_{2i})|}{|S|}-\frac{|S\cap \range(a_{2i-1})|}{|S|}\\
&\leq \frac{|\range(b_{2i})|}{|F|}-\frac{|\range(a_{2i-1})|}{|F|}+2\eps\\
&= \frac{|F_{2i}|}{|F|}+2\eps\leq 3\eps.
\end{split}
\]
Combining the last inequality in~\eqref{sfu} bounds~\eqref{all3}, as
\begin{equation}\label{all4}
\begin{split}
\left|\sum_{j=1}^{2m}\frac{(a_{j}-a_{j-1})|S\cap r_j|}{|S|}-\frac{\cost(S\cap\range(r),x)}{|S|}\right|
&\leq \sum_{i=1}^m (a_{2i}-a_{2i-1})\cdot 3\eps=3\eps a_{2m}.
\end{split}
\end{equation}

Using~\eqref{allx}, expression~\eqref{all2} is bounded by
\[
\begin{split}
&\sum_{j=1}^{2m}(a_{j}-a_{j-1})\cdot\left|\frac{|r_j|}{|F|}-\frac{|S\cap r_j|}{|S|}\right|\\
&=\sum_{j=1}^{m}(a_{2j}-a_{2j-1})\cdot\left|\frac{|F|-|\range(b_{2j})|}{|F|}-\frac{|S|-|S\cap \range(b_{2j})|}{|S|}\right|\\
&\leq \sum_{j=1}^{m}(a_{2j}-a_{2j-1})\cdot \eps= \eps a_{2m}.
\end{split}
\]
Combining~\eqref{888},~\eqref{all4} and the last inequality bounds the left hand side of~\eqref{allTriangle}, as
\[
\begin{split}
\left|\frac{\cost(\range(r),x)}{|F|}-\frac{\cost(S\cap\range(r),x)}{|S|}\right|
&\leq \eps a_{2m} +\eps a_m+3\eps a_{2m}\\
&=5\eps a_{2m}\leq 5\eps r.
\end{split}
\]
\end{proof}

By plugging Theorem~\ref{enettheorem} in Theorem~\ref{enet2} we obtain the following corollary.

\begin{theorem}\label{enet4}
Let $F$ be a set of functions from $X$ to $\image$, and let $\eps \in (0,1)$. Let $S$ be a sample of
\begin{equation*}%\label{sdefz}
|S|=\frac{c}{\eps^2}\left(\dim(F)+\log\frac{1}{\delta}\right)
\end{equation*}
i.i.d items from $F$, where $c$ is a sufficiently large constant. Then, with probability at least $1-\delta$, $S$ is an $\eps$-approximation of $F$.
\end{theorem}

%*****************************************************

\section{$\eps$-Approximations for High and Infinite Dimensional Spaces}
%: proof of Theorem~\ref{thm:gen}
\label{sec:a_hd}
Suppose that we have a range space of a high (maybe infinite) dimension $d$.
In this section we show that for several natural families of high dimensional range spaces, a small $\eps$-approximation can be constructed that approximates (not all, but rather) a {\em subset} of the ranges in the range space.
This weaker type of $\eps$-approximation suffices to solve certain optimization problems in high dimensional space.
Towards this end, we will define the notion of a generalized range space, the notion of a corresponding function space, and the notion of $\eps$-approximation in this context.
As before, these notions will play a major role in our analysis.
%We restate Definitions~\ref{shatter2} and \ref{dimfunctionsHigh} from Section~\ref{???}.
\begin{definition}[generalized range space.]\label{shatter22}
Let $F$ be a set. Let $\RANGES$ be a function that maps every subset $S\subseteq F$ to a set $\RANGES(S)$ of subsets of $F$. The pair $(F, \RANGES)$ is a \emph{generalized range space} if for every two sets $S, G$ such that $S\subseteq G\subseteq F$, we have $\RANGES(S)\subseteq\RANGES(G)$.
The \emph{dimension} of a generalized range space $(F, \RANGES)$ is the smallest integer $d$, such that \[\forall S\subseteq F:\Big|\br{S\cap\range \mid \range\in\RANGES(S)}\Big|\leq |S|^{d}\enspace.\]
\end{definition}

We now define the generalized dimension of a family of functions:

\begin{definition}[function space.]\label{dimfunctionsHigh2}
Let $F$ be a finite set of functions from a set $X$ to $\image$. Let $\XX$ be a function that maps every subset $S\subseteq F$ to a set of items $\XX(S)\subseteq X$. The pair $(F, \XX)$ is called a \emph{function space}, if the pair $(F, \RANGES)$ is a generalized range space, where $\RANGES$ is defined as follows. For every $x\in X$ and $r\geq 0$, let $\range(x,r)=\br{f\in F\mid f(x)\leq r}$. For every $S\subseteq F$, let $\RANGES(S)=\br{\range(x, r)\mid x\in \XX(S), r\geq 0}$.
The \emph{dimension $\dim(F, \XX)$ of the function space} $(F,\XX)$ is the dimension of the generalized range space $(F, \RANGES)$.
\end{definition}

%For example, suppose that $F$ represents a set $P$ of points. That is, every $f_p\in F$ is a distance function $f_p(x)$ from a point $p$ to a center $x\in\REAL^d$. Let \[\XX(F)=\br{\sum_{p\in Q}p/|Q| \mid Q\subseteq P, |Q|=100}\subseteq\REAL^d\] denote the union over every center of mass $\sum_{p\in Q}p/|Q|$ of a subset $Q$ of $P$, of size \mr{should be removed: at least} \mb{$|Q| = 100$}. More generally, for every $S\subset F$ (that corresponds to a subset of $P$), let $\XX(S)=\br{\sum_{p\in Q}p/|Q| \mid Q\subseteq S, |Q|=100}$. Then the pair $\XX(\cdot)$ is \mr{text not completed} ..
\mw{add example}
We note that it is not hard to verify that for $\XX\equiv X$ it holds that $\dim(F,X)=\dim(F,\XX)$.
For a subset $S$ of $F$, let $F_{\ind \XX(S)}:\ \XX(S) \rightarrow [0,\infty)$ be the function set which is defined by {\em restricting} the functions $F$ to inputs in $\XX(S)$.
The following theorem is an immediate consequence of the proof in \cite{li00improved} and can be seen as a corollary of Theorem~\ref{enettheorem}.

\begin{theorem}[$\eps$-approximation for a function space]\label{mycor2}
Let $(F,\XX)$ be a function space of dimension $d$ from $X$ to $[0,\infty)$. Let $\eps,\delta>0$. Let $S$ be a sample of
\begin{equation*}%\label{sdef}
|S|=\frac{c}{\eps^2}\left(d+\log\frac{1}{\delta}\right)
\end{equation*}
i.i.d functions from $F$, where $c$ is a sufficiently large constant. Then, with probability at least $1-\delta$, $S$ is an $\eps$-approximation of the range space $(F,\RANGES(S))$.
%$F_{\ind \XX(S)}$.
\end{theorem}

The following is a simple corollary of Theorem~\ref{enet2} that connects between the notion of $\eps$-approximation for range spaces and  $\eps$-approximation for function sets in the generalized setting.

\begin{corollary}\label{lemma:mycor2}
Let $(F,\XX)$ be a function space of dimension $d$. Let $S$ be an $\eps$-approximation of the range space $(F,\RANGES(S))$ for some $\eps>0$.
Then $S$ is an $\eps$-approximation of $F_{\ind \XX(S)}$.
\end{corollary}

%\mikel{Need to add proof - current proof is incorrect}

%\begin{proof}
%For every $x\in \XX(S)$ and $r\geq 0$, let \[
%\range(F_{\ind\XX(S)},x,r)=\br{f\in F_{\ind\XX(S)}\mid f(x)\leq r},
% \]
%and \[
%\ranges=\br{\range(F_{\ind\XX(S)},x, r)\mid x\in \XX(S), r\geq 0}\enspace.
% \]
%Similarly, let
%\[
%\range(F,x,r)=\br{f\in F\mid f(x)\leq r},
%\]
%and
%\[
%\RANGES(S)=\br{\range(F,x,r)|x\in \XX(S), r\geq 0}\enspace.
%\]
%Hence,
%\begin{equation}\label{rangesRANGES}
%\begin{split}
%\RANGES(S)
%&=\br{\range(F,x,r)|x\in \XX(S), r\geq 0}\\
%&=\br{\range(F_{\ind\XX(S)},x,r)|x\in \XX(S), r\geq 0}\\
%&=\ranges
%\end{split}
%\end{equation}
%
%By the assumption of this lemma, $S$ is an $\eps$-approximation of the range space $(F,\RANGES(S))$. For every $\range\in\RANGES(S)$ we have $f\in\range$ if and only if $f_{\ind\RANGES(S)}\in\range$. Hence, $S$ is also an $(\eps^2/25)$-approximation of the range space $(F_{\ind\XX(S)},\RANGES(S))$.
%By this and ~\eqref{rangesRANGES}, we have that $S$ an $(\eps^2/25)$-approximation of the range space $(F_{\ind\XX(S)},\ranges)$.
%By substituting $F=F_{\ind\XX(S)}$ and $X=\XX(S)$ in Definition~\ref{dimfunctionsHigh}, a set is an $\eps$-approximation of $F_{\ind\XX(S)}$ if it is an $(\eps^2/25)$-approximation of the range space $(F_{\ind\XX(S)}, \ranges)$. Hence, $S$ is an $\eps$-approximation of $F_{\ind\XX(S)}$.
%\end{proof}

%\subsection{Proof of Theorem~\ref{thm:gen}}
Using Corollary~\ref{lemma:mycor2} with Theorem~\ref{mycor2}, we now conclude:
\begin{theorem}\label{functionspacerand}
Let $(F,\XX)$ be a function space of dimension $d$. Let $0<\eps,\delta<1$, and let $S$ be a random sample of at least $$\frac{c}{\eps^2}\left(d+\log\frac{1}{\delta}\right)$$ i.i.d functions from $F$, where $c$ is a sufficiently large constant.
Then, with probability at least $1-\delta$, $S$ is an $\eps$-approximation of $F_{\ind\XX(S)}$.
\end{theorem}

\section{From $\eps$-approximations to
$(\gamma,\eps)$-coresets}

\label{sec:step2}

In this section we define and analyze the notion of $(\gamma,\eps)$-coresets: a relaxed notion of coresets (that we refer to as {\em robust} coresets) that we will use in our study of {\em robust medians} discussed in the Introduction. Roughly speaking, we show that $\eps$-approximators for $F$ are also $(\gamma,\eps)$-coresets.

%\mikel{Do we want this section to use the notion of a generalized function space as well?}

\begin{definition}[$(\gamma,\eps)$-coreset.]\label{def:three_levels_of_gamma}
Let $\eps\in (0,1/2)$, and $\gamma\in (0,1]$. Let $F$ and $S$ be two sets of functions from a set $X$ to $\image$. For every $x\in X$:
\begin{itemize}
\item Let $F_x$ denote the $\big\lceil \gamma|F|\big\rceil$ functions $f\in F$ with the smallest value $f(x)$
\item Let $S_x$ denote the $\big\lceil (1-\eps)\gamma|S|\big\rceil$ functions $f\in S$ with the smallest value $f(x)$
\item Let $G_x\subseteq F_x$ denote the $\big\lceil(1-2\eps)\gamma|F|\big\rceil$ functions $f\in F$ with the smallest value $f(x)$
\end{itemize}
The set $S$ is $(\gamma,\eps)$-good for $F$ if
\begin{equation}
\label{myprove15}
\forall x\in X:
(1-\eps)\cdot\frac{\cost(G_x, x)}{|G_x|}\leq \frac{\cost(S_x, x)}{|S_x|}\leq \frac{\cost(F_x, x)}{|F_x|}\cdot (1+\eps)\enspace.
\end{equation}
The set $S$ is a $(\gamma,\eps)$-coreset of $F$ if for every $\gamma'\in[\gamma,1]$, and $\eps'\in[\eps,1/2)$, we have that $S$ is $(\gamma',\eps')$-good for $F$.
\end{definition}

Our definition of robust coresets has the flavor of approximating with {\em outliers}.
Namely, in our definition, we allow a portion of the functions in both $F$ and $S$ to be neglected when considering the quality of $S$.
In what follows, we show that an $\eps$-approximation $S$ to a function set $F$ is also a robust coreset.
 %We start by ...
%\mikel{Add text}

\begin{theorem}\label{smaller}
Let $\eps\in(0,1)$. Let $F$ be a set of functions from $X$ to $\image$, and let $S$ be an $(\eps/7)$-{\sample} of the range space corresponding to $F$. Suppose that $|F|,|S|\geq 5/\eps$. Let $\gamma\in (0,1]$, and for every $x\in X$:
\begin{itemize}
\item Let $F_x$ denote the $\lceil \gamma\cdot |F|\rceil$ functions $f\in F$ with the smallest value $f(x)$
\item Let $S_x$ denote the $\lceil \gamma\cdot |S|\rceil$ functions $f\in S$ with the smallest value $f(x)$
\end{itemize}
Then
\begin{equation*}
%\label{toprove9}
\forall x\in X:
\left|\frac{\cost(F_x, x)}{|F|}
-\frac{\cost(S_x, x)}{|S|}\right|
\leq \eps\cdot\max_{f\in F_x\cup S_x}f(x)\enspace.
\end{equation*}
\end{theorem}

\begin{proof}
Let $\eps\in(0,1/7)$, and let $S$ be an $\eps$-approximation to the range space corresponding to $F$. By Theorem~\ref{enet2}, $S$ is also an $\eps$ approximation to $F$.
Let $S_x$ denote the $\lceil \gamma\cdot |S|\rceil$ functions $f\in S$ with the smallest value $f(x)$. Let $\gamma$, $S_x$, and $F_x$ be defined as in the statement of the theorem. We will prove that
\begin{equation}
\label{toprove9}
\forall x\in X:
\left|\frac{\cost(F_x, x)}{|F|}
-\frac{\cost(S_x, x)}{|S|}\right|
\leq 7\eps\cdot\max_{f\in F_x\cup S_x}f(x)\enspace.
\end{equation}
This suffices to prove the theorem for $\eps\in(0,1)$.

Indeed, for every $x\in X$ and $r\geq 0$, we define $\range(x, r)=\br{f\in F\mid f(x)\leq r}$.
By our definitions,
\begin{equation}\label{god3}
\forall x\in X, r\geq 0:\left|\frac{\cost(\range(x,r),x)}{|F|}
-\frac{\cost(S\cap \range(x,r),x)}{|S|}\right|
\leq \eps r\enspace,
\end{equation}
and
\begin{equation}\label{god5}
\forall x\in X, r\geq 0: \left| \frac{|\range(x,r)|}{|F|}-
\frac{|S\cap \range(x,r)|}{|S|}\right|
\leq \eps.
\end{equation}

Fix $x\in X$, and let $r=\max_{f\in F_x\cup S_x}f(x)$, $\gm=\br{f\in F\mid f(x)<r}$.
\newcommand{\fx}{F_x}
\newcommand{\sx}{S_x}
We have
\[
\cost(\fx,x)=\cost(\fx\cap \gm,x)+\cost(\fx\setminus \gm,x)
=\cost(\fx\cap \gm,x)+r\cdot |\fx|-r\cdot |\fx\cap \gm|.
\]
Similarly,
\[
\cost(\sx,x)=\cost(\sx\cap \gm,x)+\cost(\sx\setminus \gm,x)
=\cost(\sx\cap \gm,x)+r\cdot |\sx|-r\cdot |\sx\cap \gm|.
\]
Let $c_1=5$. Since $|S|,|F|\geq c_1/\eps$, we have that
\begin{equation}\label{epssmall}
\left|\frac{|\fx|}{|F|}-\frac{|\sx|}{|S|}\right|
\leq \max\br{\frac{1}{|F|}, \frac{1}{|S|}}
\leq \frac{\eps}{c_1}\enspace.
\end{equation}
Using the last equations, we have
\begin{equation}\label{frc}
\begin{split}
&\left|\frac{\cost(\fx,x)}{|F|}-\frac{\cost(\sx,x)}{|S|}\right|
\\
&=\left|\frac{\cost(\fx\cap \gm,x)+r\cdot |\fx|-r\cdot |\fx\cap \gm|}{|F|}
-\frac{\cost(\sx\cap \gm,x)+r\cdot |\sx|-r\cdot |\sx\cap \gm|}{|S|}\right|\\
&\leq \left|\frac{\cost(\fx\cap \gm,x)-r\cdot |\fx\cap \gm|}{|F|}
-\frac{\cost(\sx\cap \gm,x)-r\cdot |\sx\cap \gm|}{|S|}\right|+ \left| \frac{r\cdot|F_x|}{|F|}-\frac{r\cdot|S_x|}{|S|}\right|\\
&\leq \left|\frac{\cost(\fx\cap \gm,x)}{|F|}-\frac{\cost(\sx\cap \gm,x)}{|S|}\right|
+r\cdot\left|\frac{|\fx\cap \gm|}{|F|}-\frac{|\sx\cap \gm|}{|S|}\right|+\frac{\eps r}{c_1}\enspace.
\end{split}
\end{equation}

We now bound each of the terms in the right hand side of~\eqref{frc}. Using the triangle inequality,
\[
\begin{split}
&\left|\frac{\cost(\gm\cap\fx ,x)}{|F|}-\frac{\cost( \gm\cap\sx,x)}{|S|}\right|\\
&\leq\left|\frac{\cost(\gm\cap\fx ,x)}{|F|}-\frac{\cost(\gm ,x)}{|F|}\right|
+\left|\frac{\cost(\gm,x)}{|F|}-\frac{\cost(\gm\cap S,x)}{|S|}\right|
+\left|\frac{\cost(\gm\cap S,x)}{|S|}-\frac{\cost(\gm\cap \sx,x)}{|S|}\right|\\
&=
\frac{\cost(\gm,x)}{|F|}-\frac{\cost(\gm\cap\fx,x)}{|F|}
+\left|\frac{\cost(\gm,x)}{|F|}-\frac{\cost(\gm\cap S,x)}{|S|}\right|+\frac{\cost(\gm\cap S,x)}{|S|}-\frac{\cost(\gm\cap \sx,x)}{|S|}\\
&\leq \frac{r\cdot|\gm\setminus \fx|}{|F|}+
\left|\frac{\cost(\gm,x)}{|F|}-\frac{\cost(\gm\cap S,x)}{|S|}\right|
+\frac{r\cdot|\gm\cap S\setminus \sx|}{|S|}\enspace.
\end{split}\]
Similarly,
\[
\begin{split}
\left|\frac{|\gm\cap\fx|}{|F|}-\frac{|\gm\cap \sx|}{|S|}\right|
&\leq \frac{|Y|}{|F|}-\frac{|Y\cap \fx|}{|F|}+
\left|\frac{|\gm|}{|F|}-\frac{|\gm\cap S|}{|S|}\right|
+\frac{|Y\cap S|}{|S|}-\frac{|Y\cap \sx|}{|S|}
\\
&\leq\frac{|\gm\setminus \fx|}{|F|}
+\left|\frac{|\gm|}{|F|}-\frac{|\gm\cap S|}{|S|}\right|+\frac{|\gm\cap S\setminus \sx|}{|S|}\enspace.
\end{split}\]
Combining the last two equations in~\eqref{frc} yields
\begin{equation}\label{tobound3}
\begin{split}
\left|\frac{\cost(\fx,x)}{|F|}-\frac{\cost(\sx,x)}{|S|}\right|
&\leq\left|\frac{\cost(\gm,x)}{|F|}-\frac{\cost(\gm\cap S,x)}{|S|}\right|
+r\cdot\left|\frac{|\gm|}{|F|}-\frac{|\gm\cap S|}{|S|}\right|\\
&\quad
+2r\cdot\frac{|\gm\setminus \fx|}{|F|}
+2r\cdot\frac{|\gm\cap S\setminus \sx|}{|S|}+\frac{\eps r}{c_1}\enspace.
\end{split}
\end{equation}
By
%substituting $r=\max_{f\in Y}f(x)$ in
\eqref{god3} we bound the first term in the right hand side of~\eqref{tobound3} by $\eps r$. Using~\eqref{god5} we bound the second term by $\eps$. We thus obtain
\begin{equation}\label{boundfirst}
\left|\frac{\cost(\gm,x)}{|F|}-\frac{\cost(\gm\cap S,x)}{|S|}\right|
+r\cdot\left|\frac{|\gm|}{|F|}-\frac{|\gm\cap S|}{|S|}\right|
\leq 2\eps r\enspace.
\end{equation}

We now bound the other terms in the right hand side of~\eqref{tobound3}.
By the definition of $r$ and $\gm$, we have either $\gm\subset \fx$, or $\gm\cap S\subset \sx$ (or both). Hence, $|\gm|< |\fx|$ or $|\gm\cap S|< |\sx|$.
By~\eqref{god5} we have
\begin{equation}\label{epsbig}
\left|\frac{|Y|}{|F|}-\frac{|Y\cap S|}{|S|} \right|
\leq\eps\enspace.
\end{equation}
Using the last three equations and~\eqref{epssmall}, we obtain
\[
\frac{|\gm|}{|F|}-\frac{|\fx|}{|F|}\leq \max\br{0, \frac{|\gm\cap S|}{|S|}-\frac{|\fx|}{|F|}+\eps}\leq
\max\br{0, \frac{|\sx|}{|S|}-\frac{|\fx|}{|F|}+\eps}\leq \frac{\eps}{c_1}+\eps.
\]
Since both $\fx$ and $\gm$ contain the functions with the smallest values $f(x)$, we have $|\fx\cap \gm|=\min\br{|\fx|,|\gm|}$. Together with the previous equation, we obtain
\begin{equation}\label{ymg}
\frac{|\gm\setminus \fx|}{|F|}=\frac{|\gm|-|\fx\cap \gm|}{|F|}
\leq \max\br{0, \frac{|\gm|-|\fx|}{|F|}}\leq \frac{\eps}{c_1}+\eps\enspace.
\end{equation}

Similarly, we bound the rightmost term in~\eqref{tobound3}.
As stated above, we have $|\gm|< \fx$ or $|\gm\cap S|< |\sx|$. Using~\eqref{epsbig} with the last two inequations yields
\[
\frac{|\gm\cap S|-|\sx|}{|S|}\leq \max\br{0, \frac{|\gm|}{|F|}+\eps-\frac{|\sx|}{|S|}}\leq
\max\br{0, \frac{|\fx|}{|F|}+\eps-\frac{|\sx|}{|S|}}\leq \frac{\eps}{c_1}+\eps\enspace,
\]
where the last derivation follows from~\eqref{epssmall}.
We have $|\gm\cap S\cap \sx|=\min\br{|\gm\cap S|,|\sx||}$. Together with the previous equation, we obtain
\begin{equation*}\label{ymg2}
\frac{|\gm\cap S\setminus \sx|}{|S|}
=\frac{|\gm\cap S|}{|S|}-\frac{|\gm\cap S\cap \sx|}{|S|}\leq \max\br{0, \frac{|\gm\cap S|-|\sx|}{|S|}}\leq
\frac{\eps}{c_1}+\eps\enspace.
\end{equation*}

Combining~\eqref{boundfirst}, ~\eqref{ymg} and the last equation in~\eqref{tobound3} proves~\eqref{toprove9} as follows.
\begin{equation*}\label{tobound}
\begin{split}
\left|\frac{\cost(\fx,x)}{|F|}-\frac{\cost(\sx,x)}{|S|}\right|
&\leq 2\eps r+2r\left(\eps+\frac{\eps}{c_1}\right)+2r\left(\eps+\frac{\eps}{c_1}\right)+\frac{\eps r}{c_1}\\
&   = 6\eps r+\frac{5\eps r}{c_1}\leq 7\eps r= 7\eps\cdot\max_{f\in F_x\cup S_x}f(x)\enspace.
\end{split}
\end{equation*}
\end{proof}

We are now ready to state the connection between $\eps$-approximations and $(\gamma,\eps)$ coresets.

%The following theorem states that an $\eps'$-{\sample} is also a $(\gamma,\eps)$-coreset. See Definition~\ref{def:three_levels_of_gamma} of such a coreset.
\begin{theorem}
\label{thm:three_levels_of_gamma}
Let $\eps\in (0,1/4)$, and $\gamma\in (0,1]$. Let $F$ be a set of functions from a set $X$ to $\image$, and let $S$ be an $(\eps^2\gamma/63)$-{\sample} of the range space corresponding to $F$ (and thus also of the function set $F$), such that $\displaystyle |S|,|F|\geq 5/(\eps^2\gamma)$. Then $S$ is a $(\gamma,\eps)$-coreset of $F$.
\end{theorem}
\begin{proof}
Let $\eps\in(0,1/12)$ and let $S$ be an $(\eps^2\gamma/7)$-{\sample} of $F$ such that $|S|\geq 5/(\eps^2\gamma)$. We will prove that $S$ is $(\gamma,3\eps)$-good for $F$; see Definition~\ref{def:three_levels_of_gamma}. By our definitions, $S$ is also an $(\eps'^2 \gamma'/7)$-approximation of $F$, for every $\gamma'\geq \gamma$ and $\eps'\geq \eps$. Hence, $S$ is $(\gamma',3\eps')$-good for every $\gamma'\geq \gamma$ and $\eps'\geq \eps$. This suffices to prove that $S$ is a $(\gamma,\eps)$-coreset by replacing $\eps$ with $\eps/3$.

Indeed, let $G_x$ be the $\big\lceil(1-6\eps)\gamma|F|\big\rceil$ functions $f\in F$ with the smallest value $f(x)$, and $S_x$ denote the $\big\lceil (1-3\eps)\gamma|S|\big\rceil$ functions $f\in S$ with the smallest value $f(x)$.
In order to prove that $S$ is $(\gamma,3\eps)$-good for $F$, we need to prove that
\begin{equation}
\label{myprove15a}
\forall x\in X:
(1-3\eps)\cdot\frac{\cost(G_x, x)}{|G_x|}\leq \frac{\cost(S_x, x)}{|S_x|}\leq \frac{\cost(F_x, x)}{|F_x|}\cdot (1+3\eps)\enspace.
\end{equation}

Fix $x\in X$, and let $H_x$ denote the $\big\lceil \gamma(1-3\eps)|F| \big\rceil$ functions $f\in F$ with the smallest value $f(x)$. We first bound the right hand side of~\eqref{myprove15a}. By Theorem~\ref{smaller}, we have
\begin{equation}\label{enet72}
\begin{split}
\frac{\cost(S_x,x)}{|S|}\leq \frac{\cost(H_x,x)}{|F|}+\eps^2 \gamma \max_{f\in H_x\cup S_x} f(x),
\end{split}
\end{equation}
Since $1\leq\eps\gamma |F|$, we have
\[
|H_x|< (1-3\eps)\gamma|F|+1
\leq (1-2\eps)\gamma|F|\leq (1-2\eps)|F_x|\enspace.
\]
By the last equation and Markov's inequality,
\begin{equation}\label{maxs2}
\max_{f\in H_x} f(x)\leq \frac{1}{2\eps}\cdot\frac{\cost(F_x,x)}{|F_x|}\enspace.
\end{equation}

Let $U=\br{f\in F\mid f(x)< \max_{f\in S_x}f(x)}$. Since $S\cap U\subset S_x$, we have
\begin{equation}\label{su}
|S\cap U|\leq (1-3\eps)\gamma|S|.
\end{equation}
Since $S$ is an $(\eps^2\gamma/7)$-approximation of $(F,\ranges(F))$, we have
\begin{equation*}\label{enet6}
\left| \frac{|U|}{|F|}-
\frac{|S\cap U}{|S|}\right|
\leq \frac{\eps^2\gamma}{7}\enspace.
\end{equation*}
By~\eqref{su} and the last equation, we obtain
\[
\begin{split}
\frac{|U|}{|F|} \leq \frac{|S\cap U|}{|S|}+\frac{\eps^2\gamma}{7}
&\leq (1-3\eps)\gamma+\eps\gamma\\&\leq(1-2\eps)\gamma\leq (1-2\eps)\cdot\frac{|F_x|}{|F|}\enspace.
\end{split}
\]
Hence,
\[
\left|\br{f\in F_x \mid f(x)\geq \max_{f\in S_x}f(x)}\right|
=|F_x|-|F_x\cap U|\geq |F_x|-|U|> 2\eps|F_x|.
\]
Using the last equation with Markov's inequality, we conclude that $\max_{f\in S_x}f(x)<\cost(F_x, x)/(2\eps|F_x|)$. By this and~\eqref{maxs2}, we obtain
\begin{equation}\label{eq:max Hx Sx}
\max_{f\in H_x\cup S_x}f(x)\leq \frac{\cost(F_x, x)}{2\eps|F_x|}.
\end{equation}

Since this theorem assumes $\eps\gamma|F|\geq 1$, we have
\[
\frac{1}{|F|}
= \frac{(1-2\eps)\gamma}{(1-2\eps)\gamma|F|}
\leq \frac{(1-2\eps)\gamma}{(1-3\eps)\gamma|F|+1}
< \frac{(1-2\eps)\gamma}{|H_x|}\enspace.
\]
Combining the last equation and~\eqref{eq:max Hx Sx} in~\eqref{enet72} yields
\begin{equation}\label{costsxx}
\begin{split}
\frac{\cost(S_x,x)}{|S|}&\leq \frac{\cost(H_x,x)}{|F|}+\eps^2 \gamma \max_{f\in H_x\cup S_x} f(x)\\
&\leq (1-2\eps)\gamma\cdot\frac{\cost(H_x,x)}{|H_x|}
+\eps \gamma \cdot\frac{\cost(F_x, x)}{2|F_x|}\enspace.
\end{split}
\end{equation}

Since $H_x$ contains the $|H_x|$ functions $f\in F_x$ with the smallest value $f(x)$, we have that $\cost(H_x, x)/|H_x|\leq \cost(F_x, x)/|F_x|$. Using this in~\eqref{costsxx} yields
\[
\begin{split}
\frac{\cost(S_x,x)}{|S|}&\leq(1-2\eps)\gamma\cdot\frac{\cost(F_x,x)}{|F_x|}
+\eps \gamma \cdot\frac{\cost(F_x, x)}{2|F_x|}
\leq (1-\eps)\gamma\cdot\frac{\cost(F_x,x)}{|F_x|}\enspace.
\end{split}
\]
Multiplying the last equation by $|S|/|S_x|$ bounds the right hand side of~\eqref{myprove15a} as follows.
\begin{equation}\label{righthand}
\frac{\cost(S_x,x)}{|S_x|}
\leq \frac{(1-\eps)\gamma}{(1-3\eps)\gamma}\cdot\frac{\cost(F_x,x)}{|F_x|}
\leq (1+3\eps)\cdot\frac{\cost(F_x,x)}{|F_x|}\enspace.
\end{equation}

We now bound the left hand side of~\eqref{myprove15a} in a similar way.
Let $T_x$ denote the $\big\lceil
 \gamma(1-6\eps)|S|\big\rceil$ functions $f\in S$ with the smallest value $f(x)$.
Since $1\leq \eps\gamma|S|$, we have
\[
\begin{split}
|T_x|&< (1-6\eps)\gamma|S|+1
\leq (1-5\eps)\gamma|S| \\
&\leq (1-2\eps)(1-3\eps)\gamma|S|
\leq (1-2\eps)|S_x|.
\end{split}
\]
By the last equation and Markov's inequality,
\begin{equation}\label{maxs}
\max_{f\in T_x} f(x)\leq \frac{\cost(S_x,x)}{2\eps|S_x|}\enspace.
\end{equation}

Let $Y=\br{f\in F\mid f(x)< \max_{f\in G_x}f(x)}$. Since $Y\subset G_x$, we have $|Y|\leq (1-6\eps)\gamma|F|$. Since $S$ is an $(\eps^2\gamma/7)$-approximation of $F$, substituting $r=\max_{f\in Y}f(x)$ in Definition~\ref{shatter} yields
\[
\frac{|S\cap Y|}{|S|} \leq \frac{|Y|}{|F|}+\frac{\eps^2\gamma}{7}
< (1-2\eps)(1-3\eps)\gamma\leq (1-2\eps)\cdot\frac{|S_x|}{|S|}\enspace.
\]
That is, $|S\cap Y|< (1-2\eps)|S_x|$. Hence,
\[
\left|\br{f\in S_x \mid f(x)\geq \max_{f\in G_x}f(x)}\right|
=|S_x|-|S_x\cap Y|\geq |S_x|-|S\cap Y|>
2\eps|S_x|\enspace.
\]

Using the last equation with Markov's inequality, we conclude that $\max_{f\in G_x}f(x)< \cost(S_x, x)/(2\eps|S_x|)$. By this and~\eqref{maxs}, we obtain
\begin{equation*}\label{max_Gx_TX}
\max_{f\in G_x\cup T_x}f(x)\leq \frac{\cost(S_x, x)}{2\eps|S_x|}.
\end{equation*}
Since $\eps\gamma|S|\geq 1$, we have
\[
\frac{1}{|S|}=\frac{(1-5\eps)\gamma}{(1-5\eps)\gamma|S|}\leq \frac{(1-5\eps)\gamma}{(1-6\eps)\gamma|S|+1}\leq \frac{(1-5\eps)\gamma}{|T_x|}\enspace.
\]
By Theorem~\ref{smaller}, we have
\begin{equation*}%\label{enet7}
\begin{split}
\frac{\cost(G_x,x)}{|F|}\leq\frac{\cost(T_x,x)}{|S|}+\eps^2\gamma \max_{f\in G_x\cup T_x} f(x).
\end{split}\end{equation*}
Combining the last three equations yields
\[
\begin{split}
\frac{\cost(G_x,x)}{|F|}
&\leq \frac{\cost(T_x,x)}{|S|}+\eps^2\gamma\cdot \max_{f\in G_x\cup T_x} f(x)\\
&\leq (1-5\eps)\gamma\cdot\frac{\cost(T_x,x)}{|T_x|}+\eps\gamma\cdot\frac{\cost(S_x, x)}{2|S_x|}\\
&\leq (1-5\eps)\gamma\cdot\frac{\cost(S_x,x)}{|S_x|}+\eps\gamma\cdot\frac{\cost(S_x, x)}{|S_x|}\leq (1-4\eps)\gamma\cdot \frac{\cost(S_x,x)}{|S_x|}.
\end{split}
\]
Multiplying the last equation by $(1-3\eps)|F|/|G_x|$ yields
\[
\begin{split}
(1-3\eps)\cdot\frac{\cost(G_x,x)}{|G_x|}
&\leq \frac{(1-3\eps)(1-4\eps)\gamma|F|}{|G_x|}\cdot \frac{\cost(S_x,x)}{|S_x|} \\
&\leq \frac{(1-3\eps)(1-4\eps)}{1-6\eps}\cdot \frac{\cost(S_x,x)}{|S_x|}\leq \frac{\cost(S_x,x)}{|S_x|}\enspace.
\end{split}
\]
The last equation and~\eqref{righthand} proves~\eqref{myprove15a} as desired.
\end{proof}

Using Theorems~\ref{enettheorem} and~\ref{thm:three_levels_of_gamma}, we get the following corollary.
\begin{corollary}\label{corcoreset}
Let $\eps\in (0,1/4)$, and $\gamma\in (0,1]$. Let $F$ be a set of functions from a set $X$ to $\image$. Let $S$ be a sample of at least
$$\frac{c}{\eps^4\gamma^2}\left(\dim(F)+\log\left(\frac{1}{\delta}\right)\right)$$ i.i.d functions from $F$, where $c$ is a sufficiently large constant. Suppose $|F|\geq |S|$. Then, with probability at least $1-\delta$, $S$ is a $(\gamma,\eps)$-coreset of $F$.
\end{corollary}

\section{Robust medians: From ($\gamma,\eps)$-coresets to $(\gamma,\eps,\alpha,\beta)$-medians}
\label{sec:robustapprox}
In this section we discuss the notion of {\em robust medians} stated in the Introduction and tie it to the notion of $(\gamma,\eps)$-coresets discussed in the last section.
{Roughly speaking, a robust median is a subset of points $Y$ from $X$ that acts as a {\em bi-criteria} clustering of $F$ when considering outliers.
More specifically, our robust medians will be parametrized by four parameters: $\gamma, \eps, \alpha$ and $\beta$. The parameter $\gamma$ (or to be precise $1-\gamma$) will specify the fraction of outliers considered. The parameter $\eps$ is a slackness parameter crucial to the proof of our theorems to come. The parameter $\alpha$ is the approximation ratio between the obtained clustering by $Y$ and the optimal $1$-median clustering.
Finally, the parameter $\beta$ will denote the size of $Y$. In several cases, we will just take $\beta$ to be $1$, and will remove the parameter $\beta$ from our notation.}

\begin{definition}[cost to a set of items]
For a set $Y\subseteq X$, we denote
$\Cost(F,Y)=\sum_{f\in F}\min_{y\in Y}f(y).$
\end{definition}

\begin{definition}[robust median]\label{robustapprox}
Let $F$ be a set of $n$ functions from a set $X$ to $\image$. Let $0<\eps,\gamma<1$, and $\alpha>0$. For every $x\in X$, let $F_x$ denote the $\big\lceil \gamma n\big\rceil$ functions $f\in F$ with the smallest value $f(x)$. Let $Y\subseteq X$, and let $G$ be the set of the $\lceil(1-\eps)\gamma n\rceil$ functions $f\in F$ with smallest value $f(Y)=\min_{y\in Y}f(y)$. The set $Y$ is called a \emph{$(\gamma,\eps,\alpha,\beta)$-median} of $F$, if $|Y|=\beta$ and
\[
\Cost(G, Y)\leq \alpha\min_{x\in X}\cost(F_x, x)\enspace.
\]
For simplicity of notation, a \emph{$(\gamma,\eps,\alpha)$-median} is a shorthand for a $(\gamma,\eps,\alpha,1)$-median.
\end{definition}
Let $F$ be a set of functions from $X$ to $\image$.
In the previous section we proved that a small $(\gamma,\eps)$-coreset of $F$ can be constructed using algorithms that compute $\eps$-approximation of $F$.
In particular, a random sample $S$ of $F$ is such a $(\gamma,\eps)$-coreset.
%, if the dimension of $F$ is low.
In this section we prove that the $(\gamma,\eps,\alpha)$-median of $S$ is also an $(O(\gamma),O(\eps),\alpha)$-median of $F$. In other words, if we have a (possibly inefficient) algorithm for computing the $(\gamma,\eps)$-median of a small coreset $S$, then we can compute a similar median for the original set $F$ in time linear in $n$.

\begin{theorem}\label{robust}
Let $F$ be a set of functions from a set $X$ to $\image$. Let $\eps\in(0,1/10)$, $\gamma\in (0,1]$. Suppose that $S$ is a $(\gamma,\eps)$-coreset of $F$, {and that $|F| \geq |S|\geq 2/(\eps\gamma)$}. Let $\alpha>0$. Then a $\big((1-\eps)\gamma,\eps,\alpha)$-median of $S$ is also a $(\gamma,4\eps,\alpha)$-median of $F$.
\end{theorem}

\begin{proof}
%Let $c_0=1, c_1=4, c_2=3, c_3=1, c_4=1$.
For every $x\in X$, let $F_x$ denote the $\big\lceil \gamma|F|\big\rceil$ functions $f\in F$ with the smallest value $f(x)$.
\begin{itemize}
\item Let $x^*\in X$ and $F^*\subseteq F$, such that $|F^*|= \big\lceil \gamma|F|\big\rceil$ and $\cost(F^*, x^*)= \min_{x\in X}\cost(F_x,x)$.
\item Let $x'$ be a $\big((1-\eps)\gamma,\eps,\alpha)$-median for $S$
\item Let $G$ denote the $\big\lceil(1-4\eps)\gamma|F|\big\rceil$ functions $f\in F$ with the smallest value $f(x')$
\item Let $S'$ denote the $\big\lceil(1-3\eps)\gamma|S|\big\rceil$ functions $f\in S$ with the smallest value $f(x')$
\item Let $S^*$ denote the $\big\lceil(1-\eps)\gamma|S|\big\rceil$ functions $f\in S$ with the smallest value $f(x^*)$
\end{itemize}

We have
\begin{equation}\label{S'bound}
|S'|=\big\lceil(1-3\eps)\gamma|S|\big\rceil\leq \big\lceil(1-\eps)(1-\eps)\gamma|S| \big\rceil\enspace.
\end{equation}
%c_3\leq c_0
Since $1 \leq 1$, we have $|S^*|=\lceil (1-\eps)\gamma|S|\rceil\geq \lceil (1-\eps)\gamma|S|\rceil$. Using this,~\eqref{S'bound} and the fact that $x'$ is a $\big((1-\eps)\gamma,\eps,\alpha\big)$-median of $S$, we have
\begin{equation}\label{costst}
\cost(S', x')\leq \alpha\cdot\cost(S^{*}, x^*)\enspace.
\end{equation}
Since $S$ is a $(\gamma,4\eps)$-coreset of $F$, it is $(\gamma,\eps)$-good for $F$; see Definition~\ref{def:three_levels_of_gamma}.
By this, and since $|G|\leq \big\lceil (1-2\eps)\gamma|F|\big\rceil$, and $|S'|\geq (1-\eps)\gamma|S|$, we obtain
\begin{equation*}
\label{myprove5}
(1-\eps)\cdot\frac{\cost(G, x')}{|G|}
\leq \frac{\cost(S', x')}{|S'|}\enspace.
\end{equation*}
Since $S$ is a $(\gamma,\eps)$-coreset of $F$, we have that
\begin{equation*}
\label{myprove6}
\frac{\cost(S^{*}, x^*)}{|S^{*}|}\leq (1+\eps)\cdot\frac{\cost(F^{*}, x^*)}{|F^{*}|}\enspace.
\end{equation*}
By~\eqref{costst} and the last two equations, we obtain
\begin{equation}\label{costGX}
\begin{split}
\cost(G, x')
&\leq \frac{|G|\cost(S', x')}{(1-\eps)|S'|}\\
&\leq \frac{|G|\alpha}{(1-\eps)|S'|}\cdot\cost(S^{*}, x^*)\\
&\leq \frac{|G|\alpha}{(1-\eps)|S'|}\cdot\frac{|S^*|(1+\eps)}{|F^{*}|}\cdot\cost(F^{*}, x^*)\enspace.
\end{split}
\end{equation}

By the assumption of the theorem, we have $|S|\geq 2/(\eps\gamma)$, so $1\leq \eps\gamma|S|/2$. Hence, \[
|S^*|\leq (1-\eps)\gamma|S|+1\leq (1-\eps/2)\gamma|S|.
 \]
Similarly, since $1\leq 4\eps\gamma|F|/2$,
\[
|G|\leq (1-4\eps)\gamma|F|+1\leq (1-4\eps/2)\gamma|F|.
\]
Therefore,
\[
\begin{split}
\frac{|G|\alpha}{(1-\eps)|S'|}\cdot\frac{|S^*|(1+\eps)}{|F^{*}|}
&\leq  \frac{(1-4\eps)\gamma|F|\alpha}{(1-\eps)(1-3\eps)\gamma|S|}\cdot
\frac{(1-\eps)\gamma|S|\cdot(1+\eps)}{\gamma|F|}\\
&=  \frac{(1-4\eps)\alpha}{(1-\eps)(1-3\eps)}\cdot
\left(1-\eps\right)(1+\eps)
\leq
\alpha\enspace.
\end{split}
\]
Using the last equation with~\eqref{costGX} yields
\[
\cost(G, x')\leq \alpha\cdot\cost(F^{*}, x^*)\enspace.
\]
Hence, $x'$ is a $(\gamma, 4\eps,\alpha)$-median of $F$ as desired.
\end{proof}

In the following (immediate) corollary, we use the same parameters as in Theorem~\ref{robust}.
\begin{corollary}\label{cor y}
Let $Y\subseteq X$ be a set of size $\beta$ that contains a $\big((1-\eps)\gamma,\eps,\alpha)$-median of $S$. Then $Y$ is a $(\gamma,4\eps,\alpha,\beta)$-median of $F$.
\end{corollary}

%\mikel{Removed proof as it is trivial (if I am not missing anything)
%\begin{proof}
%Let $y\in Y$ denote a $\big((1-\eps)\gamma,\eps,\alpha)$-median of $S$. Since $y\in Y$, we have
%\begin{equation}\label{costFY}
%\Cost(F,Y)\leq \cost(F,y).
%\end{equation}
%Let $G$ denote the set of $\lceil (1-4\eps)\gamma\rceil$ functions $f\in F$ with the smallest value $f(y)$. For every $x\in X$, let $F_x$ denote the set of $\lceil \gamma|F|\rceil$ functions with the smallest value $f(x)$.
%By Theorem~\ref{robust}, $y$ is a $(\gamma,4\eps,\alpha)$-median of $F$. Hence,
%\[
%\cost(F,y)\leq \alpha\min_{x\in X}\cost(F_x,x).
%\]
%Using~\eqref{costFY} with the last equation yields $\Cost(F,y)\leq \alpha\min_{x\in X}\cost(F_x,x)$, which proves the corollary.
%\end{proof}
%}

Suppose that for a small subset $S$ from $F$, we can compute a $(\gamma,\eps,\alpha,\beta)$-median $Y$ for $\beta\geq1$.
For $\beta=1$, we showed in Lemma~\ref{robust} that if $S$ is a robust coreset for $F$ then $Y$ is a robust median for $F$.
Unfortunately, this does not hold for $\beta>1$. However, if we use stronger assumptions on the set $S$, the following theorem proves that $Y$ is indeed a robust median in this case.
More specifically, we will need $S$ to be an approximation  to an enhanced version of the function set $F$.
The enhanced function set corresponding to $F$ is one which takes as input {\em subsets} $Y \subset X$ (and naturally outputs the minimum evaluation over points in $Y$).
In a later section, will will use the theorem below to construct efficient bicriteria approximation algorithms from  inefficient ones.

\begin{theorem}Let $\beta\geq 1$ be an integer, $0\leq \eps\leq 1/10$, $0<\gamma\leq 1$, and $\alpha>0$.
\begin{itemize}
\item Let $F$ be a set of functions from $X$ to $[0,\infty)$ such that $|F|\geq 1/(\eps^2 \gamma)$.
\item For every $f\in F$ define $h_f:X\cup X^{\beta}\rightarrow [0,\infty)$ as $h(Y)= \min_{y \in Y} f(y)$.
\item Let $S$ be a $(\gamma,\eps)$-coreset for $H=\br{h_f\mid f\in F}$, such that $|S|\geq 1/(\eps^2\gamma)$.
\item Let $Y$ be a $((1-\eps)\gamma,\eps,\alpha,\beta)$-median for $S_{\ind X}$.
\end{itemize}
Then $Y$ is a $(\gamma,4\eps,\alpha(1+10\eps),\beta)$-median for $F_{\ind X}$.
\end{theorem}
\begin{proof}
Let $G\subseteq H$ denote the $\lceil (1-4\eps)\gamma|F|\rceil$ functions $h_f\in H$ with the smallest value $h_f(Y)=\min_{y\in Y}f(y)$.
Let $S_Y$ denote the $\lceil (1-2\eps)\gamma|S|\rceil$ functions $f\in S$ with the smallest value $f(Y)$.
Since $S$ is a $(\gamma,\eps)$-coreset for $H$, it is also $(\gamma,2\eps)$-good for $H$; see Definition~\ref{def:three_levels_of_gamma}. Hence,
\begin{equation}\label{bi1}
(1-2\eps)\cdot\frac{\cost(G,Y)}{|G|}\leq \frac{\cost(S_Y,Y)}{|S_Y|}.
\end{equation}
For every $x\in X$, let $S_x$ denote the $\lceil (1-\eps)\gamma|S|\rceil$ functions $f\in S$ with the smallest value $f(x)$.
Let $z$ be the item that minimizes $\cost(S_z,z)$ over $z\in X$. The theorem assumes $|S|\geq 1/(\eps^2\gamma)$. Therefore
\[
\begin{split}
|S_Y|&\leq (1-2\eps)\gamma|S|+1 \\&= (1-\eps)^2\gamma|S|+1-\eps^2\gamma|S|\leq (1-\eps)^2\gamma|S|.
\end{split}
\]
By this and the definition of $Y$,
\begin{equation}\label{bi2}
\cost(S_Y,Y)\leq \alpha\cost(S_z,z).
\end{equation}

For every $x\in X$, let $F_x$ denote the $\lceil \gamma |F|\rceil$ functions $f\in F$ with the smallest value $f(x)$.
Let $x^*$ be a center that minimizes $\cost(F_x,x)$ over $x\in X$. By definition of $z$,
\begin{equation}\label{bi3}
\cost(S_z, z)\leq \cost(S_{x^*},x^*).
\end{equation}
Since $S$ is a $(\gamma,\eps)$-coreset for $H$, we have
\begin{equation}\label{bi4}
\frac{\cost(S_{x^*},x^*)}{|S_{x^*}|} \leq (1+\eps)\cdot\frac{\cost(F_{x^*},x^*)}{|F_{x^*}|}.
\end{equation}
Combining~\eqref{bi1},~\eqref{bi2},~\eqref{bi3} and~\eqref{bi4} yields
\begin{equation}\label{cost G Y}
\begin{split}
\cost(G,Y)\leq \frac{|G|\cdot\cost(S_Y,Y)}{(1-2\eps)\cdot|S_Y|}
&\leq \frac{|G|\alpha\cost(S_z,z)}{(1-2\eps)|S_Y|}\\
&\leq \frac{|G|\alpha\cost(S_{x^*},x^*)}{(1-2\eps)|S_Y|}\\
&\leq \frac{|S_{x^*}|}{|S_Y|}\cdot \frac{|G|}{|F_{x^*}|}\cdot \frac{(1+\eps)\alpha\cdot\cost(F_{x^*},x^*)}{1-2\eps}
\end{split}
\end{equation}
Since $\eps^2\gamma|S|\geq 1$, we have
\begin{equation}\label{SS}
|S_{x^*}|\leq  (1-\eps)\gamma|S|+1\leq \gamma|S|.
\end{equation}

Since $\eps^2\gamma|F|\geq 1$, we have
\begin{equation}\label{GG}
|G|\leq  (1-4\eps)\gamma|F|+1\leq \gamma|F|.
\end{equation}
By plugging~\eqref{GG} and~\eqref{SS} in~\eqref{cost G Y}, we infer that
\[
\begin{split}
\cost(G,Y)\leq \frac{1}{1-2\eps}\cdot \frac{(1+\eps)\alpha\cost(F_{x^*},x^*)}{1-2\eps}
&\leq (1+10\eps)\alpha\cdot\cost(F_{x^*},x^*),
\end{split}
\]
where in the last derivation we used the assumption $\eps\leq 1/10$ of the theorem.
This proves that $Y$ is a $(\gamma,\eps,\alpha(1+10\eps),\beta)$-median of $F_{\ind X}$.
\end{proof}

We conclude this section with a lemma (similar in nature to Theorem~\ref{robust}) that addresses generalized range spaces.

\begin{lemma}\label{functions}
Let $(F,\XX)$ be a function space of dimension $d$.
Let $\gamma\in (0,1]$, $\eps \in (0,1/10)$, $\delta\in (0,1/10)$, $\alpha>0$.
Let $S$ be a random sample of
\[
s=\frac{c}{\eps^4\gamma^2}\left(d+\log\frac{1}{\delta}\right),
\]
i.i.d functions from $F$, where $c$ is a sufficiently large constant that is determined in the proof. Suppose that $x\in\XX(S)$ is a $((1-\eps)\gamma,\eps,\alpha)$-median of $S$, and that $|F|\geq s$. Then, with probability at least $1-\delta$, $x$ is a $(\gamma,4\eps,\alpha)$-median of $F$.
\end{lemma}
\begin{proof}
Let $x^*$ be a $(\gamma,0,1)$-median of $F$, and for all $S \subseteq F$ let $X^{+}(S)=\XX(S)\cup \br{x^*}$.
Notice that $(F,X^{+})$ is a generalized range space as in Definition~\ref{dimfunctionsHigh2}.
The number of ranges in $X^{+}(S)$ is larger by at most $|S|$ than the number of ranges in $\XX(S)$. Hence, $\dim(F,X^{+})\leq d+1$.
Hence, applying Theorem~\ref{enettheorem} and then Corollary~\ref{lemma:mycor2} with $c$ large enough, we obtain that, with probability at least $1-\delta$, $S$ is an $(\eps^2\gamma/63)$-approximation of $F_{\ind X^{+}(S)}$. Assume that this event indeed occurs.
By Theorem~\ref{thm:three_levels_of_gamma}, $S$ is also a $(\gamma,\eps)$-coreset of $F_{\ind X^{+}(S)}$.

Since $X^{+}(S)\subseteq X$, we have that $x$ is a $((1-\eps)\gamma,\eps,\alpha)$-median of $S_{\ind X^{+}(S)}$.
Using Theorem~\ref{robust} with $F=F_{\ind X^{+}(S)}$ and $S=S_{\ind X^{+}(S)}$, we obtain that $x$ is a $(\gamma,4\eps,\alpha)$-median of $F_{\ind X^{+}(S)}$. Since $x^*\in X^{+}(S)$, we infer that $x$ is a $(\gamma,4\eps,\alpha)$-median for $F$.
\end{proof}

\subsection{Techniques for Computing a Robust Median}
\label{sec:tech}
In this section, we use the results of Section~\ref{sec:step2} to reduce the problem of computing the robust median for a set of $n$ points to easier problems on smaller (usually, of size independent of $n$) sets. We assume that sampling $s$ functions from $F$ uniformly can be done in time $O(s)$. Using Theorem~\ref{corcoreset}, Theorem~\ref{robust}, and Corollary~\ref{cor y}, we get the following corollary.

\begin{corollary}\label{robustUsingSlowRobust}
Let $\eps\in(0,1/10)$ and $\delta,\gamma\in (0,1]$. Let $F$ be a set of $n\geq 1/(\eps\gamma)$ functions from $X$ to $[0,\infty)$. Suppose that we have an algorithm that receives a set $S\subseteq F$ of size
    \[\displaystyle |S|=\Theta\left(\frac{\dim(F)+\log(1/\delta)}{\gamma^2\eps^4}\right)\enspace,\]
and returns a set $Y$, $|Y|\leq \beta$ that contains a $\big((1-\eps)\gamma, \eps,\alpha\big)$-median of $S$ in time $\satime$.\\
Then a $(\gamma,4\eps,\alpha,\beta)$-median of $F$ can be computed, with probability at least $1-\delta$, in time $\satime + O(|S|)$.
\end{corollary}

The reduction stated in the corollary above (approximately) preserves the quality of the median with respect to $\gamma$. In cases, it is useful to show a connection between medians for $S$ with $\gamma =1$ and medians for $F$ which arbitrary $\gamma$. This point is addressed in the next corollary.

\begin{corollary}\label{cor4}
Let $\eps\in(0,1/4)$ and $\delta,\gamma\in (0,1]$. Let $F$ be a set of $n\geq 1/(\eps\gamma)$ functions from a set $X$ to $\image$. Suppose that we have an algorithm that receives a set $S\subseteq F$ of size         \[\displaystyle |S|=\Theta\left(\frac{\dim(F)+\log(1/\delta)}{\gamma^2\eps^4}\right)\enspace,\]
and returns a $(1,\eps,\alpha)$-median of $S$ in time $\onemedtime$.
Then a $(\gamma,4\eps,\alpha)$-median of $F$ can be computed, with probability at least $1-\delta$, in time $$\robusttime=O\big(\onemedtime\cdot t\cdot\exp\br{2\gamma |S|\ln |S|}\big)\enspace\enspace,$$
where $\ftime$ is the time it takes to compute $f(x)$ for a pair $f\in F$ and $x\in X$.
\end{corollary}

\xmr{The proof below needs a brush up - I found it hard to follow - I added the points which I think are mistaken. Danny: I rewrote.}

\begin{proof}
We first compute a $((1-\eps)\gamma,\eps,\alpha)$-median $z^*$ for $S$.
Let $x^*$ be a $((1-\eps)\gamma,0,\alpha)$ for $S$. Let $T^*$ be the $\gamma'=\lceil (1-\eps)\gamma\rceil$ functions $f\in S$ with the smallest value $f(x^*)$.
Let $y$ be a $(1,0)$-median of $T^*$. Hence, $\cost(T^*,y)\leq \cost(T^*,x^*)$.
Let $z$ be a $(1,\eps)$-median of $T^*$. For every $x\in X$, let $T_x$ denote the $\lceil(1-\eps)\gamma'\rceil$ functions $f\in S$ with the smallest value $f(x)$.  Therefore, $\cost(T_z,z)\leq \cost(T^*,y)$.

We compute a $(1,\eps,\alpha)$-median for every set $T\subseteq S$ of size $\gamma'$, and choose $z^*$ to be the median that minimizes $\cost(T_{z^*},z^*)$. Hence, $\cost(T_{z^*},z^*)\leq \alpha\cost(T_z,z)$.
Combining the last equations yields
\[
\cost(T_{z^*},z^*)\leq \alpha\cost(T_z,z)\leq \alpha\cost(T^*,y)\leq \alpha\cost(T^*,x^*).
\]
Hence, $z^*$ is a $((1-\eps)\gamma,\eps,\alpha)$ for $S$ as desired.

We compute $z^*$ using exhaustive search over all possible $|S|^{O(|T^*|)}\leq \exp\br{2\gamma |S|\ln |S|}$ subsets of size $|T^*|$ of $S$. The proof now follows by applying Corollary~\ref{robustUsingSlowRobust} with $\beta=1$.
\end{proof}

%We now prove Theorems \ref{thm:median} and \ref{thm:out} restated here as the following Corollary.
\section{Centroid Sets}
\label{sec:centaa}

\mw{I made this a section. Need to change ``road map''.}

In this section we define and analyze the notion of a {\em centroid} set.
Roughly speaking, a centroid set in a subset of the centers $X$ that includes a robust median for every subset $S \subseteq F$.
The notion of centroid sets will be later tied to that of {\em weak} coresets as outlined in the Introduction.

Recall that by Corollary~\ref{cor4}, in order to compute a $(\gamma,4\eps,\alpha,\beta)$-median of $F$ for $0< \gamma\leq 1$ in time \emph{independent} in $n$, it suffices to compute a $(1,\eps,\alpha)$ median for a small set $S$ in some \emph{finite} time (even exponential in $|S|$).

\begin{definition}\label{def:cent}
Let $F$ be a set of functions from $X$ to $[0,\infty)$.
A $(\gamma,\eps,\alpha,\beta)$-centroid set for $F$ is a set $\cent\subseteq X^{\beta}$ that contains as an element a $(\gamma,\eps,\alpha,\beta)$-median of $S$, for every $S\subseteq F$. A $(\gamma,\eps,\alpha)$-centroid set is a shorthand for a $(\gamma,\eps,\alpha,1)$-centroid set.
\end{definition}

We start with the following simple lemmas that follows directly by our definitions.

\begin{lemma}\label{above}
Let $F$ be a set of functions from $X$ to $[0,\infty)$. Let $\alpha,\beta,\gamma>0$ be parameters. Then, for every two parameters $1>\eps'\geq \eps\geq 0$ a $(\gamma,\eps,\alpha,\beta)$-median of $F$ is also a $(\gamma,\eps',\alpha,\beta)$-median of $F$.
\end{lemma}

\begin{lemma}\label{relax}
Let $F$ be a set of non-negative functions, $\gamma\in(0,1]$ and $\eps',\gamma'\in [0,1]$. Then every $(\gamma,0,\alpha,\beta)$-centroid set of $F$ is a $(\gamma',\eps',\alpha,\beta)$-centroid set of $F$.
\end{lemma}
\begin{proof}
Let $\cent$ be a $(\gamma,0,\alpha,\beta)$-centroid set for $F$.
Let $S \subseteq F$.
We will show that $\cent$ includes a $(\gamma',\eps,\alpha,\beta)$ median for $S$.
Then using Lemma~\ref{above} and Definition~\ref{def:cent}, we can conclude our assertion.
Let $x^*$ be a $(\gamma',0,1)$-median of $S$.
Let $m=\lceil \gamma'|S| \rceil$, and let $G$ denote the $\lfloor (m-1)/\gamma\rfloor+1$ functions $f\in S$ with the smallest value $f(x^*)$.
By Definition~\ref{def:cent} $\cent$ contains a $(\gamma,0,\alpha,\beta)$-median $Y$ for $G$.
Let $H$ denote the $\lceil \gamma|G|\rceil$ functions $f\in S$ with the smallest value $f(x^*)$.
Let $V$ denote the $\lceil\gamma|G|\rceil$ functions $f\in S$ with the smallest value $f(Y)$.
Hence,\begin{equation}\label{cv}
\Cost(V,Y)\leq \alpha\Cost(H,x^*).
\end{equation}

By denoting $a=|G|-(m-1)/\gamma$, and noting that $0<a\leq 1$, we have
\[
|V|=|H|=\lceil \gamma |G|\rceil
=\left\lceil \gamma \left(\frac{m-1}{\gamma}+a\right)\right\rceil
=\lceil m-1+\gamma a\rceil
=m=\lceil \gamma'|S| \rceil,
\]
where in the last deviation we used the assumption $\gamma>0$.
By the previous equation and~\eqref{cv}, we have that $Y$ is a $(\gamma',0,\alpha,\beta)$-median for $S$.
Using Lemma~\ref{above}, $Y$ is also a $(\gamma',\eps',\alpha,\beta)$-median for $S$.
Since the proof holds for every $S\subseteq F$, we conclude that $\cent$ is a  $(\gamma',\eps',\alpha,\beta)$-centroid set for $F$.
\end{proof}

\mw{We not present a natural connection between the concept of centroid sets for $F$ to centroid sets to extensions of the family $F$ that may take as input subsets $Y$ of $X$.}

\mrx{In the lemma below - why don't we prove things for general $\gamma$ and $\eps$? Danny: It is already too complicated, and we don't use such generalization. I had problems with your changes and revert. Please reread.}

\begin{lemma}\label{cent2}
Let $F$ be a set of functions from $X$ to $[0,\infty)$. Let $\cent$ be a $(1,0,\alpha,\beta)$-centroid set for $F$. For every $f\in F$
define $f_k$ as the function that for $\ell \leq k$ takes as input $x=(x_1,\cdots,x_\ell)\in X^{1}\cup \cdots\cup X^{k}$ and returns $f_k(x)=\min_{1\leq i\leq \ell} f(x_i)$.
Let $F_k=\br{f_k\mid f\in F}$.
%Let $\cent_k=\cent^k$ be the collection of all $k$-tuples of $\cent$.
%

For every $k$-tuple $Y=(Y_1,\cdots,Y_k)\in \cent^k$, let
\[
\Pi(Y)=\br{(x_1,\cdots,x_k), (x_{k+1},\cdots,x_{2k}),\cdots}\in (X^k)^\beta,
\]
be a partition of $Y_1\cup \cdots \cup Y_k$ into $\beta$ disjoint sets, each of size at most $k$. Let $\cent_k=\{\Pi(Y) \mid Y \in \cent^k \}$.
Then $\cent_k$ is a $(1,0,\alpha,\beta)$-centroid set of size $|\cent_k|=|\cent|^k$ for $F_k$.
\end{lemma}
\begin{proof}
Let $S_k\subseteq F_k$. Let $x^*=(x_1^*,\cdots,x_k^*)\in X^k$ be a $(1,0)$-median for $S_k$, and let $T=\br{f\in F\mid f_k\in S_k}$ be the corresponding functions in $F$. Let $(T_1,\cdots, T_k)$ be a partition of $T$, such that $T_i=\br{f\in T\mid f(x^*_i)=f_k(x^*)}$ for every $1\leq i\leq k$. Fix $i$, $1\leq i\leq k$.
Let $Y_i=\br{x_1,\cdots,x_{\beta}}\in\cent$ be a $(1,0,\alpha,\beta)$-median for $T_i$.
Hence,
\begin{equation}
\label{eq:pprev}
\Cost(T_i,Y_i)\leq \alpha\cost(T_i,x_i^*).
\end{equation}

Let $Y= (Y_1,\ldots\,Y_k)\in\cent^k$.
%Let $\Pi(Z)$ be an arbitrary partition of $Y$ into subsets of size at most $k$.
Summing~\eqref{eq:pprev} over every $1\leq i\leq k$ yields
\[
\Cost(S_k, \Pi(Y))
=\sum_{f\in S_k}\min_{1\leq i\leq k}\min_{y\in Y_i}f(y)
\leq \sum_{i=1}^k \Cost(T_i, Y_i)
\leq \alpha\sum_{i=1}^k\cost(T_i,x_i^*)
= \alpha\cost(S_k,x^*).
\]
Hence, $\Pi(Y)$ is a $(1,0,\alpha,\beta)$ for $S_k$.
Since $\Pi(Y) \in \cent_k$, we conclude that $\cent_k$ is a $(1,0,\alpha,\beta)$-centroid set for $F_k$.
\end{proof}

\mrx{Need to verify proof of next lemma ... Danny:please read again}

\begin{lemma}\label{cent1}
Let $F$ and $F_k$ be defined as in Lemma~\ref{cent2}.
Let $\gamma\in (0,1]$, $\eps \in [0,1)$, $\alpha>0$.
Let $\cent$ be a $(1,0,\alpha)$-centroid set for $F$.
Then there is $x\in \cent^k$ which is a $(\gamma,\eps,\alpha)$-median for $F_k$.
%Let $F$ and $F_k$ as in.. Let $\XX$ be cent for $F$.
%There is x\in \XX^k which is a median of F_k.
\end{lemma}
\begin{proof}
Let $x^*=(x_1^*,\cdots,x_k^*)$ be a $(\gamma,0)$-median for $F_k$.
Let $H_k$ denote the $\lceil\gamma\rceil$ functions $f_k\in F_k$ with the smallest value $f_k(x^*)$. Let $G=\br{f\in F \mid f_k\in H_k}$.
Let $(G_1,\cdots, G_k)$ be a partition of $G$, such that $G_i=\br{f\in G\mid f(x^*_i)=f_k(x^*)}$ for every $1\leq i\leq k$.

For every $1\leq i\leq k$, let $x_i\in \cent$ be a $(1,0,\alpha)$-median for $G_i$. Hence, $\cost(G_i,x_i)\leq \alpha\cost(G_i,x_i^*)$. Let $x=(x_1,\cdots, x_k)\in\cent^k$. We thus have,
\begin{equation}
\label{eq:eq2}
\cost(H_k,x)\leq \sum_{i=1}^k\cost(G_i,x_i)
\leq \sum_{i=1}^k \alpha\cost(G_i,x_i^*)
=\alpha\cost(H_k,x^*).
\end{equation}
That is, $x$ is a $(\gamma,0,\alpha)$-median for $F_k$.
Hence, $x$ is also a $(\gamma,\eps,\alpha)$-median for $F_k$.
%By Lemma~\ref{}, $x$ is a $((1-\eps/4)\gamma,\eps/4,\alpha)$-median for $F_k$.
%

%Let $V^*$ be a partition of $\cent$ into $k$-tuples, each of size at most $k$, such that $x\in V^*$.
%Applying Corollary~\ref{cor y} with $\eps/4$, we have that $V^*$ is a $(\gamma,\eps,\alpha,\beta)$-median for %$F_k$. Since $\cost(F_k,V)=\cost(F_k,V^*)$ we conclude that $V$ is also a $(\gamma,\eps,\alpha,\beta)$-median for $F_k$.
\end{proof}

\section{From $(\gamma,\eps,\alpha,\beta)$-medians to bicriteria approximations}
\label{sec:b}

\begin{definition}[Bicriteria $(\alpha,\beta)$-approximation]
Let $F$ be a set of functions from $X$ to $\image$.
An \emph{$(\alpha,\beta)$-bicriteria approximation} for $F$ is a $(1,0,\alpha,\beta)$-median of $F$.
\end{definition}
\label{sec:bicriteria}

\begin{figure*}
%\centering
\subfigure{\fbox{\begin{minipage}{15cm}
\begin{tabbing}
\noindent\textbf{Algorithm} {\bicriteria}$(F, \eps,\alpha,\beta)$\\
\end{tabbing}
\vspace{-2.5em}
\begin{codebox}
\li $i\gets 1$; $F_1\gets F$
\label{definitions}
\li \While $\displaystyle |F_i|\geq 10/\eps$ \Do \label{boundq}
%\li $\mincost \gets\infty$
\li $\M_i\gets$ A $(3/4,\eps,\alpha,\beta)$-median of $F_i$
\label{lineAlgone}
\li $G_i\gets$ The set of the $\big\lceil (1-5\eps)\cdot3|F_i|/4\big\rceil$ functions $f\in F_i$
  with the smallest value $f(\M_i)$\label{linefour}
\li $F_{i+1}\leftarrow F_i\setminus G_i$\label{addf}
\li $i\gets i+1$ \;
\End
\li $\M_i\gets$ A $(1,0,\alpha,\beta)$-median of $F_i$\label{lastptas}
\li $G_i\gets F_i$\label{Gi_last}
\li \Return $\br{(G_1, Y_1),\cdots, (G_i, Y_i)}$
\end{codebox}
\end{minipage}}}\caption{The algorithm~{\bicriteria}. (A slight change in the algorithm compared to that presented in the Introduction.)\label{fig:alg_bi2}}
\end{figure*}

Let $F$ be a set of $n$ functions from some set $X$ to $\image$. Recall that for a set $X'\subseteq X$, we define $\cost(F, X')=\sum_{f\in F}\min_{x\in X'}f(x)$.
In this section we present the algorithm~{\bicriteria} that receives a set $F$ of $n$ functions, and parameters $\eps\in (0,1)$. It returns a set $X'\subseteq X$, $|X'|\leq \log_2 n$, such that $\cost(F, X')\leq (1+\eps)\cdot\min_{x\in X}\cost(F, x)$. See Fig.~\ref{fig:alg_bi2}.
%We call $X'$ a bicriteria approximation, because it approximates $\min_{x\in X}\cost(F,x)$ by using two criterions: (i) a multiplicative $(1+\eps)$-factor to $\min_{x\in X}\cost(F,x)$, as in regular
%PTAS approximations, and (ii) by using a small set of items $X'\subseteq X$ instead of a single $x\in X$.
The algorithm~{\bicriteria} uses (calls) the following two algorithms:

\begin{itemize}
\item An algorithm that computes a robust-median for a given subset of $F$; see Definition~\ref{robustapprox}
\item A (possibly inefficient) algorithm that receives a set $S\subseteq F$ of size $O(1/\eps)$, and returns a set $Y$ such that $\cost(S,Y)\leq (1+\eps)\min_{x\in X}\cost(S,x)$.
\end{itemize}

The second algorithm receives an input of size independent of $n$, and thus can be inefficient. Algorithms for computing a robust-median of $n$ functions in time linear in $n$ are presented in Section~\ref{sec:tech}.
%Given such an algorithm for computing the $(3/4,\eps)$-median, the algorithm~{\bicriteria} also takes time linear in $n$. This holds for any set of functions $F$, regardless of its dimension. More generally, the algorithm~{\bicriteria} reduces the problem of computing a bicriteria approximation of $F$ to the problem of computing a $(\gamma,\eps)$-median for subsets of $F$.

\begin{theorem}\label{costfx}
Let $F$ be a set of $n$ functions from a set $X$ to $\image$, and let $\alpha,\beta\geq 0$, $0<\eps\leq 1$. Let $\B$ be the set that is returned by the algorithm $\bicriteria(F, \eps/100, \alpha,\beta)$; see Fig.~\ref{fig:alg_bi2}.
Then $Z=\cup_{(G,Y) \in B} Y$ is a $((1+\eps)\alpha,\beta\log n)$-approximation for $F$. That is, $|Z|\leq \beta\log_2 n$ and
\[
\Cost(F, Z)\leq (1+\eps)\alpha\cdot \min_{x\in X}\cost(F, x)\enspace.
\]
\end{theorem}

\begin{proof}
Since $|F|$ is reduced by more than half in each ``while" iteration, there are at most $\log_2 n$ iterations. In every iteration we compute $Y$ such that $|Y|\leq \beta$, so $|Z|\leq \beta\log n$. It is left to bound $\Cost(F,Z)$.

Let $B$ be the set that is returned by a call to the algorithm $\bicriteria(F, \eps,\alpha,\beta)$.
We will prove that
\begin{equation}\label{cost_F_X'}
\Cost(F,Y)=\sum_{(G,Y)\in B}\Cost(G, Y)\leq (1+100\eps)\alpha\cdot \min_{x\in X}\cost(F, x)\enspace.
\end{equation}
which suffices to prove our assertion.
%Since $\br{G\mid (G,Y)\in B}$ is a partition of $F$, and $Y\subseteq Z$ for every $(G,Y)\in B$, this would prove the lemma for a call to $\bicriteria(F, \eps/100,\alpha,\beta)$ with a corresponding $0<\eps\leq 1$.

For every $x\in X$, let $F_x$ denote the $\lceil 3|F|/4 \rceil$ functions $f\in F$ with the smallest value $f(x)$. Let $x^*$ be an item that minimizes $\cost(F_x, x)$ over all $x\in X$. Fix $i$, $1\leq i\leq |B|-1$.  Let $F_i^*$ denote the $\lceil 3|F_i|/4\rceil$ functions $f\in F_i$ with the smallest value $f(x^*)$. Since $Y_i$ is a $(3/4,\eps,\alpha,\beta)$-median of $F_i$, we have
(by the definition of $G_i$) that
\begin{equation}\label{costgi}
\begin{split}
\Cost(G_i, Y_i)\leq \alpha \cost(F^*_i, x^*)\enspace.
\end{split}
\end{equation}

\mw{As above, I added the parameter $\alpha$ in several places below - please check to see that I did not miss anything. I did not mark these changes.}

We denote the functions in $F$ by $F=\br{f_1, \cdots, f_{n}}$, such that $f_a(x^*) \leq f_b(x^*)$ for every $1\leq a< b\leq n$, where ties are broken arbitrarily. Let
\begin{equation}\label{defb}
U_i=\br{f_1, \cdots, f_{n-|F_i|}}, \quad V_i=\br{f_{n-|F_i|+1}, \cdots, f_{n-|F_i|+|F_i^*|}}.
\end{equation}
During the first $(i-1)$ ``while" iterations, an overall of $n-|F_i|$ functions were removed from $F$. Hence,
\begin{equation*}%\label{fstar}
|(U_i\cup V_i)\cap F_i|\geq |U_i|+|V_i|-(n-|F_i|)=|V_i|=|F^*_i|.
\end{equation*}
We thus have $U_i\cup V_i\supseteq F^*_i$. The set $V_i$ contains the $|V_i|=|F^*_i|$ functions $f\in U_i\cup V_i$ with the largest values $f(x^*)$. Hence, $\cost(F^*_i, x^*)\leq \cost(V_i, x^*)$.
Combining~\eqref{costgi} with the last equation yields
\begin{equation*}\label{costfi}
\Cost(G_i, Y_i)\leq \alpha \cost(F^*_i, x^*)\leq \alpha \cost(V_i, x^*)\enspace.
\end{equation*}
By Lines~\ref{lastptas} and~\ref{Gi_last} of the algorithm, we have \begin{equation}\label{prob1}
\Cost(G_{|B|}, Y_{|B|})=\Cost(F_{|B|}, Y_{|B|})\leq \alpha \cdot\cost(F_{|B|}, x^*)\enspace.
\end{equation}
Let $V_{|B|}=F_{|B|}$. Using the last three inequations, we obtain
\begin{equation}\label{boundbyB}
\begin{split}
\sum_{(G,Y)\in B}\Cost(G, Y)
&\leq \alpha \cdot\cost(F_{|B|}, x^*)
+\sum_{i=1}^{|B|-1}\cost(G_i, Y_i)\\
&\leq \alpha \sum_{i=1}^{|B|}\cost(V_i, x^*)\enspace.
\end{split}
\end{equation}

Let,  $1\leq i\leq |B|-1$. We now prove that
\begin{equation}
\label{proveme1}
|V_{i+1}\cap V_i|\leq 24\eps|V_{i+1}|,
\end{equation}
and that for every integer $j$ such that $i+2\leq j\leq |B|$, we have
\begin{equation}\label{proveme2}
V_j\cap V_i=\emptyset.
\end{equation}

Indeed, let $j$ be an integer such that $i+1 \leq j\leq |B|$, and assume $V_j\cap V_i\neq \emptyset$.
We have $|F_{j}|=|F_i|-\sum_{k=i}^{j-1} |G_k|$.
Using the last equation and~\eqref{defb}, we get
\begin{equation}\label{bjk}
\begin{split}
|V_{j}\cap V_i|&\leq n-|F_i|+|F_i^*|-(n-|F_{j}|+1)+1\\
&\leq |F_{j}|-|F_i|+|F_i^*| = |F_i^*|-\sum_{k=i}^{j-1} |G_k|\enspace.
\end{split}
\end{equation}

We have $|G_i|\geq (1-5\eps)\cdot|F_i^*|\geq |F_i^*|/(1+6\eps)$, where in the last deviation we use the assumption $\eps\leq 1/100$ from the beginning of this proof.
Hence,
\begin{equation}\label{ffi}
|F_i^*|\leq (1+6\eps)|G_i|=|G_i|+6\eps|G_i|\enspace.
\end{equation}
Since $i\leq |B|-1$, we have by \mbx{Line~\ref{boundq} that $|F_i|\geq 10/\eps$}. We thus have $$|G_i|\leq \frac{(1-5\eps)\cdot3|F_i|}{4}+1\leq \frac{3|F_i|}{4}\leq 3|F_{i+1}|.$$ Using the last two equations, we obtain
\[
|F_i^*|\leq |G_i|+6\eps |G_i|\leq |G_i|+18\eps|F_{i+1}|\enspace.
\]
Combining the last equation with~\eqref{bjk} yields
\begin{equation}\label{bjk2}
|V_{j}\cap V_i|\leq |G_i|+18\eps|F_{i+1}|-\sum_{k=i}^{j-1} |G_i|.
\end{equation}

We have $|F_{i+1}^*|\geq 3|F_{i+1}|/4$, i.e, $|F_{i+1}|\leq 4|F_{i+1}^*|/3$. Thus, substituting $j=i+1$ in~\eqref{bjk2} yields
\begin{equation*}\label{bjk3}
|V_{i+1}\cap V_i|\leq 18\eps|F_{i+1}|
\leq 24\eps |F_{i+1}^*|= 24\eps|V_{i+1}|\enspace,
\end{equation*}
which proves~\eqref{proveme1}.
If $j\geq i+2$, we have by~\eqref{bjk2}
\[
\begin{split}
|V_{j}\cap V_i|\leq 18\eps|F_{i+1}|-|G_{i+1}|
\leq 18\eps|F_{i+1}|-\frac{|F_{i+1}|}{2}<0\enspace,
\end{split}
\]
which contradicts the fact $|V_{j}\cap V_i|\geq 0$. Hence, the assumption $V_j\cap V_i\neq\emptyset$ implies $j=i+1$. This proves~\eqref{proveme2}.

Using~\eqref{proveme2} with~\eqref{boundbyB}, we infer that
\begin{equation}\label{proveme3}
\begin{split}
\sum_{(G,Y)\in B}\Cost(G, Y)
&\leq \alpha\sum_{i=1}^{|B|}\cost(V_i, x^*)\\
&=\alpha\cdot\cost\left(\bigcup_{1\leq i\leq |B|}V_i, x^*\right)
+\alpha\sum_{i=1}^{|B|-1}\cost( V_{i+1}\cap V_i, x^*)\\
&\leq \alpha\cost(F, x^*)+\alpha\sum_{i=1}^{|B|-1}\cost(V_{i+1}\cap V_i, x^*)\enspace.
\end{split}
\end{equation}

We have $|F_{i+1}|\leq 4|F_{i+1}^*|/3$.
The set $V_{i+1}\cap V_i$ contains the functions $f\in V_{i+1}$ with
the smallest value $f(x^*)$. Hence, Equation~\eqref{proveme1} implies
\begin{equation*}\label{Bismall}
\begin{split}
\cost(V_{i+1}\cap V_i, x^*)&\leq \frac{|V_{i+1}\cap V_i|}{|V_{i+1}|}\cdot \cost(V_{i+1}, x^*)\leq 24\eps\cdot\cost(V_{i+1}, x^*)\\
&=24\eps\cdot\cost(V_{i+1}\setminus V_i, x^*)+24\eps\cdot\cost(V_{i+1}\cap V_i, x^*).
\end{split}
\end{equation*}
That is, $$(1-24\eps)\cdot\cost(V_{i+1}\cap V_i, x^*)\leq 24\eps\cdot\cost(V_{i+1}\setminus V_i, x^*).$$ Since $\eps\leq 1/100$, combining the previous equation in~\eqref{proveme3} yields
\[
\begin{split}
\sum_{(G,Y)\in B}\Cost(G, Y)
&\leq \alpha\cdot\cost(F, x^*)+\alpha\sum_{i=1}^{|B|-1}\frac{24\eps\cdot\cost(V_{i+1}\setminus V_i, x^*)}{1-24\eps}\\
&\leq \alpha\cdot\cost(F, x^*)+100\eps\alpha\sum_{i=1}^{|B|-1}\cost(V_{i+1}\setminus V_i, x^*)\\
&\leq \alpha(1+100\eps)\cdot\cost(F, x^*)
\enspace,
\end{split}
\]
where in the last deviation we used~\eqref{proveme2}. This proves~\eqref{cost_F_X'} as desired.
\end{proof}

%\subsection{Proof of Theorem~\ref{thm:bicriteria}}
In what follows we restate Theorem~\ref{the:costfx_intro} and present its proof.
\begin{theorem}\label{73}
Let $F$ be a set of $n$ functions from a set $X$ to $\image$. Let $0<\eps,\delta<1$, $\alpha,\beta\geq 0$. Then a set $Z\subseteq X$ of size $|Z| \leq\beta\log_2 n$ can be computed such that, with probability at least $1-\delta$, \[
\cost(F, Z)\leq (1+\eps)\alpha\cdot \min_{x\in X}\cost(F, x)\enspace.
\]
This takes time $$\bicriteriatime = O(1)\cdot (n\ftime+\log^2 n\cdot\satime+\ptastime),$$ where:
\begin{itemize}
\item $\ftime$ is an upper bound on the time it takes to compute $f(Y)$ for a pair $f\in F$ and $Y\subseteq X$ such that $|Y|\leq \beta$.
\item $O(\satime)$ is the time it takes to compute, with probability at least $1-\delta/2$, a $(3/4,\eps,\alpha,\beta)$-median for a set $F'\subseteq F$.
\item $O(\ptastime)$ is the time it takes to compute a $(1,0,\alpha,\beta)$-median for a set $F'\subseteq F$ of size $|F'|=O(1/\eps)$.
\end{itemize}
\end{theorem}
\begin{proof}
We present a randomized implementation of the algorithm~{\bicriteria}$(F,\eps,\alpha,\beta)$ in Fig.~\ref{fig:alg_bi2}. The implementation succeed with probability at least $1-\delta$, and its running time is $\bicriteriatime$, as stated in the theorem. By Theorem~\ref{costfx}, this proves the theorem.

Indeed, let $B$ denote the output of a call to {\bicriteria}$(F,\eps,\alpha,\beta)$. Put $i$, $1\leq i\leq |B|$. Suppose that we have an algorithm $\median(F_i,\delta')$ that computes, with probability at least $1-\delta'$, a $(3/4,\eps,\alpha,\beta)$-median $Y_i$ for $F_i$. Calling to $\median(F_i,\delta/\log n)$ in each of the $O(\log n)$ times that Line~\ref{lineAlgone} of the algorithm~{\bicriteria} is executed, would yield an implementation for~{\bicriteria} that succeeds with probability at least $1-\delta$. However, in this implementation, we use $\delta'$ that is dependent of $n$.

Instead, in order to compute $Y_i$, we call $i$ times to $\median(F_i,\delta/2)$, and denote by $x_1,\cdots, x_i$ the returned \mbx{sets}.
\mbx{Note that, here, each $x_i$ is a subset of size $\beta$ from $X$.}
For each such \mbx{set} $x_j$, $1\leq j\leq i$, let $G_j$ denote the $\lceil (1-5\eps)3|F_i|/4 \rceil$ functions $f\in F$ with the smallest value $f(x_j)$.
Let $(G_i,Y_i)$ denote the pair that minimizes $\cost(G_j, x_j)$ over $(G_1,x_1),\cdots, (G_i,x_i)$. The algorithm then continue to Line~\ref{linefour} of the algorithm~{\bicriteria} using this construction of $Y_i$ and $G_i$.

The probability that $Y_i$ is a $(3/4,\eps,\alpha,\beta)$-median of $F_i$ is at least the probability that one or more \mbx{of the items $x_1,\cdots, x_i$ contains a $(3/4,\eps,\alpha,\beta)$-median} of $F_i$. Hence, $Y_i$ is a $(3/4,\eps,\alpha,\beta)$-median of $F_i$ with probability at least $1-(\delta/2)^i$.
By Theorem~\ref{costfx} there are at most $|B|\leq \log_2 n$ iterations. Hence, the probability that the item $Y_i$ would be a $(3/4,\eps,\alpha,\beta)$-median in the $i$th iteration, for every $i$, $1\leq i\leq |B|$, is at least $1-\sum_{i=1}^{\lceil\log_2 n\rceil}(\delta/2)^i\geq 1-\delta$.

The running time of the $i$th iteration of the algorithm~{\bicriteria} is dominated by the above implementation of Line~\ref{lineAlgone}. By the assumption of the lemma, each of the $i$ calls to $\median(F_i,\delta/2)$ takes $O(\satime)$ time.
The \mbx{computation of $G_j$ for} every $1\leq j\leq i$ takes overall of $O(i\ftime|F_i|)$ time using order statistics (\mw{add citation}). Since the size of $F$ is reduced by more than half in each ``while" iteration, the running time of Line~\ref{lineAlgone} over all the $O(\log n)$ iterations is therefore
\mbx{
\[
\begin{split}
\sum_{i=1}^{\log_2 n} O\left(\frac{n}{2^{i-1}}\cdot i\ftime+ i\cdot\satime\right)
&\leq O(n\ftime)\cdot\sum_{i=1}^{\log_2 n} \frac{i}{2^{i-1}}+O(\log^2n)\cdot\satime\\
&=O(n\ftime+\log^2 n\cdot\satime)\enspace.
\end{split}
\]}
By the assumption of this theorem, Line~\ref{lastptas} can be computed in time $\ptastime$. We conclude the that the total running time of the above implementation for $\bicriteria(F,\eps,\alpha,\beta)$ is $\bicriteriatime$ as desired.
\end{proof}

%deterministic streaming bicriteria in logn passes:
%
%see .. for ranomized.. see .. for high..

\section{Applications: Bicriteria for Projective Clustering}
\label{sec:a_app1}

\mrx{Started editing from here: 11.3.11.}

In this section we present several applications of the Theorems presented in Section~\ref{sec:b} addressing bi-criteria approximation.
Our applications are from the context of projective clustering.
We consider several settings of parameters. For each setting we prove appropriate results.
We start with some notation.

\subsection{Notation}\label{notation}
For a point $p\in\REAL^d$ and a set $Q\subseteq\REAL^d$, we define $\dist(p,Q)=\min_{q\in Q}\norm{p-q}$. More generally, for an $m$-tuple $x=(x_1,\cdots,x_m)$ of subsets of $\REAL^d$, we define
\[
\dist(p,x)=\min_{1\leq i\leq m}\dist(p,x_i)=\min_{1\leq i\leq m}\min_{q\in x_i}\norm{p-q}.
\]
We denote by $\proj(p,Q)$ the point $q\in Q_i$ such that $\dist(p,Q)=\norm{p-q}$, where ties are broken arbitrarily.
The span of $Q$ (\mbx{i.e., the affine subspace containing all points in $Q$}) is denoted by $\Span{Q}$. A \emph{$j$-flat} in $\REAL^d$ is a translated (affine) $(j-1)$-dimensional subspace of $\REAL^d$.
For example, a $1$-flat in $\REAL^d$ is a set that consists of a single point.

Let $j,k\geq 1$ be two integers.
Let $X(j,1)$ denote the set of all possible $j'$-flats in $\REAL^d$, $1\leq j'\leq j$. Let $X(j,k)=\bigcup_{m=1}^k \big(X(j,1)\big)^m$ be the union of tuples, where each tuple contains at most $k$ flats, each of dimension at most $(j-1)$. Let $P$ be a set of points in $\REAL^d$. For every point $p\in\REAL^d$, we define the corresponding function $f_p:X(j,k)\rightarrow [0,\infty)$ to be $f_p(x)=\dist(p,x)$, where $x=(x_1,\cdots, x_m)$. We define $F(P,j,k)=\br{f_p\mid p\in P}$ to be the union of these functions.
For every set $S\subseteq F(P,j,k)$, we denote $P_S=\br{p\in P\mid f_p\in S}$.

For $x=(x_1,\cdots,x_m)\in X(j,k)$, we define $\costP(P,x)=\sum_{p\in P}\dist(p,x)$. For a set of tuples, $\{y_i\}_i=Y\subseteq X(j,k)$, we define $\CostP(P,Y)=\sum_{p\in P}\min_{i}\dist(p,y_i)$.
Hence, $\costP(P,x)=\cost(F(P,j,k),x)$ and $\CostP(P,Y)=\Cost(F(P,j,k),Y)$.

\mrx{General remark: to be able to compare the bi-criteria running times for the various examples in this section - we need to add by each final theorem the running time where all constants are encapsulated in the O notation.}

\mrx{General remark: We should remove ``large $d$'' from all subsections - as $d$ is always assumed to be large.  Danny: Done.}

\subsection{$\alpha=2^j$, Small $j$, and $k$}

\mbx{We start by showing how one can obtain an $(\alpha,\beta \log{n})$ bi-criteria approximation in which the approximation ratio $\alpha$ is rather large, and the resulting $\beta$ and running time are of size exponential in $j$ and $\log{k}$.
Our proof has the following structure.

To apply our generic algorithm for bi-criteria approximation, one must (iteratively) find robust medians for given subsets of $F$.
Essentially, this is done via random sampling.
Namely, as we have shown, for any such $F' \subset F$, taking a sufficiently large sample $S$, a robust median for $S$ is also one for $F'$.
To find a $j$-subspace that acts as a $(1,0,\alpha)$-median for $S$ efficiently, we show that one does not have to consider all $j$-flats in $\REAL^d$, but rather only those spanned by $j$ points of $S$.
This effectively allows us to consider a generalized rage space corresponding to $F(P,j,k)$ of dimension $O(jk)$ (instead of the naive dimension of $(dk)$), which determines the size of the random sample $S$ to be independent in $d$.
Hence, using such small random samples $S$, and exhaustively computing for them a robust median will yield our result.
A detailed proof follows.
}

\begin{theorem}[\cite{FFSS07}]\label{ffs07}
Let $P$ be a finite set of points in $\REAL^d$. Let $0\leq j\leq d$. There is a set $M\subseteq P$, $|M|\leq j$, and a flat $x=\Span{M}$ such that, \[
\costP(P, x)\leq 2^{j}\min_{x^*\in X(j,1)}\costP(P,x^*)
\]
\end{theorem}
\begin{theorem}\label{lemma:j}
Let $P$ be a finite set of points in $\REAL^d$, \mbx{and $1\leq j\leq d+1$}. Let $S\subseteq F(P,j,k)$,
\[
\XX(S)=\br{x\in X(j,1):\,\, x=\Span{M}, M\subseteq P_S, |M|\leq j},
\]
and $\XX_k(S)=(\XX(S))^k$.  Then
\renewcommand{\labelenumi}{(\textit{\roman{enumi}})}
\begin{enumerate}
\item $\XX(S)$ \mbx{is of size $O(|S|^{j})$}, and can be computed in \mbx{$O(dj^2)\cdot |S|^{j}$} time.
\item $\dim(F(P,j,k),\XX_k)=O(jk).$
\item  $\XX_k(S)$ is a $(1,0,2^{j})$-centroid set for $S$.
\end{enumerate}
\end{theorem}
\begin{proof}
\noindent\textbf{(i)} \mbx{There are $|\XX(S)|=O(|S|^{j})$} subsets of size at most $j$ of $S$. For a fixed subset $Q$ of $|Q|\leq j$ points from $S$, we use the QR decomposition in order to compute the flat that is spanned by them. This takes $O(dj^2)$ time.\\
\noindent\textbf{(ii)} We prove the case $k=1$. The case $k\geq 1$ then follows from Lemma~\ref{clustering}.
 Fix $x\in \XX(S)$. For $r\geq 0$, let $\range(S,x,r)=\br{f\in S\mid f(x)\leq r}$. Hence, $|\br{\range(S,x,r)\mid r\geq 0}|\leq |S|$. Therefore,
\mbx{
\[
|\br{\range(S,x,r)\mid x\in \XX(S), r\geq 0}|\leq O(|S|^{j}\cdot|S|)= |S|^{O(j)}.
\]
}
By our definitions, we obtain $\dim(F(P,j,1),\XX)=O(j)$ as desired.

\noindent\textbf{(iii)}  Follows from Lemma~\ref{cent2} and Theorem~\ref{ffs07}.
\end{proof}

\mrx{Need to double check that $\eps$ can be taken to zero}

\mrx{Danny - there are a few subtle issues in the following lemma ... please check carefully}

\begin{lemma}\label{jk}
Let $P$ be a finite set of points in $\REAL^d$, and $j,k\geq 1$ be two integers. Let $\delta,\eps\in (0,1/10)$, $\gamma\in[0,1]$, and
\begin{equation}
s=\frac{1}{\gamma^2\eps^4}\left(jk+\log\frac{1}{\delta}\right).
\end{equation}
Then, a \mbx{$(\gamma,\eps,2^{j},O(s^{j})/k)$}-median for $F(P,j,k)$ can be computed, with probability at least $1-\delta$, in time $O(ds^2)+s^{O(j)}$.
\end{lemma}
\begin{proof}
Let \mbx{$F=F(P,j,1)$,} $F_k=F(P,j,k)$ and $\XX_k$ be defined as in Theorem~\ref{lemma:j}.
Let $S_k$ be a random sample of $c\cdot s$ i.i.d functions from $F_k$, where $c$ is a sufficiently large constant that will be determined later in the proof.
Here, we assume that $|F_k|\geq c\cdot s$, otherwise we set $S_k=F_k$.
Without loss of generality, we assume that the points in $P$ corresponding to $S_k$ are in $\REAL^{|S_k|}$, otherwise we compute an orthogonal base for these points in $O(ds^2)$ time using the QR decomposition. \mw{should we add a citation?)}

Let $S=\br{f\in F\mid f_k\in S_k}$. By applying Theorem~\ref{lemma:j} with $k=1$, a $(1,0,2^{j})$-centroid set $\XX(S)$, $|\XX(S)|=O(s^{j})$, for $S$ can be computed \mbx{in time $O(dj^2)\cdot s^{j}$}.
By applying Lemma~\ref{cent1} with $F=S$, $F_k=S_k$, $\cent=\XX(S_k)$, $\eps/4$ and $(1-\eps)\gamma$ there is a  $((1-\eps/4)\gamma,\eps/4,2^j)$-median $x\in (\XX(S_k))^k=\XX_k(S_k)$ for $S_k$.
Applying Lemma~\ref{functions} with the function space $(F_k,\XX_k)$ yields that with probability at least $1-\delta$, $x$ is a $(\gamma,\eps,2^j)$-median of $F_k$.

Let $V$ be an arbitrary partition of $\XX_k(S_k)$ into $\beta=\lceil |\XX_k(S_k)/k\rceil$ sets of size at most $k$.
Since $x\in \XX_k(S_k)$ we have $\cost(F_k,V)\leq\cost(F_k,x)$. Since  $x$ is a $(\gamma,\eps,2^j)$-median of $F_k$, the last equation implies that $V$ is a $(\gamma,\eps,2^j,\beta)$-median of $F_k$.

\mrx{To apply Lemma~\ref{functions} we need to change the size of $s$ and add a multiplicative factor of $1/\eps^4$. But then $\eps$ cannot be zero! Please verify ... this will imply changes in the rest of the paper.}
\mrx{Since $x\in V$, it follows that $V$ is a $(\gamma,\eps,2^{j},s^{O(j)}/k)$-median for $F_k$.}
\mrx{I did not follow two points: 1) Why are we using Lemma~\ref{cent1} at all ... Theorem~\ref{lemma:j} already addresses $k$-tuples? 2) Why are there two sets $V$ and $V'$ in the
original writeup? In general, I suggest reviewing Lemma~\ref{cent1} again ... is the partition in the lemma specific? It is not phrased that way ...}
%Since $x\in V$, $V$ is also a $(\gamma,\eps,2^{j},s^{O(j)}/k)$-median for $F_k$. Since $\Cost(F_k,V')=\Cost(F_k,\V)$ we conclude that $V'$ is a $(\gamma,\eps,2^{j},s^{O(j)}/k)$-median for $F_k$.
\end{proof}

\begin{theorem}\label{lem444}
Let $P$ be a finite set of points in $\REAL^d$, and $j,k\geq 1$ be two integers. Let $\delta\in (0,1/10)$, and let
\[
s=jk+\log\frac{1}{\delta}.
\] A $(2^{j+1},s^{O(j)}k^{-1}\log n)$-bicriteria approximation for $F(P,j,k)$ can be computed, with probability at least $1-\delta$, in time
\[
\bicriteriatime=O(nd s^{O(j)})+O(ds^2\log^2n)+s^{O(j)}\log^2 n=O(nds^{O(j)}).
\]
\end{theorem}
\begin{proof}
By Lemma~\ref{jk}, a $(\gamma,1/2,2^{j},s^{O(j)}/k)$-median for a set $F'\subseteq F(P,j,k)$ can be computed, with probability at least $1-\delta/2$, in $\satime=O(ds^2)+s^{O(j)}$ time. Similarly, using $\gamma'=1$ and $\eps'=0$ in the proof of Lemma~\ref{jk}, a $(1,0,2^{j},2^{O(j)}/k)$-median for a set $F'$ of size $|F'|=O(1)$ can be computed in $\ptastime=|\XX_k|=O(d)+2^{O(j)}$ time. The time it takes to compute the distance between a point to a set of $s^{O(j)}$ $j$-flats is $t=O(ds^{O(j)})$. By applying Theorem~\ref{73} with $\eps=1/2$, and \mbx{$\beta=k^{-1}s^{O(j)}$}, we infer that a $(2^{j},k^{-1}s^{O(j)}\log n)$-bicriteria approximation for $F(P,j,k)$ can be computed, with probability at least $1-\delta$, in time
\[
\begin{split}
\bicriteriatime &= O(1)\cdot (n\ftime+\log^2 n\cdot\satime+\ptastime)\\
&=O(nds^{O(j)})+O(ds^2\log^2n)+s^{O(j)}\log^2 n = O(nds^{O(j)})
\end{split}
\]
\end{proof}

\subsection{$\alpha=1+\eps$, Small $j$ and $k$}

\mbx{We now address an $(\alpha,\beta \log{n})$ bi-criteria approximation in which the approximation ratio $\alpha$ is small.
Our proof follows a similar structure to that given in the previous case of $\alpha = 2^j$.
The main difference here is that we need to present an efficient way to find an $(1,0,1+\eps)$-median for random samples $S$ of $F$.
We first show, as before, that one need not consider all $j$-flats in $\REAL^d$, but rather only $j$-flats contained in the span of approximately $jk/\eps$ points in $S$.
As there are infinitely many such $j$-flats, this will not suffice for our needs, and thus we turn to discretize the set of potential medians to obtain a final set of potential medians of size roughly $|S|^{jk/\eps}$.
Once our potential set of medians (i.e., our centroid set) has been established, we continue as we did in the previous section.
A detailed proof follows.
We start by presenting a few known assertions.}

\begin{theorem}[\cite{ShyVar07}]\label{shyvar}
Let $P$ be a set of points in $\REAL^d$, $1\leq j\leq d$, and let $0<\eps<1/4$.
Let $x^*$ be a $j$-flat that minimizes $\costP(P,x^*)$ over every $x^*\in X(j,1)$.
Then there is a set $M\subseteq P$, $\displaystyle |M|\leq\frac{10j\log(1/\eps)}{\eps}$, and a $j$-flat $x\subseteq \Span{M}$ such that:
\begin{enumerate}
\item $\costP(P,x)\leq  (1+\eps)\costP(P,x^*)$.
\item Given $x^*$, $x$ can be computed in $O(ndM)$ time.
\end{enumerate}
\end{theorem}

\mrx{Moved Lemma below to this location}

\begin{lemma}[\cite{SA, FMSW08}] \label{dim:1median}\label{centfms}
Let $P$ be a set of points in $\REAL^d$, and $j,k\geq 1$ be two integers. Then
\begin{enumerate}
\renewcommand{\labelenumi}{(\roman{enumi})}
\item $\dim(F(P,j,k))=O(dk)$.
\item A $(1,0,1+\eps)$-centroid set $C$ for $F(P,j,k)$ of size $|C|=n^{O(djk\log(1/\eps))}$ can be constructed in $O(|C|)$ time.
\end{enumerate}
\end{lemma}

\mrx{It is surprising that the running time above of $O(|C|)$ does not depend on the dimension $d$ - please double check. Danny: There is exponential dependancy}

\mbx{We now present a technical lemma that we will use in our proofs to come.}

\begin{lemma}\label{jm}
Let $Q$ be an $m$-dimensional subspace of $\REAL^d$, and let $Q'$ be an $(m+1)$-dimensional subspace that contains $Q$. Put $p\in \REAL^d$. There is a point $p'\in Q'$ such that for every $j$, $1\leq j\leq m-1$, and a $j$-flat $x\subseteq Q$ we have
\[
\dist(p,x)=\dist(p',x).
\]
Moreover, $p'$ can be computed in $O(d)$ time.
\end{lemma}
\begin{proof}
Let $p\in \REAL^d$. Let $p'\in Q'$, such that $\proj(p',Q)=\proj(p,Q)$ and $\dist(p',Q)=\dist(p,Q)$. The point $p'$ can be computed by projecting $p$ on $Q$ and then translate it in a direction that is orthogonal to $Q$.
Let $x\subseteq Q$ be a $j$-flat. By the Pythagorean Theorem and the construction of $p'$, for every $q\in Q$ we have
\[
\norm{p-q}=\sqrt{\big(\dist(p,Q)\big)^2+\norm{\proj(p,Q)-q}^2}=\sqrt{\big(\dist(p',Q)\big)^2+\norm{\proj(p',Q)-q}^2}=\norm{p'-q}.
\]
Since $x\subseteq Q$, we have by the last equation that $\dist(p,x)=\min_{q\in x}\norm{p-q}=\dist(p',x)$ as desired.
\end{proof}

The following is a generalization of Theorem~\ref{lemma:j}(i).
\begin{lemma}\label{highd}
Let $P$ be a finite set of points in $\REAL^d$. Let $m\geq j\geq 1$ and $k\geq 1$ be integers. For every set $S\subseteq F(P,j,k)$, let
\begin{equation}\label{def:xxS}
\XX(S)=\br{x\in X(j,1):\,\, x\subseteq \Span{M}, M\subseteq P_S, |M|\leq m},
\end{equation}
and $\XX_k(S)=(\XX(S))^k$. Then $\dim(F(P,j,k),\XX_k)=O(mk)$.
\end{lemma}
\begin{proof}
We prove the case $k=1$. The case $k\geq 1$ follows from Lemma~\ref{clustering}.
Put $S\subseteq F(P,j,1)$, $M\subseteq P_S$ such that $|M|\leq m$, and $Q=\Span{M}$.
Let $X_Q=\br{x\in X(j,1)\mid x\subseteq Q}$ denote all the flats of dimension at most $j$ that are contained in $Q$.
Let $Q'$ be an $(m+1)$-subspace that contains $Q$.
By Lemma~\ref{jm}, for every $p\in P_S$ there is a point $p'\in Q'$ such that
\begin{equation}\label{dd}
\text{$\dist(p,x)=\dist(p',x)$ for every $x\in X_Q$}.
\end{equation}
For every $p\in P$, define $f_{p'}:X_Q\rightarrow [0,\infty)$ to be $f(x)=\dist(p',x)$.
Let $S'=\br{f_{p'} \mid p\in P_S}$ be the union of these functions.

Since both $P_{S'}$ and the flats of $X_{Q}$ are contained in the $(m+1)$-dimensional subspace $Q'$, applying Lemma~\ref{dim:1median}(i) with $d=m+1$ implies that $\dim(S')=O(m)$. By definition of $\dim(\cdot)$, we obtain
\begin{equation}\label{dimd}
\big|\br{\range(S',x,r)\mid x\in X_{Q}, r\geq 0}\big|\leq |S'|^{\dim(S')}\leq |S|^{O(m)}.
\end{equation}
By~\eqref{dd}, for every $r\geq 0$, $x\in X_Q$ and a set $\range(S,x,r)=\br{f\in S\mid f(x)\leq r}=f_{p_1},f_{p_2},\cdots$ there is a corresponding
distinct set: $\range(S',x,r)=\br{f\in S'\mid f(x)\leq r}=f_{p'_1},f_{p'_2},\cdots$. Therefore,
\begin{equation*}
\big|\br{\range(S,x,r)\mid x\in X_Q, r\geq 0}\big|= \big|\br{\range(S',x,r)\mid x\in X_Q,  r\geq 0}\big|.
\end{equation*}
%Since $X_Q\subseteq X_{Q'}$, we thus have
%\begin{equation*}
%\big|\br{\range(S,x,r)\mid x\in X_Q,  r\geq 0}\big|\leq \big|\br{\range(S',x,r)\mid x\in X_{Q'},  r\geq 0}\big|.
%\end{equation*}

Using the last equations with~\eqref{dimd}, we obtain
\[
\big|\br{\range(S,x,r)\mid x\in X_Q, r\geq 0}\big|\leq |S|^{O(m)}.
\]
Taking the union over every possible choice of $Q$ yields
\[
\bigcup_{Q\in\br{\Span{M}:\,\, M\subseteq P_S, |M|\leq m}}\big|\br{\range(S,x,r)\mid x\in X_Q, r\geq 0}\big|\leq |P_S|^{O(m)}\cdot |S|^{O(m)}=|S|^{O(m)}.
\]
Using~\eqref{def:xxS} with the last equation yields
\[
\begin{split}
\big|\br{\range(S,x,r)\mid x\in \XX(S), r\geq 0}\big|
&\leq \bigcup_{Q\in\br{\Span{M}:\,\, M\subseteq P_S, |M|\leq m}}\big|\br{\range(S,x,r)\mid x\in X_Q, r\geq 0}\big|\\
&= |S|^{O(m)}.
\end{split}
\]
By our definitions, we obtain $\dim(F(P,j,1),\XX)=O(m)$ as desired.
\end{proof}

\begin{theorem}\label{1e}
Let $P$ be a finite set of points in $\REAL^d$, and $k\geq 1$ be an integer. Let $S\subseteq F(P,j,k)$,
\begin{equation}
\XX(S)=\br{x\in X(j,1):\,\, x\subseteq \Span{M}, M\subseteq P_S, |M|\leq \frac{10j\log(1/\eps)}{\eps}},
\end{equation}
and $\XX_k(S)=(\XX(S))^k$. Then
\renewcommand{\labelenumi}{(\textit{\roman{enumi}})}
\begin{enumerate}
\item  \[
 \dim(F(P,j,k),\XX_k)=O\left(\frac{jk\log(1/\eps)}{\eps}\right).
\]
\item $\XX_k(S)$ is a (possibly infinite) $(1,0,1+\eps,1)$-centroid set for $S$.
\end{enumerate}
\end{theorem}
\begin{proof}
\noindent\textbf{(i)} Follows from Lemma~\ref{highd}.
\noindent\textbf{(ii)} Follows from Lemma~\ref{cent2} and Theorem~\ref{shyvar}.
\end{proof}

The following centroid set that is constructed using the bound of \mbx{Theorem~\ref{dim:1median}} is similar to the larger and somewhat less general centroid set that is constructed in~\cite{santosh}.

\begin{lemma}\label{1cent}
Let $P$ be a set points in $\REAL^d$.
Let $S\subseteq F(P,j,k)$, and let $\eps\in (0,1)$.
A $(1,0,1+\eps)$-centroid set $C$ for $S$ can be computed in $O(d\cdot|S|^2+|C|)$ time, where
\[
|C|=|S|^{O(j^2k\log^2(1/\eps)/\eps)}.
\]
Moreover, $C\subseteq \XX_k(S)$, where $\XX_k(S)$ is defined in Theorem~\ref{1e}.
\end{lemma}
\begin{proof}
We prove the case $k=1$. The case $k\geq 1$ follows by applying Lemma~\ref{cent2} with $F=S$ and $\beta=1$.
Let $\eps'=\eps/3$, $m=10j\log(1/\eps')/\eps'$, and $M\subseteq P_S$ such that $|M|\leq m$. Let $Q=\Span{M}$,  $X_Q=\br{x\in \XX(S)\mid x\subseteq Q}$, and let $Q'$ be an $(m+1)$-subspace that contains $Q$.
By Lemma~\ref{jm}, for every $p\in P_S$ there is a point $p'\in Q'$ such that
\begin{equation}\label{mydist}
\text{$\dist(p,x)=\dist(p',x)$ for every $x\in X_Q$}.
\end{equation}

For every $p\in P$, define $f_{p'}:X_Q\rightarrow [0,\infty)$ to be $f(x)=\dist(p',x)$.
Let $S_Q=\br{f_{p'} \mid p\in P_S}$ be the union of these functions.
Substituting $P=S_Q$ and $d=m+1$ in Lemma~\ref{centfms}(ii) yields that a $(1,0,1+\eps')$-centroid set $C_Q$ for $S_Q$ of size \mbx{$|C_Q|=|S|^{O(mj\log(1/\eps))}$} can be computed in $O(|C_Q|)$ time.
By~\eqref{mydist}, $C_Q$ is also a $(1,0,1+\eps')$-centroid set for $S_{\ind X_Q}$. Let $\XX_k(S)=\bigcup_Q X_Q$ where the union is over every $Q=\Span{M}$ such that $M\subseteq P_S, |M|\leq m$. Hence, $C=\bigcup_{Q}C_Q$ is a $(1,0,1+\eps')$-centroid set for $S_{\ind \XX_k(S)}$.
By Theorem~\ref{1e}(ii), $\XX_k(S)$ is a $(1,0,1+\eps')$-centroid set for $S$. Hence, \mbx{by definition}, $C$ is a $(1,0,(1+\eps')^2)$-centroid set for $S$.
Since $(1+\eps')^2\leq 1+3\eps'\leq 1+\eps$, $C$ is a $(1,0,1+\eps)$-centroid set for $S$, as desired.

\mbx{The size of $C$ is \[
|C|=\left|\bigcup_{Q}C_Q\right|=|P_S|^{O(m)}\cdot |C_Q|=|S|^{O(j^2\log^2(1/\eps)/\eps)}.
\]}
For the running time, we may compute a base for $\Span{S}$ using, for example, the QR decomposition in $d|S|^2$ time, and then compute $C$ on the $|S|$-dimensional space.
\mrx{Need to change statement of lemma according to change above - an see how this effects future parts of the paper. Danny:Done}
\end{proof}

\begin{lemma}\label{jk2}
Let $P$ be a finite set of points in $\REAL^d$, and $j,k\geq 1$ be two integers. Let $\delta,\eps\in (0,1/10)$ and $\gamma\in(0,1]$.
A $(\gamma,\eps,1+\eps)$-median for $F(P,j,k)$ can be computed, with probability at least $1-\delta$, in time $O(ds^2)+s^{O(j^2k\log^2(1/\eps)/\eps)}$, where
\begin{equation*}
s=\frac{1}{\eps^4\gamma^2}\left(\frac{jk\log(1/\eps)}{\eps}+\log\frac{1}{\delta}\right).
\end{equation*}
\end{lemma}

\begin{proof}
Let $S$ be a random sample of $c\cdot s$ i.i.d functions from $F$, for some constant $c\geq 1$ that will be determined later. Here, we assumed that $|F|\geq c\cdot s$. Otherwise, let $S=F$.
By Lemma~\ref{1cent}, a $(1,0,1+\eps)$-centroid set $C$ for $S$ can be computed in $O(|C|+ds^2)$ time, where

\mrx{Size may change due to previous remarks: Danny: Done}
\[
|C|=s^{O(j^2k\log^2(1/\eps)/\eps)}.
\]
By \mbx{Lemma~\ref{relax}},
%Theorem~\ref{lemma:j}(ii),
$C$ is also a $((1-\eps/4)\gamma,\eps/4,1+\eps)$-centroid set for $S$.
Using exhaustive search over $C$, a $((1-\eps/4)\gamma,\eps/4,1+\eps)$-median $x\in C$ of $S$ can be computed in $O(ds^2+|C|)$ time.
\mrx{Don't we need to do a computation fopr each $x \in C$ that will take time $O(s)$? So it may need to be $O(ds^2+s|C|)$. Danny: Right. This is hidden in the O notation of $|C|$.}
Let $\XX_k(\cdot)$ be defined as in Theorem~\ref{1e}. By Theorem~\ref{1e}, $\XX_k(S)$ is a $(1,0,1+\eps)$-centroid set for $S$, and $\dim(F,\XX_k)=O(jk\log(1/\eps)/\eps)$. By Theorem~\ref{1cent}, $C$ is contained in $\XX_k(S)$, so $x\in \XX_k(S)$. By Theorem~\ref{functions}, for a large enough constant $c$ we have that, with probability at least $1-\delta$, $x$ is a $(\gamma,\eps,1+\eps)$-median for $F(P,j,k)$.
\end{proof}

\mrx{Why are we using the parameter $r$ instead of $s$ in the next theorem? It is confusing.}

\begin{theorem}\label{smalljk}
Let $P$ be a finite set of points in $\REAL^d$, and $k,j\geq 1$ be two integers. Let $\delta\in (0,1/10)$ and
\begin{equation*}
r=\frac{1}{\eps^4}\left(\frac{jk\log(1/\eps)}{\eps}+\log\frac{1}{\delta}\right).
\end{equation*}
Then a $(1+\eps,\log n)$-bicriteria approximation for $F(P,j,k)$ can be computed, with probability at least $1-\delta$, in time
\[
\bicriteriatime = O(ndjk)+O(dr^2)+r^{O(j^2k\log^2(1/\eps)/\eps)}\log^2 n.
\]
\end{theorem}
\begin{proof}
By applying Lemma~\ref{jk2} with $\gamma=3/4$ and $\delta/2$, a $(3/4,\eps,1+\eps)$-median for a set $F'\subseteq F(P,j,k)$ can be computed, with probability at least $1-\delta/2$, in $\satime=O(dr^2)+r^{O(j^2k\log^2(1/\eps)/\eps)}$ time. For a set $S\subseteq F(P,j,k)$, $|S|=O(1/\eps)\leq r$, a $(1,0,1+\eps)$-median $x$ of $S$ can be computed in $\satime$ time using exhaustive search on the centroid set in Lemma~\ref{1cent}.
\mrx{Are we applying an approximation in the last step of our algorithms - namely instead of finding a $(1,0,1+\eps)$ median we are finding a $((1-\eps/4)\gamma,\eps/4,1+\eps)$-median? We should say something about this. Danny:No, this was a bug. Fixed.}
By applying Theorem~\ref{73} with $\beta=1$ and $t=djk$, a $(1+\eps,\log n)$-bicriteria approximation for $F(P,j,k)$ can be computed, with probability at least $1-\delta$, in time
\[
\bicriteriatime = O(ndjk)+O(dr^2)+r^{O(j^2k\log^2(1/\eps)/\eps)}\log^2 n.
\]
\end{proof}

\subsection{$\alpha=1+\eps$, Large $k$, Small $j$}

\mw{We now address an $(\alpha,\beta \log{n})$ bi-criteria approximation in which the ratio $\alpha$ is small, but as apposed to the previous section, the parameter $k$ may depend on $n$.
The main difference in our analysis evolves in the fact the we study ...}
\mw{Danny please add}

\begin{lemma}\label{jk3}
Let $P$ be a finite set of points in $\REAL^d$, and $j,k\geq 1$ be two integers. Let $\delta,\eps\in (0,1/10)$ and $\gamma\in(0,1]$.
Let $\beta=s^{\Theta(j^2\log^2(1/\eps)/\eps)}$, where
\begin{equation*}
s=\frac{1}{\eps^4\gamma^2}\left(\frac{jk\log(1/\eps)}{\eps}+\log\frac{1}{\delta}\right).
\end{equation*}
Then a $(\gamma,\eps,1+\eps,\beta)$-median for $F(P,j,k)$ can be computed in $O(ds^2+k\beta)$ time.
\end{lemma}
\begin{proof}
Let $F_k=F(P,j,k)$ and $F=F(P,j,1)$. Let $S_k$ be a random sample of $c\cdot s$ i.i.d functions from $F_k$, for some constant $c\geq 1$. Here, we assumed that $|F|\geq c\cdot s$. Otherwise, let $S_k=F_k$. Let $S=\br{f\in F\mid f_k\in S_k}$.
By applying Lemma~\ref{1cent} with $k=1$, a $(1,0,1+\eps)$-centroid set $\XX(S)$ for $S$, $|\XX(S)|=k\beta$, can be computed in $O(ds^2+k\beta)$ time. Applying Lemma~\ref{cent1} with $F=S$, $F_k=S_k$, yields that
there is $x\in (\XX(S))^k$ which is a $((1-\eps/4)\gamma,\eps/4,1+\eps)$-median for $S_k$.
Let $\XX_k(S_k)=(\XX(S))^k$. Applying Lemma~\ref{functions} with the function space $(F_k,\XX_k)$ yields that with probability at least $1-\delta$, $x$ is a $(\gamma,\eps,1+\eps)$-median of $F_k$. Assume that this event indeed occurs.

Let $V$ be an arbitrary partition of $\XX_k(S_k)$ into $\beta=\lceil |\XX_k(S_k)/k\rceil$ sets of size at most $k$. Since $x\in \XX_k(S_k)$ we have $\cost(F_k,V)\leq\cost(F_k,x)$. Since  $x$ is a $(\gamma,\eps,1+\eps)$-median of $F_k$, the last equation implies that $V$ is a $(\gamma,\eps,1+\eps,\beta)$-median of $F_k$.
\end{proof}

\mrx{In following theorem I suggest dividing $\beta$ by $k$ and changing text accordingly. This will be more consistent with the lemma above. The current form is very confusing.}

\mrx{In theorem below we still get a running time greater than $k^j$ ... I am not sure I understand why this is good for large $k$. Danny: It is polynomial in $k$}

\mrx{In proof of theorem below there are several mix ups between $s$ and $r$ ... please fix. Danny:fixed}

\begin{theorem}
Let $P$ be a finite set of points in $\REAL^d$, and $j,k\geq 1$ be two integers. Let $\eps,\delta\in (0,1/10)$,
\begin{equation*}
r=\frac{1}{\eps^4}\left(\frac{jk\log(1/\eps)}{\eps}+\log\frac{1}{\delta}\right),
\end{equation*}
and $\beta=r^{{\Theta(j^2k\log^2(1/\eps)/\eps)}}$. Then a $(1+\eps,\beta k^{-1}\log n)$-bicriteria approximation for $F(P,j,k)$ can be computed in time
\[
\bicriteriatime = O(nd\beta)+O(dr^2\log^2n)+r^{O(j^2\log^2(1/\eps)/\eps)}\log^2 n.
\]
\end{theorem}
\begin{proof}
Let $\beta=r^{\Theta(j^2\log^2(1/\eps)/\eps)}/k$. By applying Lemma~\ref{jk3} with $\gamma=3/4$ and $\delta/2$, a $(3/4,\eps,1+\eps,\beta/k)$-median $x$ for a set $F'\subseteq F(P,j,k)$ can be computed, with probability at least $1-\delta/2$, in $\satime=O(dr^2+k\beta)$ time.

For a set $S\subseteq F(P,j,k)$, $|S|=O(1/\eps)$, a $(1,0,1+\eps)$-centroid set $\XX(S)$ for $S$, $|\XX(S)|=k\beta$, can be computed in $O(dr^2+\beta)$ time using Lemma~\ref{1cent}. Applying Lemma~\ref{cent1} with $F=S$, $F_k=S_k$, $\gamma=1$ and $\eps=0$ yields that there is $x\in (\XX(S))^k$ which is a $(1,0,1+\eps)$-median for $S_k$. Hence, an arbitrary partition $V$ of $(\XX(S))^k$ to $k$-tuples is a $(1,0,1+\eps,\beta/k)$-median for $S_k$ that can be computed in $\ptastime=O(k\beta)$ time.

The time it takes to compute the distance between a point to a set of $\beta$-flats is $t=O(d\beta)$.
By Theorem~\ref{73} a $(1+\eps,\beta k^{-1}\log n)$-bicriteria approximation for $F(P,j,k)$ can thus be computed, with probability at least $1-\delta$, in time
\[
\bicriteriatime = O(nd\beta)+O(dr^2\log^2n)+k\beta\log^2 n.
\]
\end{proof}

\subsection{$\alpha=1+\eps$, Large $j$ and $k$}

\mw{Please add some explanatory text of main difference from last sections.}

\begin{lemma}\label{jk4}
Let $P$ be a set of $n$ points in $\REAL^d$, and $j,k\geq 1$ be two integers. Let $\delta,\eps\in (0,1/10)$ and $\gamma\in(0,1]$. Let $S$ be a random sample of
\begin{equation*}
s=\frac{c}{\eps^4\gamma^2}\left(\frac{jk\log(1/\eps)}{\eps}+\log\frac{1}{\delta}\right)
\end{equation*}
i.i.d functions from $P$, where $c$ is a sufficiently large constant that is determined in the proof.
Let \[
Y=\br{(x_1,\cdots, x_k)\in X(j,k)\mid x_i\subseteq \Span{S} \text{for every $1\leq i\leq k$}}.
\]
Then, with probability at least $1-\delta$, $Y$ is a $(\gamma,\eps,1+\eps,\infty)$-median for $F(P,j,k)$.
\end{lemma}
\begin{proof}
Let $\XX_k$ be defined as in Theorem~\ref{1e}. By Theorem~\ref{1e}(ii), $\XX_k(S)$ is a $(1,0,1+\eps)$-centroid set for $S$. Let $\gamma'\leq 1$ and $\eps'\geq 0$. By Lemma~\ref{relax}, $\XX_k(S)$ is also a $(\gamma',\eps',1+\eps)$-centroid set for $S$. Hence, there is a $(\gamma',\eps',1+\eps)$-median $x\in \XX_k(S)$ for $S$. Since $\XX_k(S)\subseteq Y$, we have that $Y$ is a $(\gamma',\eps',1+\eps,\infty)$-median for $S$.

For $\eps'=\eps/4$ and $\gamma'=(1-\eps/4)\gamma$, there is a $((1-\eps/4)\gamma,\eps/4,1+\eps)$-median $x\in \XX_k(S)$ for $S$.
By Theorems~\ref{1e}(i), we have $\dim(F(P,j,k),\XX_k)\leq j^2k\log^2(1/\eps)/\eps$. Using this with Theorem~\ref{functions}, we infer that there is a constant $c$ such that, with probability at least $1-\delta$, $x$ is a $(\gamma,\eps,1+\eps)$-median for $F(P,j,k)$.
Assume that this event indeed occurs. Since $x\in \XX_k\subseteq Y$, we have that $Y$ is a $(\gamma,\eps,1+\eps,\infty)$-median for $F(P,j,k)$.
\end{proof}

\begin{theorem}\label{bi_high_dim}
Let $P$ be a finite set of points in $\REAL^d$, and $k\geq 1$, $j\geq 1$ be two integers. Let $\eps,\delta\in (0,1/10)$ and
\begin{equation*}
r=\frac{1}{\eps^4}\left(\frac{jk\log(1/\eps)}{\eps}+\log\frac{1}{\delta}\right).
\end{equation*}
Then, with probability at least $1-\delta$, an $O(r\log n)$-dimensional subspace $Z$ of $\REAL^d$ that satisfies
\[
\costP(P,Z)\leq (1+\eps)\min_{x^*\in X(j,k)}\costP(P,x^*)
\]
can be computed in time
\[
\bicriteriatime = O(ndr)+O(dr^2\log^2n).
\]
\end{theorem}
\begin{proof}
By applying Lemma~\ref{jk4} with $\gamma=3/4$ and $\delta/2$, a $(3/4,\eps,1+\eps,\infty)$-median $Y$ of a set $F'\subseteq F(P,j,k)$ can be computed, with probability at least $1-\delta/2$, such that all the $k$-flats of $Y$ are contained in an $O(r)$-flat. For a set $F'$ of size $O(1/\eps)$, the span of \mbx{(the points corresponding to)} $F'$ contains a $(1,0,1+\eps,1)$-median of $F'$.

By definition of $Y$, for every $p\in\REAL^d$ we have $\dist(p,Y)=\dist(p,\Span{S})$.
After computing an orthogonal base for $S$ in $O(dr^2)$ time, the time it takes to compute $\dist(p,Y)$ is $t=O(dr)$.
By Theorem~\ref{73} an $O(r\log n)$-flat $Z$ that, with probability at least $1-\delta$, satisfies
\[
\costP(P,Z)\leq (1+\eps)\min_{x^*\in X(j,k)}\costP(P,x^*)
\]
can be computed in time
\[
\bicriteriatime = O(ndr)+O(dr^2\log^2n).
\]
\end{proof}

\subsection{$k$-Median in a Metric Space}

\mw{Please add some explanatory text of main difference from last sections. Danny: I rewrote the proof.}

\begin{theorem}\label{kmedianbi}
Let $(P,\dist)$ be a metric space of $n$ points. Let $k\geq 1$ be an integer, $\eps>0$ and
\[
\beta=\Theta\left(\frac{k+\log(2/\delta)}{\eps^4}\right)\enspace.
\]
A set $B\subseteq P$ of $O(\beta\log n)$ points can be computed in $O(ndk+\log^2 n\beta)$ time such that, with probability at least $1-\delta$,
\[
\cost(P,B) \leq (2+\eps)\cdot \min_{x\in P^k}\cost(P,x).
\]
\end{theorem}
\begin{proof}
For every $p\in P$, define $f_p:P^k\rightarrow [0,\infty)$ to be $f_p(x)=\dist(p,x)$. Let $F=\br{f_p \mid p\in P}$.
For every set $S\subseteq F$ that corresponds to $\T\subseteq P$, let $\XX(S)= \T^k$.
For every $x\in \XX(S)$ and $r\geq 0$, let $\ranges(S,x,r)=\br{f\in S\mid f(x)\leq r}$. Hence,
\[
|\ranges(S)|
=|\br{\range(S,x,r)\mid x\in \XX(S), r\geq 0}|
\leq |S|^k\cdot |S|\leq |S|^{k+1},
\]
so $\dim(F,\XX)=O(k)$.

\mrx{Please check proof for mixups between $S$ and $\T$.}

Let $\gamma=3/4$, $\eps\in(0,1/10)$, $\alpha=2$.
Let $F'\subseteq F$. If $|F'|\geq \beta$, let $S$ be a random sample of $\beta$ i.i.d functions from $F'$. Othersise, we define 	$S=F'$.
Let $x^*=(x_1^*,\cdots,x_k^*)$ be a $(\gamma,\eps,1)$-median for $S$.
Let $y=(y_1,\cdots, y_k)\in \T^k$, such that $y_i$ is the closest point to $x^*_i$ in $\T$, for every $1\leq i\leq k$. Let $\T_{x^*}$ denote the closest \mbx{$\lceil (1-\eps) \gamma |S|\rceil$} points \mbx{of $\T$} to $x^*$. Fix $p\in \T_{x^*}$, and let $x_p$ denote the closest point in \mbx{$x^*$} to $p$. By the triangle inequality,
$\dist(p,y)\leq \dist(p,x_p)+\dist(x_p,y)$, and by definition of $y$, $\dist(x_p,y)\leq \dist(x_p,p)$. Hence, $\dist(p,y)\leq 2\dist(x_p,p)= 2\dist(p,x^*)$. Summing over every $p\in \T_{x^*}$ yields $\cost(\T_{x^*},y)\leq 2\cost(\T,x^*)$. By the last inequality, $y$ is a $((1-\eps)\gamma,\eps, \alpha)$-median of $S$. Since $y\in\T^k$, we have that $\T$ contains a $((1-\eps)\gamma,\eps, \alpha)$-median of $S$.

If $|F'|\geq 1/\eps$, by applying Corollary~\ref{robustUsingSlowRobust} with $F=F'$ and $Y=S$
we can compute, with probability at least $1-\delta/2$, a \mbx{$(\gamma,4\eps,\alpha,\beta)$}-median of $F'$ in time $O(|S|)=O(\beta)$. If $|F'|<1/(\gamma\eps)$, the set $S=F'$ is a trivial $(1,0,\alpha,\beta)$ median for $F'$.

Applying Theorem~\ref{73} with $X=P^k$, $t=d$, $\satime=O(\beta)$, $\ptastime=\satime=O(\beta)$, $\ftime=O(k)$ and $16\eps$ yields that
a set $Z\subseteq P$, $|Z| \leq k\beta\log_2 n$ can be computed such that, with probability at least $1-\delta$, \[
\cost(P, Z)\leq (1+\eps/2)\alpha\cdot \min_{x\in P^k}\cost(P, x)\leq (2+\eps)\cdot \min_{x\in P^k}\cost(P, x)
\]
in time \[
\bicriteriatime = O(1)\cdot (ndk+\log^2 n\beta).
\]
\end{proof}

\section{From bicriteria to $B$-coresets}
\label{sec:app_bcoreset}

In this section we analyze the quality of the coresets obtained via algorithm \coresetA\ (Figure~\ref{fig:algA}).
We present
 of analysis which will be used in sections to come when we derive results for specific clustering problems.

\begin{figure*}
%\centering
\subfigure{\fbox{\begin{minipage}{17cm}
\begin{tabbing}
\noindent\textbf{Algorithm} {\coresetA}$(F,F',s,m,\eps)$
\end{tabbing}
\vspace{-2.5em}
\begin{codebox}
\li For each $f\in F$, let $t_f:X\rightarrow [0,\infty)$ be defined as:
$\displaystyle t_f(x)=\begin{cases}
f'(x) & f'(x)> s_f(x)\\
0& \text{otherwise}
\end{cases}\enspace
$
\li Let $T =\{t_f \mid f\in F\}$.
%\li For each $f\in F$ let $\displaystyle n_f= \left\lceil |F|\cdot\max_{\br{x\in X: f(x)\leq s(f,x)}} \frac{f(x)}{\cost(F,x)} \right\rceil$
\li For each $f\in F$ let $g_f:X\rightarrow [0,\infty)$ be defined as:
$
g_f(x)=\begin{cases}
0& f'(x)> s_f(x)\\
\frac{f(x)}{m_f} & \text{otherwise}
\end{cases}\enspace
$
\label{def:G_f}
\li Let $G_f$ consist of the $m_f$ copies of $g_f$.
\li $G \leftarrow \bigcup_{f\in F}G_f$.\label{Gdef}
\li $S \leftarrow$ An $\eps$-approximation of $G$. \label{constructS}
\End
\li $U \leftarrow \br{g_f\cdot \frac{|G|}{|S|}\quad\Big|\, g_f\in S}$.
\label{def:S}
\li \Return $C \leftarrow T\cup U$.
\end{codebox}
\end{minipage}}}\caption{The algorithm~{\coresetA}.\label{fig:algA}}
\end{figure*}

\begin{theorem}
\label{the:coresetA}
Let $F$ be a set of functions from $X$ to $[0,\infty]$, and $0<\eps<1/4$. Let \mbx{$s:(F,X)\rightarrow [0,\infty)$}, and $m:F\rightarrow \mathbb{N}\setminus \br{0}$.
For each $f \in F$ let $f'$ be a corresponding function associated with $f$, and let $F' = \{f' | f \in F\}$.
For every $x\in X$, let $M(x)=\br{f\in F: f'(x)\leq s_f(x)}$ and assume $f(x)\leq 2s_f(x)$ for every $f\in M(x)$.
Then for $C = \coresetA(F,F',s,m,\eps)$ it holds that
\begin{equation*}
%\label{toprove10}
\begin{split}
\forall x\in X:
&|\cost(F, x)-\cost(C, x)|\leq
\sum_{f\in F\setminus M(x)}\big|f(x)-f'(x)\big|+2\eps\max_{f\in M(x)}\frac{s_f(x)}{m_f}\sum_{f\in F}m_f.
\end{split}
\end{equation*}
\end{theorem}

\begin{proof}
Fix $x\in X$, and let $M=\br{f\in F: f'(x)\leq s_f(x)}$.
For every $f\in M$, we have $\cost(G_f,x)=m_f\cdot g_f(x)=f(x)$.
Moreover, by definition, for every $f \not \in M$ we have $\cost(G_f,x)=0$.
Hence,
\begin{equation}\label{eq:cost(H,x)}
\begin{split}
\cost(F,x)&=\cost(F\setminus M, x)+\sum_{f\in M}f(x)=\cost(F\setminus M, x)+\cost(G, x).
\end{split}
\end{equation}
The first term in the right hand side is approximated by $T$, up to an error of
\begin{equation}\label{cFmX}
|\cost(F\setminus M, x)-\cost(T,x)|=
\left|\sum_{f\in F\setminus M}\big(f(x)-f'(x)\big)\right|
\leq \sum_{f\in F\setminus M}\big|f(x)-f'(x)\big|.
\end{equation}
%Next, we prove that
%\begin{equation}\label{eq:|cost(G,x)-cost(S,x)|}
%|\cost(G,x)-\cost(U,x)|\leq \eps\sum_{f\in F}\max_{x\in X}\lceil s(f,x)\rceil.
%\end{equation}

Since $S$ is a $\eps$-approximation of $G$, by Lemma~\ref{enet2} we obtain
\begin{equation*}
\left|\frac{\cost(G, x)}{|G|}
-\frac{\cost(   S, x)}{|S|}\right|
\leq \eps \cdot \max_{g_f\in G}g_f(x)\enspace.
\end{equation*}
By Step~\ref{def:G_f} of our algorithm, for every $g_f\in G$, we have $f'(x)\leq s_f(x)$. By the assumption $f(x)\leq 2s(f)$ of the theorem, we thus obtain
\[
g_f(x) = \frac{f(x)}{m_f}\leq \frac{2s_f(x)}{m_f}.
\]
By the last two equations,
\begin{equation*}
\left|\frac{\cost(G, x)}{|G|}
-\frac{\cost(S, x)}{|S|}\right|
\leq \eps \cdot\max_{f\in F}\frac{ 2s_f(x)}{m_f}.
\end{equation*}
Multiplying this equation by $|G|$ yields
\[
\left|\cost(G,x)-\frac{|G|}{|S|}\cdot\cost(S,x)\right|
\leq \eps |G|\cdot\max_{f\in F}\frac{2s_f(x)}{m_f}.
\]
Recall that $U=\br{g_f\cdot |G|/|S|\mid g_f\in S}$.
Together with the previous two inequalities, we obtain
\begin{equation}\label{cGx}
\begin{split}
|\cost(G,x)-\cost(U,x)|
&=\left|\cost(G,x)-\frac{|G|}{|S|}\cdot \cost(S,x)\right|\leq \eps |G|\cdot\max_{f\in F}\frac{2s_f(x)}{m_f}.
\end{split}
\end{equation}

We have $\cost(C,x)=\cost(T,x)+\cost(U,x)$. Hence, combining~\eqref{cFmX} and~\eqref{cGx} with the triangle inequality yields
\[
\begin{split}
|\cost(F\setminus M, x)+\cost(G, x)-\cost(C,x)|&=|\cost(F\setminus M, x)+\cost(G, x)-\cost(T,x)-\cost(U,x)|\\
&\leq |\cost(F\setminus M, x)-\cost(T,x)|+|\cost(G, x)-\cost(U,x)|\\
&\leq\sum_{f\in F\setminus M}\big|f(x)-f'(x)\big|+ \eps |G|\cdot\max_{f\in F}\frac{ 2s_f(x)}{m_f}.
\end{split}
\]
Using~\eqref{eq:cost(H,x)}, this proves the theorem, as
\[
\begin{split}
|\cost(F,x)-\cost(C, x)|
&=|\cost(F\setminus M, x)+\cost(G, x)-\cost(C, x)|\\
&\leq \sum_{f\in F\setminus M}\big|f(x)-f'(x)\big|+\eps |G|\cdot \max_{f\in F}\frac{2s_f(x)}{m_f}\\
&=\sum_{f\in F\setminus M}\big|f(x)-f'(x)\big|+\eps\max_{f\in F}\frac{2s_f(x)}{m_f}\sum_{f\in F}m_f.
\end{split}
\]
\end{proof}

\mbx{We now present a few corollaries of Theorem~\ref{the:coresetA} that will be used in the sections to come.}
\mrx{Need to add short intuition for statement of below cor (namely, we are assuming a relaxed/strengthened triangle inequality).}

\begin{corollary}
\label{cor:mean1}
Let $F, X$, $s$, $M$ and $\eps$ be defined as in Theorem~\ref{the:coresetA}.
Let $b>0$. Suppose that for every $x\in X$ and $f\in M(x)$ we have
\[
m_f\geq \frac{s_f(x)}{b\cdot \cost(F,x)},
\]
and, for every $f\in F\setminus M(x)$,
\[
|f(x)-f'(x)|\leq \eps b f(x)
\]
Then for $C = \coresetA(F,F',s,m,\eps)$ it holds that
\begin{equation*}
\forall x\in X:
|\cost(F, x)-\cost(C, x)|\leq \eps b\cost(F,x)\left(1+2\sum_{f\in F}m_f\right)\enspace.
\end{equation*}
\end{corollary}
\begin{proof}
Put $x\in X$. For every $f\in M(x)$, we have
\[
\frac{s_f(x)}{m_f}\sum_{f\in F}m_f
\leq \frac{s_f(x)}{\frac{s_f(x)}{b\cost(F,x)}}\sum_{f\in F}m_f
= b\cost(F,x)\sum_{f\in F}m_f.
\]
For every $f\in F\setminus M(x)$, we have
\[
\sum_{f\in F\setminus M(x)}|f(x)-f'(x)|\leq \eps \sum_{f\in F\setminus M(x)}b f(x)\leq \eps b\cost(F,x).
\]
The Corollary follows by applying Theorem~\ref{the:coresetA} using the last inequalities.
\end{proof}

\mrx{Need to add short intuition for statement of below cor.}

\begin{corollary}\label{corrf}
\label{cor:mean}
Let $F, X, F', s$ and $\eps$ be defined as in Theorem~\ref{the:coresetA}. Let $B\subseteq X$ and $\tau>0$. Suppose that for all $x \in X$ and for all $f \in F$ it holds that
\begin{equation}\label{suppose2}
f'(x)>\frac{f(B)}{\tau}\quad \Rightarrow \quad |f(x)-f'(x)|\leq  \eps\cdot f(x).
\end{equation}

For every $f\in F$ and $x\in X$ assume $s_f(x)=f(B)/\tau$ and define
\[m_f=\left\lceil \frac{|F|\cdot f(B)}{\cost(F,B)}\right\rceil+1.\]
Then for $C = \coresetA(F,F',s,m,\tau^2)$ it holds that
\begin{equation*}
\forall x\in X:
|\cost(F, x)-\cost(C, x)|\leq \eps\cost(F,x)+4\tau\cost(F,B).
\end{equation*}
\end{corollary}
\begin{proof}
Put $x\in X$, $M(x)=\br{f\in F\mid f'(x)\leq s_f(x)}$, and $f\in F$. If $f\in M(x)$, then \mbx{using our definitions}
\[
\frac{s_f(x)}{m_f}=\frac{f(B)}{\tau m_f}\leq \frac{\cost(F,B)}{|F|\tau}.
\]
Otherwise, \mbx{$f \not \in M(x)$}.
Thus $f'(x)>s_f(x)=f(B)/\tau$, so, by~\eqref{suppose2}, $|f(x)-f'(x)|\leq \eps\cdot f(x)$.
Replacing $\eps$ with $\tau^2$ in Theorem~\ref{the:coresetA} yields
\[
\begin{split}
|\cost(F,x)-\cost(C,x)|\leq &\sum_{f\in F\setminus M(x)}\big|f(x)-f'(x)\big|+2\tau^2\max_{f\in M(x)}\frac{s_f(x)}{m_f}\sum_{f\in F}m_f\\
&\leq \sum_{f\in F}\eps f(x)+2\tau^2\cdot \frac{\cost(F,B)}{|F|\tau}\sum_{f\in F} \left(\frac{|F|f(B)}{\cost(F,B)}+1\right)\\
&\leq  \eps\cost(F,x)+4\tau\cost(F,B).
\end{split}
\]
\end{proof}

\begin{corollary}\label{hh}
%[Fix the use of $f(B)$ below to better match the uses of the cor.]
Let $F, X, F'$ and $\eps$ be defined as in Theorem~\ref{the:coresetA}.
Let $B\subseteq X$.
%Let $c_f$ be an arbitrary positive value.
For $f \in F$, let $m_f$ be an arbitrary positive value,
%=\left\lceil \frac{|F|f(B)}{\cost(F,B)}+c_f\right\rceil$
and let $\Delta_f= \frac{3m_f\cdot\cost(F,B)}{\eps^{z-1}\sum_g m_g}$.
Suppose that for all $x \in X$ and for all $f \in F$ it holds that
\begin{equation}\label{suppose4}
|f(x)-f'(x)|\leq \Delta_f.
\end{equation}
For every $f\in F$ and $x\in X$, let $h_f(x)=f(x)-f'(x)+\Delta_f$ and $H=\br{h_f\mid f\in F}$.
For every $h_f\in H$, let $s_{h_f}=h_f$ and $m_{h_f}=m_f$.
Then for $C = \coresetA(H,\emptyset,s,m,\eps^z)$ it holds
\mbx{$\forall x\in X$ that:
\begin{equation*}
|\cost(H,x)-\cost(C,x)|
=
\left|\cost(F,x)-\left(\cost(F',x)+\cost(C,x)-\sum_f\Delta_f\right)\right|\leq 12\eps \cost(F,B).
\end{equation*}
}
\end{corollary}
\begin{proof}
Let $x\in X$.
We have
\[
\begin{split}
\cost(F,x)&=\cost(F',x)+\cost(F,x)-\cost(F',x)\\&=\cost(F',x)+\cost(H,x)-\sum_f\Delta_f.
\end{split}
\]
Hence,
\[
\left|\cost(F,x)-\left(\cost(F',x)+\cost(C,x)-\sum_f\Delta_f\right)\right|
=|\cost(H,x)-\cost(C,x)|
\]
By applying Theorem~\ref{the:coresetA} with $F=F'$ as $H$, we infer that
\[
\begin{split}
|\cost(H,x)-\cost(C,x)|
&\leq 2\eps^z \max_{h_f\in H}\frac{s_{f}(x)}{m_{f}}\sum_{h_f\in H}m_{f}\\
&= 2\eps^z \max_{h_f\in H}\frac{h_f(x)}{m_{f}}\sum_{h_f\in H}m_{f}\\
& = 2\eps^z \max_{f\in F}\frac{f(x)-f'(x)+\Delta_f}{m_{f}}\sum_{h_f\in H}m_{f}\\
& \leq 2\eps^z \max_{f\in F}\frac{\Delta_f}{m_{f}}\sum_{h_f\in H}m_{f}\\
&\leq 2\eps \max_{f\in F}\frac{6m_f\cdot\cost(F,B)}{m_f\sum_g m_g}\sum_{h_f\in H}m_{f}
 = 12\eps \cost(F,B)
\end{split}
\]
%where the last deviation is due to~\eqref{suppose4}.
In the above we use the fact that $|f(x)-f'(x)|\leq \Delta_f$.
We conclude that,
\[
\left|\cost(F,x)-\left(\cost(F',x)+\cost(C,x)-\sum_f\Delta_f\right)\right|\leq 12\eps\cost(F,B).
\]
\end{proof}

\section{From B-Coresets to Metric B-Coresets}
\label{sec:metric_b_coresets}
\newcommand{\GG}{\mathbf{G}}
\newcommand{\KK}{\mathbf{L}}

We now turn to study algorithm \coresetA\ when applied to functions $F$ corresponding to a metric space.
Namely, we show an improved analysis when $F$ and the bi-criteria $B$ correspond to points in a given metric space.
We will use the analysis stated in this section in deriving improved results for specific clustering problems.

\mbx{
In what follows, our set of data elements will correspond to points $P$ in a metric space $(\MM,\dist)$.
The set of functions corresponding to $P$ may be referred to as $F$, $\GG$, $\HH$, or $\KK$ depending on our specific application.
The bi-criteria solution will also consist of points $B$ in $(\MM,\dist)$.
Finally, we will denote certain subsets of points in $(\MM,\dist)$ by $\T$, and the corresponding functions they represent by $S$ (as has been common throughout our presentation).
}
 \begin{figure*}
%\centering
\subfigure{\fbox{\begin{minipage}{15cm}
\begin{tabbing}
\noindent\textbf{Algorithm} {\dimred}$(P, B,\t,\eps,z)$\\
\end{tabbing}
\vspace{-2.5em}
\begin{codebox}
\li \For each $p\in P$\Do
\zi \label{mpdef}\hspace{3cm}$\displaystyle
m_{p}\gets \left\lceil \frac{|P|\dist^z(p,B)}{\sum_{p\in P}\dist^z(p,B)}\right\rceil+1.
$\\[-0.2cm]
\End
\li Pick a non-uniform random sample $\T$ of $\t$
points from $P$, \zi where for every $q\in \T$ and $p\in P$, we have $q=p$ with probability $m_p/\sum_{q\in P}m_q$.
\li For $p\in P$, let $p'= \proj(p,B)$.
\li \For every $p\in \T$ and set $x$ of points, define \Do
\zi \hspace{3cm}$\displaystyle
w(p,x)=\begin{cases}
\frac{\sum_{q\in P}m_q}{m_p\cdot |\T|} & \dist^z(p',x)\leq \frac{\dist^z(p,B)}{\eps^z}\\
0 & \text{otherwise}.
\end{cases}\enspace
$\label{lineTdef}
\End
\li     \For every $p\in P$ and a set $x$ of points, define
\Do
%\zi \hspace{3cm}$p'= \proj(p,B)$
\zi \hspace{3cm}$\displaystyle \label{w2}
w(p',x)=\begin{cases}
0 & \dist^z(p',x)\le \frac{\dist^z(p,B)}{\eps^z}\\
1 & \text{otherwise}.
\end{cases}\enspace
$
\End
\li $D\gets \T\cup \proj(P,B)$
\li \Return $(D,\T,w)$
\end{codebox}
\end{minipage}}}\caption{The algorithm~{\dimred}.\label{fig:alg}}
\end{figure*}

\mbx{
\begin{definition}[$\GG(\cdot)$]\label{GGdef}
Let $P$ and $B$ be two sets of points in a metric space $(\MM,\dist)$, $t \geq 1$ and let $\eps>0$.
For $p\in P$, let $p'= \proj(p,B)$, i.e., the closest point in $B$ to $p$.
For every $p\in P$, define $m_p$ as in Line~\ref{mpdef} of a call to $\dimred(P,B,t,\eps,z)$. See Fig.~\ref{fig:alg}.
Put $\tau=\eps^z/(cz)^z$.
For every $p\in P$, let $g_p:\MM \rightarrow \REAL^+$ be defined as follows:
\[
\displaystyle
g_p(x)=\begin{cases}
\frac{\dist^z(p,x)}{m_p} & \dist^z(p',x)\leq \frac{\dist^z(p,B)}{\eps^z}\\
0 & \text{otherwise}.
\end{cases}\enspace
\]
Notice the close resemblance between the definition of $w(p,x)$ in algorithm $\dimred$ and the definition of $\GG$.
For every $\T\subseteq P$, we then define \[
\GG(\T)=\GG(\T,B,\eps)=\br{g_p\mid p\in \T}.
\]
\end{definition}
}

\begin{lemma}\label{74}
Let $h$ and $h'$ be two functions from a set $X$ to $[0,\infty)$. Let $z\geq 1$, $x\in X$, $f(x)=(h(x))^z$, and $f'(x)=(h'(x))^z$.
Let $B\subseteq X$, $0< \eps<1$, and suppose that
\begin{equation}\label{suppose3}
|h(x)-h'(x)|\leq  h(B).
\end{equation}

Then
\renewcommand{\labelenumi}{(\textit{\roman{enumi}})}
\begin{equation*}\label{suppose2proof}
f'(x)\geq \frac{(18z)^z f(B)}{\eps^z}\quad \Rightarrow \quad |f(x)-f'(x)|\leq \eps f(x).
\end{equation*}
\end{lemma}
\begin{proof}
It suffices to prove that for $\eps<1/(18z)$, we have
\begin{equation}\label{suppose3proof}
f'(x)\geq \frac{f(B)}{\eps^z}\quad \Rightarrow \quad |f(x)-f'(x)|\leq 18\eps z f(x).
\end{equation}

Let $a\geq b\geq 0$.
We have
\begin{equation}\label{az-bz}
\begin{split}
a^z-b^z
&=\sum_{i=2}^{z}a^{i} b^{z-i}-\sum_{i=1}^{z-1}a^i b^{z-i}+\sum_{i=1}^{z-1}a^{z-i} b^i-\sum_{i=2}^{z}a^{z-i} b^{i}-ab(a^{z-2}-b^{z-2})\\
&\leq\sum_{i=1}^{z-1}a^{i+1} b^{z-i-1}-\sum_{i=1}^{z-1}a^i b^{z-i}+\sum_{i=1}^{z-1}a^{z-i} b^i-\sum_{i=1}^{z-1}a^{z-i-1} b^{i+1}\\
&=a\sum_{i=1}^{z-1}a^i b^{z-i-1}-b\sum_{i=1}^{z-1}a^i b^{z-i-1}+a\sum_{i=1}^{z-1}a^{z-i-1} b^i-b\sum_{i=1}^{z-1}a^{z-i-1} b^i\\
&=(a-b)\sum_{i=1}^{z-1} \big(a^i b^{z-i-1}+a^{z-i-1} b^i\big)\\
&\leq (a-b)(z-1)(2a^{z-1})\leq 2z\cdot a^{z-1}\cdot (a-b).
\end{split}
\end{equation}

By substituting $a=\max\br{h(x),h'(x)}$ and $b=\min\br{h(x),h'(x)}$ in ~\eqref{az-bz}, we obtain
\begin{equation}\label{|Delta(p,x)-Delta(p',x)|}
\begin{split}
|f(x)-f'(x)|&=a^z-b^z\leq 2z\cdot a^{z-1}\cdot |h(x)-h'(x)|
\end{split}
\end{equation}
Assume that $f'(x)\geq f(B)/\eps^z$. By taking the $z$th root, we get $h'(x)\geq h(B)/\eps$. That is, $h(B)\leq \eps h'(x)$.
Using this with~\eqref{suppose3} yields
\[
h(B)\leq  \eps h'(x)\leq \eps\cdot (h(x)+h(B))=\eps h(x) +\eps h(B),
\]
i.e, \[
h(B)\leq \frac{\eps h(x)}{1-\eps}\leq  (1+2\eps)\eps h(x)\leq 2\eps h(x).
\]
Using~\eqref{suppose3} again, we thus have $$|h(x)-h'(x)|\leq h(B)\leq 2\eps h(x).$$ Hence, $$a=\max\br{h(x),h'(x)}\leq h(x)+\cc h(x)=(1+\cc)h(x).$$ Combining the last two inequalities in~\eqref{|Delta(p,x)-Delta(p',x)|} yields~\eqref{suppose3proof}, as
\begin{equation*}
\begin{split}
|f(x)-f'(x)|&\leq  2z\cdot(1+\cc)^{z-1}h(x)^{z-1}\cc  h(x)\\
&=4z\eps (1+\cc)^{z-1}f(x)\leq 4z\eps (1+(2/z))^{z-1}f(x) \leq  2e^2 z\eps f(x)\leq 18z\eps f(x),
\end{split}
\end{equation*}
where in the last two deviations we used the assumption $\eps<1/(18z)$.
\end{proof}

\mrx{Maybe add text regarding the fact that we are constructing a weak-coreset below.}

\begin{theorem}\label{mainsub}
Let $(\MM,\dist)$ be a metric space, $P, B\subseteq \MM$, $0<\eps,\delta<1/2$, and $z,t\geq 1$.
\mbx{
Let $(D,\T,w)$ be the output of a call to the algorithm~{\dimred$(P,B,\eps/(cz),\t,z)$}, with
\[
\t \geq \frac{(cz)^{4z}}{\eps^{4z}}\left( \dim(\GG(P),\XX) +\log\frac{1}{\delta}\right),
\]
for a function space $(\GG(P,B,\eps/2),\XX) = (\GG(P),\XX)$.
}
Then, with probability at least $1-\delta$,
\begin{equation*}%\label{SD}
\begin{split}
\forall x\in \XX(\GG(\T,B,\eps^z/2)):&\left|\cost(P,x)-\sum_{p\in D}w(p,x)\cdot \dist(p,x) \right|\leq \eps \cost(P,B)+\eps\cost(P,x),
\end{split}
\end{equation*}
where $\cost(P,x):=\sum_{p\in P}\dist^z(p,x)$.
\end{theorem}
\begin{proof}
\mbx{Let $\GG(\T)=\GG(\T,B,\eps/(cz))$.}
Let $Y=\XX(\GG(\T))$. For every $p\in P$, let $f_p,f'_p:Y\rightarrow [0,\infty)$ such that $f_p(x)=\dist^z(p,x)$, $f'_p(x)=\dist^z(\proj(p,B),x)$, and $m_f=m_p$.
Let $F=\br{f_p\mid p\in P}$ and $F'=\br{f'_p\mid p\in P}$.
Put $x\in Y$ and $p\in P$.
Using Lemma~\ref{74} with $h(x)=\dist(p,x)$ yields
\begin{equation}\label{sass}
f'_p(x)>\frac{(cz)^zf_p(B)}{\eps^z} \Longrightarrow|f_p(x)-f'_p(x)|\leq \eps f_p(x).
\end{equation}

Put $\tau=\eps^z/(cz)^z$.
Let $C$ be the output of a call to ${\coresetA}(F_{\ind Y},F'_{\ind Y},s,m,\tau^2)$ where \mbx{$s_{f}(x)=f(B)/\tau$}.
Using~\eqref{sass}, applying Corollary \ref{corrf} yields
\begin{equation}\label{sas}
\forall x\in Y:|\cost(F, x)-\cost(C, x)|\leq \eps\cost(F,x)+\eps\cost(F,B).
\end{equation}

Let $T=\br{t_{f_p}\mid p\in P}$ and $G=\br{g_{f_p}\mid p\in P}$ be the sets that are defined in Lines~\ref{TTdef} and~\ref{Gdef} , respectively, of the above call to $\coresetA$; see Fig.~\ref{fig:algA}.
\mbx{Note that it holds that $G=\GG(P,B,\eps/(cz))$ by Definition~\ref{GGdef}. Therefore for $\GG(P)=\GG(P,B,\eps/(cz))$ we have that $\dim(G,\XX)=\dim(\GG(P),\XX)$.}
%\mb{
%\begin{equation}\label{GT}
% g_{f_p}(x) = \frac{|\T|}{|G|}\cdot w(p,x)\dist(p,x)=\frac{|\T|}{\sum_{f\in F}m_f}\cdot w(p,x)\dist(p,x),
%\end{equation}
%}
\mbx{In addition, it holds that $t_{f_p}(x)=w(p',x)\dist(p',x)$, where $w(p',x)$ is defined, in Line~\ref{w2} of algorithm~{\dimred}.}
Hence,
\begin{equation}\label{Tx}
\cost(T,x)=\sum_{p\in P}w(p',x)\dist(p',x).
\end{equation}
%and, by Definition~\ref{GGdef}, $G=\GG(P)$.

Let $S=\br{g_{f_p} \mid p\in \T}$.
By the construction of $\T$, we have that $S$ is a random sample of $\t$ i.i.d functions from $G$.
By using a sufficiently large constant $c$ in Theorem~\ref{mycor2}, with probability at least $1-\delta$, $S$ is thus an $\eps^{2z}/(cz)^{2z}$-approximation of $G_{\ind \XX(S)}=G_{\ind Y}$.

We have
\mbx{
$$
\cost(S,x)=\sum_{g_{f_p}\in S}g_{f_p}(x).$$}

\mbx{Also, for $w(p,x)$ defined in algorithm~{\dimred}, notice that our definitions imply that
$$
g_{f_p}(x) = \frac{|\T|}{\sum_{q\in P}m_q}\cdot w(p,x)\dist^z(p,x) = \frac{|\T|}{|G|} \cdot w(p,x)\dist^z(p,x)
$$
Here we use the fact that $G$ is defined in algorithm $\coresetA$ to take $m_f$ copies of each $g_f$.}

Suppose that $S$ was used in Line~\ref{Sdef} of the above call to $\coresetA$.
Using the last equation and~\eqref{Tx} with the construction of $C$, yields
\[
\begin{split}
\cost(C,x)=\cost(T,x)+\frac{|G|}{|S|}\cost(S,x)
&=\sum_{p\in P}w(p',x)\dist^z(p',x)+\sum_{p\in \T}w(p,x)\dist^z(p,x)\\
&=\sum_{p\in D}w(p,x)\dist^z(p,x).
\end{split}
\]
We also have $\cost(F,x)=\sum_{p\in P}\dist^z(p,x)=\cost(P,x)$.
By the last two equations and~\eqref{sas}, we obtain
\[
\forall x\in Y:\left|\cost(P,x)-\sum_{p\in D}w(p,x)\dist^z(p,x)\right|\leq \eps\cost(P,x)+\eps\cost(P,B).
\]
\end{proof}

\subsection{Smaller Coresets}

\mrx{Add discussion, why are we defining $\KK$ and how much does it reduce our sample size ...}

\begin{definition}[$\KK(\cdot)$]\label{GGdef3}
Let $P$ and $B$ be two set of points in a metric space $(\MM,\dist)$, and let $t\geq 1$, and $\eps>0$.
For every $p\in P$, define $m_p$ as in Line~\ref{mpdef} of a call to $\dimred(P,B,t,\eps)$. See Fig.~\ref{fig:alg}.
\mbx{For every $p\in P$, let $\ell_p:\MM \rightarrow \REAL^+$ be defined as follows:}
$$\ell_p(x)=\frac{\dist(p,x)-\dist(\proj(p,B),x)}{m_p}+\frac{3\cdot\cost(P,B)}{\sum_q m_q}.$$
For every $\T\subseteq P$, we then define \[
\mbx{\KK(\T)=\KK(\T,B,\eps)=\br{\ell_p\mid p\in \T}.}
\]
\end{definition}

\mrx{In theorem below - it is not clear why we are using a large $c$ in the call \dimred$(P,B,\eps/c,\t)$. This appeared in the original version so I did not change.}

\begin{theorem}\label{mainsub3}
\mbx{
Let $(\MM,\dist)$ be a metric space, $P, B\subseteq \MM$, $0<\eps,\delta<1/2$, and $t\geq 1$.
Let $(D,\T,w)$ be the output of a call to the algorithm~{\dimred$(P,B,\eps/c,\t)$}, with
\[
\t \geq \frac{c}{\eps^2}\left( \dim(\KK(P),\XX) +\log\frac{1}{\delta}\right),
\]
for
a function space $(\KK(P,B,\eps/c),\XX) = (\KK(P),\XX)$ where $c$ is a sufficiently large constant.
}
For every $p\in\T$, let \[
w(p)=\frac{\sum_{z\in P}m_z}{m_p\cdot |\T|}
\]

Then, with probability at least $1-\delta$,
\begin{equation*}%\label{SD}
\begin{split}
&\forall x\in \mbx{\XX(\KK(\T,B,\eps/c))}:\\
&\left|\cost(P,x)-\left(\cost(\proj(P,B),x)+\sum_{p\in \T}w(p)\dist(p,x)-\sum_{p\in \T}w(p) \dist(\proj(p,B),x)\right) \right|\\
&\leq \eps \cost(P,B).
\end{split}
\end{equation*}
\end{theorem}

\begin{proof}
\mbx{Let $\KK(\T)=\KK(\T,B,\eps/c)$}.
Let $X=\XX(\KK(\T))$.
For every $p\in P$, let $p'=\proj(p,B)$, and $h_p:X\rightarrow [0,\infty)$ be defined as
$$h_p(x)=\dist(p,x)-\dist(p',x)+\frac{3m_p\cdot\cost(P,B)}{\sum_q m_q}.$$
Let $s_{h_p}:X\rightarrow [0,\infty)$ be defined as $s_{h_p}(x)=h(x)$, and
$H=\br{h_{p}\mid p\in P}$. Let $C$ be the output of a call to $\coresetA(H,\emptyset,s,m,\eps)$; see Fig.~\ref{fig:algA}.

Let $G=\br{g_{h_p}\mid p\in P}$ be the set that is defined in Line~\ref{Gdef} of the above call to $\coresetA$. Note that for every $p\in \T$ we have
\begin{equation*}%\label{GT2}
 g_{h_p}(x)=\frac{h_p(x)}{m_p}.
\end{equation*}
We thus have $G=\KK(P)$, so $\dim(G,\XX)=\dim(\KK(P),\XX)$.
Let $S=\br{g_{h_p} \mid p\in \T}=\KK(\T)$.
By the construction of $\T$, we have that $S$ is a random sample of $\t$ i.i.d functions from $G$. By Theorem~\ref{mycor2}, with probability at least $1-\delta$ we have that $S$ is an $\eps$-approximation of $G_{\ind \XX(S)}=G_{\ind X}$.
Assume that this event indeed occurs, and suppose that $S$ was used in Line~\ref{Sdef} of the above call to $\coresetA$.

Put $x\in X$. We start by proving that the functions $h_p$ are positive.
Namely, for $\Delta_p =\frac{3m_p\cdot\cost(P,B)}{\sum_q m_q}$ we show that $|\dist(p,x)-\dist(p',x)| \leq \Delta_p$.
By the triangle inequality, for $p \in P$
\[
|\dist(p,x)-\dist(p',x)| \leq \dist(p,B).
\]
Thus it suffices to prove that
$$
\dist(p,B) \leq \frac{3m_p\cdot\cost(P,B)}{\sum_q m_q}
$$
Now,
\[
\begin{split}
&  \frac{3m_p\cdot\cost(P,B)}{\sum_q m_q} \geq \frac{3\frac{|P|\dist(p,B)}{\cost(P,B)}\cdot\cost(P,B)}{\sum_q \left(\frac{|P|\dist(p,b)}{\cost(P,B)}+1\right)} \\
& = \frac{3\left( \frac{|P|\dist(p,B)}{\cost(P,B)}\right)\cdot\cost(P,B)}{2|P|} > \dist(p,B)
 \end{split}
\]

For every $p\in P$, let $f_p,f'_p:X\rightarrow [0,\infty)$ such that $f_p(x)=\dist(p,x)$ and $f'_p(x)=\dist(p',x)$.
Let $F=\br{f_p\mid p\in P}$ and $F'=\br{f'_p\mid p\in P}$.
By Corollary~\ref{hh},
\begin{equation*}
\left|\cost(F,x)-\left(\cost(F',x)+\cost(C,x)-\sum_p\Delta_p\right)\right|\leq 12\eps\cost(F,B).
\end{equation*}
It also holds that
$
\sum_p \Delta_p = 3\cost(F,B).
$
Thus,
\begin{equation*}
|\cost(F,x)-\left(\cost(F',x)-3\cost(F,B)+\cost(C,x)\right)|\leq 12\eps\cost(F,B).
\end{equation*}
We also have
\[
\begin{split}
\sum_{p\in \T}w(p)\cdot\frac{m_p\cost(P,B)}{\sum_{q\in P}m_q}=\cost(P,B)=\cost(F,B)
\end{split}
\]
and
\[
\begin{split}
\sum_{p\in \T}w(p)\big(\dist(p,x)-\dist(\proj(p,B),x)\big)
&=\frac{\sum_{q\in P}m_q}{|S|}\sum_{p\in \T}\frac{h_p(x)-\Delta_p}{m_p}\\
&=\frac{|G|}{|S|}\sum_{p\in \T}\frac{h_p(x)}{m_p}-\frac{\sum_{q\in P}m_q}{|S|}\sum_{p\in \T}\frac{\Delta_p}{m_p}\\
\\&=\frac{|G|}{|S|}\sum_{g_{h_p}\in S} g_{h_p}(x)
-\sum_{p\in \T}w(p)\Delta_p
\\&=\cost(C,x)-\sum_{p\in \T}w(p)\cdot\frac{3m_p\cost(P,B)}{\sum_{q\in P}m_q}
\end{split}
\]
Using the last three inequalities,

\mrx{Need to elaborate derivation below - I wasn't able to verify (although I recall verifying this a while back.}

\[
\begin{split}
&\left|\cost(P,x)-\left(\cost(\proj(P,B),x)+\sum_{p\in \T}w(p)\big(\dist(p,x)-\dist(\proj(p,B),x)\big)\right)\right|
\\&=\left|\cost(F,x)-\left(\cost(F',x)+\cost(C,x)-\sum_{p\in \T}w(p)\cdot\frac{3m_p\cost(P,B)}{\sum_{q\in P}m_q}\right)\right|\\
&=|\cost(F,x)
-\left(\cost(F',x)+\cost(C,x)-3\cost(F,B)\right)|
\leq 12\eps\cost(F,B)=6\eps\cost(P,B).
\end{split}
\]
\end{proof}

\begin{theorem}\label{mainsub4}
\mbx{
Let $(\MM,\dist)$ be a metric space, $P, B\subseteq \MM$, $0<\eps,\delta<1/2$, and $z\geq 1$.
Let $(D,\T,w)$ be the output of a call to the algorithm~{\dimred$(P,B,\eps/c,\t,2)$}, with
\[
\t \geq \frac{c}{\eps^2}\left( \dim(\KK(P),\XX) +\log\frac{1}{\delta}\right),
\]
for a function space $(\KK(P,B,\eps/c),\XX) = (\KK(P),\XX)$ where $c$ is a sufficiently large constant.
}
For every $p\in\T$, let \[
w(p)=\frac{\sum_{q\in P}m_q}{m_q\cdot |\T|}
\]

Then, with probability at least $1-\delta$,
\begin{equation*}%\label{SD}
\begin{split}
&\forall x\in \mbx{\XX(\KK(\T,B,\eps/c))}:\\
&\left|\cost(P,x)-\left(\cost(\proj(P,B),x)+\sum_{p\in \T}w(p)\dist^2(p,x)-\sum_{p\in \T}w(p) \dist^2(\proj(p,B),x)\right) \right|\\
&\leq \eps \cost(P,B).
\end{split}
\end{equation*}
\end{theorem}
\begin{proof}
\mbx{Let $\KK(\T)=\KK(\T,B,\eps/c)$}.
Let $X=\XX(\KK(\T))$.
For every $p\in P$, let $p'=\proj(p,B)$, and $h_p:X\rightarrow [0,\infty)$ be defined as
$$h_p(x)=\dist(p,x)-\dist(p',x)+\frac{3m_p\cdot\cost(P,B)}{\sum_q m_q}.$$
Let $s_{h_p}:X\rightarrow [0,\infty)$ be defined as $s_{h_p}(x)=h(x)$, and
$H=\br{h_{p}\mid p\in P}$. Let $C$ be the output of a call to $\coresetA(H,\emptyset,s,m,\eps)$; see Fig.~\ref{fig:algA}.

Let $G=\br{g_{h_p}\mid p\in P}$ be the set that is defined in Line~\ref{Gdef} of the above call to $\coresetA$. Note that for every $p\in \T$ we have
\begin{equation*}
 g_{h_p}(x)=\frac{h_p(x)}{m_p}.
\end{equation*}
We thus have $G=\KK(P)$, so $\dim(G,\XX)=\dim(\KK(P),\XX)$.
Let $S=\br{g_{h_p} \mid p\in \T}=\KK(\T)$.
By the construction of $\T$, we have that $S$ is a random sample of $\t$ i.i.d functions from $G$. By Theorem~\ref{mycor2}, with probability at least $1-\delta$ we have that $S$ is an $\eps$-approximation of $G_{\ind \XX(S)}=G_{\ind X}$.
Assume that this event indeed occurs, and suppose that $S$ was used in Line~\ref{Sdef} of the above call to $\coresetA$.

Put $x\in X$. We start by proving that the functions $h_p$ are positive.
Namely, for $\Delta_p =\frac{3m_p\cdot\cost(P,B)}{\sum_q m_q}$ we show that $|\dist(p,x)-\dist(p',x)| \leq \Delta_p$.
By the triangle inequality, for $p \in P$
\[
|\dist(p,x)-\dist(p',x)| \leq \dist(p,B).
\]
Thus it suffices to prove that
$$
\dist(p,B) \leq \frac{3m_p\cdot\cost(P,B)}{\sum_q m_q}
$$
Now,
\[
\begin{split}
&  \frac{3m_p\cdot\cost(P,B)}{\sum_q m_q} \geq \frac{3\frac{|P|\dist(p,B)}{\cost(P,B)}\cdot\cost(P,B)}{\sum_q \left(\frac{|P|\dist(p,b)}{\cost(P,B)}+1\right)} \\
& = \frac{3\left( \frac{|P|\dist(p,B)}{\cost(P,B)}\right)\cdot\cost(P,B)}{2|P|} > \dist(p,B)
 \end{split}
\]

For every $p\in P$, let $f_p,f'_p:X\rightarrow [0,\infty)$ such that $f_p(x)=\dist(p,x)$ and $f'_p(x)=\dist(p',x)$.
Let $F=\br{f_p\mid p\in P}$ and $F'=\br{f'_p\mid p\in P}$.
By Corollary~\ref{hh},
\begin{equation*}
\left|\cost(F,x)-\left(\cost(F',x)+\cost(C,x)-\sum_p\Delta_p\right)\right|\leq 12\eps\cost(F,B).
\end{equation*}
It also holds that
$
\sum_p \Delta_p = 3\cost(F,B).
$
Thus,
\begin{equation*}
|\cost(F,x)-\left(\cost(F',x)-3\cost(F,B)+\cost(C,x)\right)|\leq 12\eps\cost(F,B).
\end{equation*}
We also have
\[
\begin{split}
\sum_{p\in \T}w(p)\cdot\frac{m_p\cost(P,B)}{\sum_{q\in P}m_q}=\cost(P,B)=\cost(F,B).
\end{split}
\]
Using the last two inequalities,

\mrx{Need to elaborate derivation below - I wasn't able to verify (although I recall verifying this a while back.}

\[
\begin{split}
&\left|\cost(P,x)-\left(\cost(\proj(P,B),x)+\sum_{p\in \T}w(p)\big(\dist(p,x)-\dist(\proj(p,B),x)\big)\right)\right|
\\&=\left|\cost(F,x)-\left(\cost(F',x)+\cost(C,x)-\sum_{p\in \T}w(p)\cdot\frac{3m_p\cost(P,B)}{\sum_{q\in P}m_q}\right)\right|\\
&=|\cost(F,x)
-\left(\cost(F',x)+\cost(C,x)-3\cost(F,B)\right)|
\leq 12\eps\cost(F,B)=6\eps\cost(P,B).
\end{split}
\]
\end{proof}

\section{$k$-Median in a Metric Space}
\label{sec:a:kmed}

\mbx{We now present the results obtained by applying our framework on the $k$-median problem in metric spaces. We start by presenting a constant factor approximation.} We assume that the time to compute the distance between two points in the metric space is $O(d)$.

\subsection{Constant Factor Approximation}
\begin{theorem}\label{constfactor}
Let $(P,\dist)$ be a metric space of $n$ points. Let $0<\delta<1/2$.
A set $x\in P^k$ can be computed in $O(ndk+k^2+\log^2(1/\delta)\log^2 n)$ time, such that, with probability at least $1-\delta$,
\[
\sum_{p\in P}\dist(p,x)\leq O(1)\cdot \min_{x^*\in P^k}\sum_{p\in P}\dist(p,x^*).
\]
\end{theorem}
\begin{proof}
Let $\beta=k+\log(2/\delta)$.
Let $x^*$ denote the $k$-tuple that minimizes $\sum_{p\in P}\dist(p,x)$ over every $x\in P^k$.
By Theorem~\ref{kmedianbi}, a set $B\subseteq P$ of $O(\beta\log n)$ points can be computed in $O(1)\cdot (ndk+\log^2 n\beta)$ time such that, with probability at least $1-\delta$,
\begin{equation}\label{PBMedian}
\cost(P,B) \leq O(1)\cdot\cost(P,x^*).
\end{equation}

Let $x\in B^k$ be a set such that
\begin{equation}\label{bestx}
\cost(\proj(P,B),x)\leq O(1)\min_{y^*\in B^k}\cost(\proj(P,B),y^*).
\end{equation}
Since $\proj(P,B)$ contains at most $|B|$ distinct weighted points, such a \mbx{set $x$ can} be computed in $|B|^2$ time; see survey in~\cite{MetPla04}.

Fix $p\in P$. Using the triangle inequality,
\[
\dist(p,x)\leq \dist(p,\proj(p,B))+\dist(\proj(p,B),x)=\dist(p,B)+\dist(\proj(p,B),x).
\]
Summing this over every $p\in P$ yields
$\cost(P,x)\leq \cost(P,B)+ \cost(\proj(P,B),x)$.
By this and~\eqref{PBMedian}, we obtain
 \begin{equation}\label{yess}
\cost(P,x)\leq \cost(P,B)+ \mbx{O(1)\cdot}\cost(\proj(P,B),x^*).
\end{equation}

Fix $p\in P$. Using the triangle inequality,
\[
\dist(\proj(p,B),x^*)\leq \dist(\proj(p,B),p)+\dist(p,x^*)=\dist(p,B)+\dist(p,x^*).
\]
Summing this over every $p\in P$ yields
$\cost(\proj(P,B),x^*)\leq \cost(P,B)+\cost(P,x^*)$. Using~\eqref{yess} with the last inequality yields
\[
\cost(P,x)\leq \cost(P,B)+ \mbx{O(1)\cdot}\cost(\proj(P,B),x^*)
\leq \mbx{O(1)\cdot}\cost(P,B)+ \mbx{O(1)\cdot}\cost(P,x^*).
\]
By this and~\eqref{PBMedian}, we obtain $\cost(P,x)\leq O(1)\cost(P,x^*)$, which proves this theorem.
\end{proof}

\begin{figure*}
\subfigure{\fbox{\begin{minipage}{15cm}
\begin{tabbing}
\noindent\textbf{Algorithm} {\rkmedian$(P,B,t,\eps,z)$}
\end{tabbing}
\vspace{-0.8cm}
\begin{codebox}
\li \For each $b\in B$ \Do
\li     $P_b\gets$ the set of points in $P$ whose closest point in $B$ is $b$. Ties are broken arbitrarily.
\End
\li \For each $b\in B$ and $p\in P_b$ \Do \hspace{3cm}
\zi \hspace{3cm}$\displaystyle
m_{p}\gets \left\lceil \frac{|P|\dist^z(p,B)}{\sum_{p\in P}\dist^z(p,B)}\right\rceil+1.
$
\End
\li Pick a non-uniform random sample $\T$ of $t$ \label{scons_real}
points from $P$, \mbx{where the probability \zi that a point in $\T$ equals $p\in P$, is $m_p/\sum_{q\in P}m_q$}.
\li     \For each $p\in \T$ \Do
\zi \hspace{2cm}$\displaystyle
w(p)\gets \frac{\sum_q m_q}{|\T|\cdot m_p}.
$
\End
\li     \For each $b\in B$ \Do
\li \hspace{2cm}$\displaystyle
w(b)\gets (1+10\eps)|P_b|-\sum_{p\in \T\cap P_b}w(p).\label{wb}
$
\End
\li $\DD\gets \T\cup B$
\li \Return $(\DD,\T,w)$
\end{codebox}
\end{minipage}}}\caption{The algorithm~{\rkmedian}.\label{fig:kmedian} \mrx{Need to change as above also in body of paper}}
\end{figure*}

\subsection{Strong Coresets for Metric $k$-Median}

The following is a generalization of Theorem~\eqref{enettheorem}, as appeared in \cite{li00improved}.
Although the original claim uses another definition of dimensionality (analogous to the VC-dimension), it can be easily verified that it also holds for our weaker definition of dimensionality.
%In fact, we can use the simpler version of bins and balls from..

\begin{theorem}[\cite{li00improved}]\label{dvv}
Let $F$ be a set of functions from $X$ to $[0,1]$, and let $u,v,\delta>0$. Let $\prr:F\rightarrow [0,1]$ be a distribution on $F$.
Let $c$ be a sufficiently large constant.
Let $S$ be a non-uniform random sample of
\begin{equation*}%\label{sdef}
|S|=\frac{c}{u^2v}\big(\dim(F)\cdot \log(1/v)+\log(1/\delta)\big)
\end{equation*}
functions from $F$, where for every $s\in S$ and $f\in F$, we have $\pr(s=f)=\prr(f)$.
Then, with probability at least $1-\delta$,
\[
\forall x\in X: \frac{|\f(x)-\s(x)|}{\f(x)+\s(x)+v}\leq u,
\]
where $\f(x)=\sum_{f\in F}\prr(f)\cdot f(x)$, and $\s(x)=\sum_{f\in S}f(x)/|S|$.
\end{theorem}

\mbx{We start by proving a technical lemma regarding the weights defined in algorithm~{\rkmedian$(P,B,t,\eps)$}, see Fig.~\ref{fig:kmedian}.}

\begin{corollary}\label{corkmed}
Let $P,B$ be two finite sets of points in a metric space, and $0<\delta,\eps<1/2$. Let $c$ be the constant from Theorem~\ref{dvv}, and
\[
t\geq \frac{2c|B|}{\eps^2}\left(3\log|B|+\log(1/\delta)\right).
\]
Let $(\DD,w)$ be the pair that is returned from a call to the algorithm~{\rkmedian$(P,B,t,\eps)$}, see Fig.~\ref{fig:kmedian}. Then, with probability at least $1-\delta$, we have
\[
\forall p\in \DD: w(p)\geq 0.
\]
\end{corollary}
\begin{proof}
Let $u=\eps$ and $v=1/(2^{u/2}|B|)$.
Let $\T$ be the sample that is constructed during the execution of Line~\ref{scons_real} of the algorithm; see Fig.~\ref{fig:kmedian}.
Hence,

\mrx{Need to fix so below works out with $|B|=1$ - we discussed this by phone also.}

\begin{equation}\label{ssize}
\begin{split}
|\T|&=t\geq \frac{2c|B|}{\eps^2}\big(2\log(|B|)+\log(|B|/\delta)\big)\\
&\geq \frac{2c|B|}{\eps^2}\big(\log(2^{\eps/2}|B|)+\log(|B|/\delta)\big)\\
&\geq \frac{c}{u^2v}\big(\log(1/v))+\log(|B|/\delta)\big).
\end{split}
\end{equation}

For every $p\in P$, define $f_p:B\rightarrow [0,1]$ as
\[f_p(b)=\begin{cases}
\displaystyle\frac{|P|}{|B|\cdot |P_b|m_p }& p\in P_b\\
0 & p\not\in P_b
\end{cases}\enspace.
\]
Let $F=\br{f_p\mid p\in P}$ and $S=\br{f_p\in F\mid p\in \T}$. By the construction of $\T$, for every $f\in F$ and $s\in S$, we have $s=f_p$
with probability $\prr(p)=m_p/\sum_{q\in P} m_q$.
%Since every subset of $B$ is of size at most $|B|=2^{\log( |B|)}$, we have $\dim(F)\leq\log |B|$.
%Using~\eqref{ssize}, we apply Theorem~\ref{dvv} with $d=\log |B|$, and infer that, with probability at least $1-\delta$,
We apply Theorem~\ref{dvv} with $\delta/|B|$, $d=1$ and $X=\br{b}$ for some fixed $b\in B$, and infer that, with probability at least $1-\delta/|B|$,
\mbx{
\begin{equation}\label{fms}
\frac{|\f(b)-\s(b)|}{\f(b)+\s(b)+v}\leq u,
\end{equation}
}
where $\f(b)=\sum_{f_p\in F}\prr(p)f_p(b)$ and $\s(b)=\sum_{f\in S}f_p(b)/|S|$.
Assume that~\eqref{fms} holds for every for every $b\in B$, which happens with probability at least $1-\delta$.

By~\eqref{fms},
\[
\begin{split}
\s(b)&\leq \f(b)+u(\f(b)+\s(b)+v)\\
&=\f(b)(1+u)+u\s(b)+uv.
\end{split}
\]
That is, $\s(b)\leq \big(uv+\f(b)(1+u)\big)/(1-u)$. Since $u\leq \eps\leq 1/2$, we obtain
\[
\begin{split}
\s(b)&\leq \big(uv+\f(b)(1+u)\big)(1+2u)\\
&\leq  2uv+\f(b)(1+4u)
\leq \frac{2\eps}{|B|}+\f(b)\left(1+4\eps\right).
\end{split}
\]
We have \[
\f(b)=\sum_{f_p\in F}\prr(p)f_p(b)=\sum_{p\in P_b}\frac{1}{\sum_{q\in P}m_q}\cdot \frac{|P|}{|B|\cdot |P_b|}
=\frac{|P|}{|B|\sum_{q\in P}m_q}.
\]
By the last two inequalities,
\mbx{
\[
\s(b)\leq \frac{2\eps}{|B|}+\frac{|P|(1+4\eps)}{|B|\sum_{q\in P}m_q}.
\]
}
Since $1\leq 3|P|/\sum_{q\in P}m_q$ (\mbx{by the definition of $m_p$)}, we obtain
\begin{equation}\label{ineq}
\s(b)\leq \frac{2\eps\cdot 3|P|}{|B|\sum_{q\in P}m_q}+\frac{(1+4\eps)|P|}{|B|\sum_{q\in P}m_q} \leq \frac{(1+10\eps)|P|}{|B|\sum_{q\in P}m_q}.
\end{equation}

For every $p\in P_b$, we have \mbx{$w(p)=\sum_{q\in P}m_q/(|\T|\cdot m_p)$}, so
\[
\sum_{p\in \T\cap P_b}w(p)
=\frac{\sum_{q\in P}m_q}{|\T|}\sum_{p\in \T\cap P_b}\frac{1}{m_p}
=\frac{|B|\cdot |P_b|\sum_{q\in P}m_q}{|P|} \sum_{f\in S}\frac{f_p(b)}{|S|}
=\frac{|B|\cdot |P_b|\sum_{q\in P}m_q}{|P|}\cdot \s(b).
\]
By this and~\eqref{ineq},
\[
\sum_{p\in \T\cap P_b}w(p)\leq (1+10\eps)|P_b|
\]
Hence,
\[
w(b)=(1+10\eps)|P_b|-\sum_{p\in \T\cap P_b}w(p)
\geq 0.
\]
Together with the fact that $w(p)\geq 0$ for every $p\in \T$, we conclude that $w(p)\geq 0$ for every $p\in \T\cup B=\DD$.
\end{proof}

\mbx{We are now ready to address strong coresets for metric $k$-median.}

\begin{theorem}\label{metrickmed}
Let $(P,\dist)$ be a metric space of $n$ points. Let $0<\eps,\delta<1/2$, and
\[
\t=\frac{c}{\eps^2}\cdot \big(k\log n+\log(1/\delta)\big),
\]
where $c$ is a sufficiently large constant.
Then a set $\DD\subseteq P$, $|\DD|=t$, with a weight function $w:D\rightarrow [0,\infty)$ can be computed
 such that, with probability at least $1-\delta$,
\[
\forall x\in P^k: \left|\sum_{p\in P}\dist(p,x)-\sum_{p\in D}w(p)\dist(p,x)\right|\leq \eps \sum_{p\in P}\dist(p,x).
\]
The running time is \mbx{$O(nk +\log^2(1/\delta)\log^2 n+ k^2)$}.
\end{theorem}
\begin{proof}
By Theorem~\ref{constfactor}, a set $B\subseteq P$ of $k$ points can be computed in $O(nk)+(k+\log(2/\delta)\log n)^{2}$ time such that, with probability at least $1-\delta$,
\begin{equation}\label{PBMedian2}
\cost(P,B) \leq O(1)\min_{x\in P^k}\cost(P,x).
\end{equation}
Consider the set of functions $\KK(P)$; see Definition~\ref{GGdef3}.
Since $|P|=n$, we have $\dim(\KK(P))= O(\log n)$ for the case $k=1$.
Using Lemma~\ref{clustering}, $\dim(\KK(P))= O(k\log n)$ for any $k\geq 1$.

Let $(D,\T,w)$ be the output of a call to the algorithm~{\rkmedian$(P,B,t,\eps)$}.
By Corollary~\ref{corkmed}, with probability at least $1-\delta$, the weight function $w$ is non-negative. Assume that this event indeed occurs.
Let $(D',\T',w')$ be the output of a call to the algorithm~{\dimred$(P,B,\t,\eps)$}.
Since $\T$ and $\T'$ have the same distribution, we assume w.l.o.g. that $\T=\T'$.
\mrx{The above sentence is not very clear.}

\mrx{Changed $w'$ below to $w$:}

By Theorem~\ref{mainsub3}, with probability at least $1-\delta$,
\mbx{
\begin{equation}\label{SD2}
\begin{split}
&\forall x\in P^k:\\
&\left|\cost(P,x)-\left(\cost(\proj(P,B),x)+\sum_{p\in \T}w(p)\dist(p,x)-\sum_{p\in \T}w(p) \dist(\proj(p,B),x)\right) \right|\\
&\leq O(\eps) \cost(P,B).
\end{split}
\end{equation}
}
Assume that~\eqref{SD2} indeed holds.
Since $\proj(p,B)=b$ for every $p\in P_b$, we have

\begin{equation}\label{eqq}
\begin{split}
&  \mbx{\cost(\proj(P,B),x)+\sum_{p\in \T}w(p)\dist(p,x)-\sum_{p\in \T}w(p) \dist(\proj(p,B),x)}\\
&=\sum_{b\in B}\left(|P_b|-\sum_{p\in \T\cap P_b}w(p)\right) \cdot \dist(b,x)+\sum_{p\in \T}w(p)\dist(p,x)\\
&=\sum_{p\in D}w(p)\dist(p,x)-\sum_{b\in B}10\eps|P_b|\dist(b,x).
\end{split}
\end{equation}

For every $p\in P_b$, we have $\dist(b,x)\leq \dist(b,p)+\dist(p,x)\leq \dist(p,B)+\dist(p,x)$.
Summing over every $p\in P_b$ and $b\in B$ yields
\[
\sum_{b\in B}|P_b|\dist(b,x)\leq \cost(P,B)+\cost(P,x).
\]
Hence, \[
\sum_{b\in B}10\eps|P_b|\dist(b,x)\leq O(\eps)\cost(P,B)+O(\eps)\cost(P,x).
\]

Combining the last inequality with~\eqref{SD2} and~\eqref{eqq} yields
\[
\forall x\in P^k: \left|\cost(P,x)-\sum_{p\in D}w(p)\dist(p,x)\right|\leq O(\eps)\cost(P,B)+O(\eps) \cost(P,x)
\]
Given $B$, the set $\DD$ can be constructed in $O(nk)$ by taking multiple copies of each point and then use uniform random sampling; see Fig~\ref{fig:algA}.
\mrx{[Re you referring here to the implementation of random sampling? Maybe elaborate a bit.]}
By using~\eqref{PBMedian2} and a sufficiently large constant $c$, this proves the theorem.
\end{proof}

\subsection{Strong Coreset for Metric $k$-Means and Distances to the Power of $z$}
\begin{theorem}\label{metrickmean}
Let $(P,\dist)$ be a metric space of $n$ points. Let $k\geq1$ be an integer, $0<\eps,\delta<1/10$, and
\[
\t \geq \frac{c}{\eps^{2z}}\left( \dim(\KK(P),\XX) +k\log k+\log\frac{1}{\delta}\right),
\]
where $c$ is a sufficiently large constant.
Then a set $\DD\subseteq P$, $|\DD|=t$, with a weight function $w:D\rightarrow [0,\infty)$ can be computed such that, with probability at least $1-\delta$,
\[
\forall x\in P^k: \left|\sum_{p\in P}\dist^z(p,x)-\sum_{p\in D}w(p)\dist^z(p,x)\right|\leq \eps \sum_{p\in P}\dist^z(p,x).
\]
The running time is \mbx{$O(nk +\log^2(1/\delta)\log^2 n+ k^2)$}.
\end{theorem}
\begin{proof}
We construct a set $D$ and a weight function $w$ such that
\[
\forall x\in P^k: \left|\sum_{p\in P}\dist^z(p,x)-\sum_{p\in D}w(p)\dist^z(p,x)\right|\leq c\eps \sum_{p\in P}\dist^z(p,x),
\]
with probability at least $1-c\delta$. Replacing $\eps$ and $\delta$ in the proof with $\eps/c$ and $\delta/c$ respectively, would then prove the theorem.

By Theorem~\ref{constfactor}, a set $B\subseteq P$ of $k$ points can be computed in $O(nk)+(k+\log(2/\delta)\log n)^{2}$ time such that, with probability at least $1-\delta$,
\begin{equation}\label{PBMedian22}
\cost(P,B) \leq O(1)\min_{x\in P^k}\cost(P,x).
\end{equation}

Let $M(x)=\br{p\in P\mid |\dist^z(p,x)-\dist^z(p',x)| \leq \dist^z(p,B)/\eps^{z-1}}$.
\mbx{Let $\KK(\T)=\KK(\T,B,\eps/c)$} be defined as Definition~\ref{GGdef3}, where $\dist(\cdot,\cdot)$ is replaced by $\dist^z(\cdot,\cdot)$.
Let $X=\XX(\KK(\T))$.
For every $p\in P$, let $p'=\proj(p,B)$, and $h_p:X\rightarrow [0,\infty)$ be defined as
$$h_p(x)=\dist^z(p,x)-\dist^z(p',x)+\frac{3m_p\cdot\cost(P,B)}{\eps^{z-1}\sum_{q\in P} m_q}$$
if $p\in M(x)$, and $h_p(x)=0$ otherwise.
Let $H=\br{h_{p}\mid p\in P}$. Let $C$ be the output of a call to $\coresetA(H,\emptyset,s,m,\eps^z)$, where $s(h)=h$ for every $h\in H$; see Fig.~\ref{fig:algA}.

Let $G=\br{g_{h_p}\mid p\in P}$ be the set that is defined in Line~\ref{Gdef} of the above call to $\coresetA$. Note that for every $p\in \T$ we have
\begin{equation*}%\label{GT2}
 g_{h_p}(x)=\frac{h_p(x)}{m_p}.
\end{equation*}
We thus have $G=\KK(P)$, so $\dim(G,\XX)=\dim(\KK(P),\XX)$.
Let $S=\br{g_{h_p} \mid p\in \T}=\KK(\T)$.
By the construction of $\T$, we have that $S$ is a random sample of $\t$ i.i.d functions from $G$.
By Theorem~\ref{mycor2}, with probability at least $1-\delta$ we have that $S$ is an $\eps^z$-approximation of $G_{\ind \XX(S)}=G_{\ind X}$.
Assume that this event indeed occurs, and suppose that $S$ was used in Line~\ref{Sdef} of the above call to $\coresetA$.

Put $x\in X$. We start by proving that the functions $h_p$ are non-negative.
Namely, for $p\in M(x)$ and $\Delta_p =\frac{3m_p\cdot\cost(P,B)}{\eps^{z-1}\sum_{q\in P} m_q}$ we show that $|\dist^z(p,x)-\dist^z(p',x)| \leq \Delta_p$.
Since $p\in M(x)$ we have
\[
|\dist^z(p,x)-\dist^z(p',x)| \leq \frac{\dist^z(p,B)}{\eps^{z-1}}.
\]
Thus, it suffices to prove that
$$
\dist^z(p,B) \leq \frac{3m_p\cdot\cost(P,B)}{\sum_{q\in P} m_q}.
$$
Now,
\[
\begin{split}
&  \frac{3m_p\cdot\cost(P,B)}{\sum_{q\in P} m_q} \geq \frac{3\frac{|P|\dist^2(p,B)}{\cost(P,B)}\cdot\cost(P,B)}{\sum_{q\in P} \left(\frac{|P|\dist^2(p,B)}{\cost(P,B)}+1\right)} \\
& = \frac{3\left( \frac{|P|\dist^2(p,B)}{\cost(P,B)}\right)\cdot\cost(P,B)}{2|P|} \geq \dist^2(p,B).
 \end{split}
\]

For every $p\in M(x)$, let $f_p,f'_p:X\rightarrow [0,\infty)$ be defined as $f_p(x)=\dist^z(p,x)$ and $f'_p(x)=\dist^z(p',x)$. For $p\in P\setminus M(x)$ we define $f_p(x)=f'_p(x)=0$.
Let $F=\br{f_p\mid p\in P}$ and $F'=\br{f'_p\mid p\in P}$.
By Corollary~\ref{hh},
\begin{equation*}
\left|\cost(F,x)-\left(\cost(F',x)+\cost(C,x)-\sum_{p\in P}\Delta_p\right)\right|\leq 12\eps\cost(F,B).
\end{equation*}
It also holds that
$
\sum_{p\in P} \Delta_p = 3\cost(F,B)/\eps^{z-1}.
$
Thus,
\begin{equation*}
|\cost(F,x)-\left(\cost(F',x)-3\cost(F,B)/\eps^{z-1}+\cost(C,x)\right)|\leq 12\eps\cost(F,B).
\end{equation*}
We also have
\[
\begin{split}
\sum_{p\in \T}w(p)\cdot\frac{m_p\cost(P,B)}{\sum_{q\in P}m_q}=\cost(P,B)=\cost(F,B).
\end{split}
\]
and
\[
\begin{split}
\sum_{p\in \T\cap M(x)}w(p)\big(\dist^z(p,x)-\dist^z(\proj(p,B),x)\big)
&=\frac{\sum_{q\in P}m_q}{|S|}\sum_{p\in \T\cap M(x)}\frac{h_p(x)-\Delta_p}{m_p}\\
&=\frac{|G|}{|S|}\sum_{p\in \T\cap M(x)}\frac{h_p(x)}{m_p}-\frac{\sum_{q\in P}m_q}{|S|}\sum_{p\in \T}\frac{\Delta_p}{m_p}\\
\\&=\frac{|G|}{|S|}\sum_{g_{h_p}\in S\cap M(x)} g_{h_p}(x)
-\sum_{p\in \T\cap M(x)}w(p)\Delta_p
\\&=\cost(C,x)-\sum_{p\in \T\cap M(x)}w(p)\cdot\frac{3m_p\cost(P,B)}{\eps^{z-1}\sum_{q\in P}m_q}.
\end{split}\]
Using the last three inequalities,

\mrx{Need to elaborate derivation below - I wasn't able to verify (although I recall verifying this a while back.}
\begin{equation}\label{almostthere}
\begin{split}
&\left|\cost(M(x),x)-\left(\cost(\proj(M(x),B),x)+\sum_{p\in \T\cap M(x)}w(p)\big(\dist^z(p,x)-\dist^z(\proj(p,B),x)\big)\right)\right|
\\&=\left|\cost(F,x)-\left(\cost(F',x)+\cost(C,x)-\sum_{p\in \T\cap M(x)}w(p)\cdot\frac{3m_p\cost(P,B)}{\eps^{z-1}\sum_{q\in P}m_q}\right)\right|\\
&=|\cost(F,x)
-\left(\cost(F',x)+\cost(C,x)-3\cost(F,B)/\eps^{z-1}\right)|
\leq 12\eps\cost(F,B)=6\eps\cost(P,B).
\end{split}
\end{equation}

Put $p\in P\setminus M(x)$. That is,
\begin{equation}\label{myass}
|\dist^z(p,x)-\dist^z(p',x)|>\frac{\dist^z(p,p')}{\eps^{z-1}}.
\end{equation}
Hence, \[
\dist^z(p,x)+\dist^z(p',x)\geq |\dist^z(p,x)-\dist^z(p',x)|>\frac{\dist^z(p,p')}{\eps^{z-1}}.
\]
By.. we have \[
\dist^z(p,x)\leq 2^{z-1}(\dist^z(p',p)+\dist^z(p',x)).
\]
Using the last two inequalities,
\[
\dist^z(p,p')< \eps^{z-1}(\dist^z(p,x)+\dist^z(p',x))
\leq  \eps(2^{z-1}+1)\dist^z(p',x)+2^{z-1}\dist^z(p',p)).
\]
So,
\[
\dist^z(p',x)
\geq \frac{(1-2^{z-1}\eps)\dist^z(p,p')}{(2^{z-1}+1)\eps}
\geq \frac{\dist^z(p,B)}{2^z\eps}.
\]
Using the last inequality in Lemma~\ref{74}  yields
\[
|\dist^z(p,x)-\dist^z(p',x)|\leq \eps\dist^z(p,x).
\]
Hence,
\[
\begin{split}
|\dist^z(p,x)-\dist^z(p',x)|
&\leq \eps\dist^z(p,x)\\
&\leq \eps\dist^z(p',x)+\eps|\dist^z(p,x)-\dist^z(p',x)|.
\end{split}
\]
Thus,
\begin{equation}\label{youneed}
|\dist^z(p,x)-\dist^z(p',x)|
\leq \frac{\eps\dist^z(p',x)}{1-\eps}
\leq 2\eps\dist^2(p',x).
\end{equation}

Summing~\eqref{youneed} over $p\in P\setminus M(x)$ yields
\begin{equation}\label{youneed2}
\begin{split}
|\cost(P\setminus M(x),x)-\cost(\proj(P\setminus M(x),B),x)|
&\leq \sum_{p\in P\setminus M(x)}|\dist^z(p,x)-\dist^z(p',x)|\\
&\leq 2\eps\sum_{p\in P\setminus M(x)}\dist^z(p',x)\\
&\leq 2\eps\cost(\proj(P,B),x).
\end{split}
\end{equation}
Let $P_b=\br{p\in P\mid \proj(p,B)=b}$ for every $b\in B$.
Summing~\eqref{youneed} over every $p\in \T\setminus M(x)$ yields
\begin{equation}\label{it}
\begin{split}
\sum_{p\in \T\setminus M(x)}w(p)(\dist^z(p,x)-\dist^z(p',x))
&\leq 2\eps\sum_{p\in \T\setminus M(x)}w(p)\dist^z(p',x)\\
&= 2\eps\sum_{b\in B}\dist^z(b,x)\sum_{p\in (\T\setminus M(x))\cap P_b}w(p)\\
&\leq 2\eps\sum_{b\in B}\dist^z(b,x)\sum_{p\in \T\cap P_b}w(p).
\end{split}
\end{equation}
Using Corollary~\eqref{corkmed}, we have that,\footnote{In fact, we can use $\eps=1/2$ below and reduce the size of the resulting coreset if we are willing to have negative weights. In this case the term $k\log k$ will be outside the parenthesis. If we want only positive weights, then the $k\log k$ should be inside anyway.}  with probability at least $1-\delta$, $w(p)>0$ for every $p\in D$.
In particular, by Line~\ref{wb} of the algorithm~$\rkmedian$ (see Fig.~\ref{fig:kmedian}), for every $b\in B$ we have
\[
\sum_{p\in \T\cap P_b}w(p)\leq (1+\eps)|P_b|\leq 2|P_b|.
\]
Assume that the last inequality holds.
Combining it with~\eqref{it} yields
\[
\begin{split}
\sum_{p\in \T\setminus M(x)}w(p)(\dist^z(p,x)-\dist^2(p',x))
&\leq 4\eps\sum_{b\in B}\dist^z(b,x)|P_b|\\
&= 4\eps\cost(\proj(P,B),x).
\end{split}
\]

Combining the last inequality and~\eqref{youneed2} yields
\[
\begin{split}
&\left|\cost(P\setminus M(x),x)
-\cost(\proj(P\setminus M(x),B),x)
+\sum_{p\in \T\setminus M(x)}w(p)\big(\dist^z(p,x)-\dist^z(\proj(p,B),x)\big)\right|\\
&\leq
6\eps\cost(\proj(P,B),x).
\end{split}
\]
Together with~\eqref{almostthere}, we obtain
\[
\begin{split}
&|\cost(P,x)-\sum_{p\in D}w(p)\dist^z(p,x)|\\
&=\Big|\left(\cost(M(x),x)-\cost(\proj(M(x),B),x)-\sum_{p\in \T\cap M(x)}w(p)\dist^z(p,x)\right)\\
&\quad+\left(\cost(P\setminus M(x),x)-\cost(\proj(P\setminus M(x),B),x)
-\sum_{p\in \T\setminus M(x)}w(p)\dist^z(p,x)\right)
\Big|\\
&\leq
\left|\left(\cost(M(x),x)-\cost(\proj(M(x),B),x)
-\sum_{p\in \T\cap M(x)}w(p)\dist^z(p,x)\right)\right|\\
&\quad+\left|\left(\cost(P\setminus M(x),x)-\cost(\proj(P\setminus M(x),B),x)
-\sum_{p\in \T\setminus M(x)}w(p)\dist^z(p,x)\right)\right|\\
&\leq 18\eps\cost(\proj(P,B),x).
\end{split}
\]
Plugging~\eqref{PBMedian22} in the last inequality then proves the theorem.
\end{proof}

%By Corollary~\ref{15.3}, we have $\forall p\in D$. In particular, for every $b\in B\subseteq D$, we have, by Line~\ref{} of $k-mediancoreset'$,
%\[
%(1+10\eps)|P_b|-\sum_{p\in S\cap P_b}w(p)\geq 0.
%\]
%
%We prove that $\sum_{p\in Q\cap D}w(p)|\dist^2(p,C)-\dist^2(p',C)|\leq \eps\cost(P,C)$.
%\[
%w(p)|\dist^2(p,C)-\dist^2(p',C)|\leq w(p)\dist^2(p,C)
%\leq 2w(p)\dist^2(p,p')+ 2w(p)\dist^2(p',C).
%\]
%
%\[
%\sum_{p\in Q\cap D}w(p)\dist^2(p,p')\leq \sum_{p\in D}w(p)\dist^2(p,p')\leq \eps\cost(P,C^*).
%\]
%\[
%\sum_{p\in Q\cap D}w(p)\dist^2(p',C)
%\leq \sum_{p\in D}w(p)\dist^2(p',C)
%= \sum_i \dist^2(c^*_i,C)|P_i\cap D|
%\leq \sum_i \dist^2(c^*_i,C)n_i
%\leq 2opt_i
%\]
%
%\end{proof}

\section{$k$-Median in $\REAL^d$}

\mbx{In the upcoming section we address the special case of $k$-median in $\REAL^d$.}

\label{sec:kminrd}
\subsection{Strong Coresets}
\mbx{We start by stating a technical lemma addressing arrangements of balls in $\REAL^d$.}
%Proof appears implicitly in \cite{SA} and follows from the fact that every cell in an arrangement of balls \mb{in $\REAL^d$} corresponds to an intersection of a subset of at most $O(d)$ balls, where each cell corresponds to a different subset of balls.}

\begin{lemma}\label{kballs1}
Let $P$ be a set of points in $\REAL^d$, and let $c\in\REAL$. For every $p\in P$, let $b:P\rightarrow\REAL^d$ be a mapping from every $p\in P$ to a point $p'=b(p)$. For every $p\in P$, let $f_p:X(j,1)\rightarrow [0,\infty)$ be defined as $f_p(x)=\dist(p,x)-\dist(p',x)+c$.  Then the dimension of $F=\br{f_p\mid p\in P}$ is $O(d(j+1))$.
\end{lemma}
\begin{proof}
Put $S\subseteq P$. For every $x\in X(j,1)$ and $r \in\REAL$, let \[
\range(x,r)=\br{p\in S\mid \dist(p,x)-\dist(p',x)\leq r-c}.
\]
Let $R^{+}=\br{\range(x,r)\mid x\in X(j,1), r-c\geq 0}$, and $R^{-}=\br{\range(x,r) \mid x\in X(j,1), r-c<0}$.
We have
\begin{equation}
\label{rangebound}
\begin{split}
|\br{\range(x,r)\mid x\in X(j,1), r\geq 0}|
&\leq |\br{\range(x,r)\mid x\in X(j,1), r\in\REAL}|\\
&\leq |R^{+}|+|R^{-}|.
\end{split}
\end{equation}
We now bound $|R^{+}|$ and then $|R^{-}|$.

Let $r\in \REAL$, $p\in S$ and $x\in X(j,1)$.
We define $\dist^2(p,x):=\big(\dist(p,x)\big)^2$.
Since $x$ is a $j$-flat, there is a tuple of $j+1$ vectors $h_0,\cdots, h_{j}\in\REAL^d$ such that
$x=\br{h_0+\sum_{i=1}^j a_ih_j\mid a_1,\ldots,a_j\in\REAL}$, and
\[
\begin{split}
\dist^2(p,x)&=\norm{p-h_0}_2^2-\sum_{i=1}^{j} \big((p-h_0)^T h_i\big)^2\\
&=\norm{p-h_0}_2^2-\sum_{i=1}^{j} (p^Th_i-h_0^T h_i)^2.
\end{split}
\]
For two vectors $(m_1,\cdots, m_s)\in\REAL^{s}$ and $(y_1,\cdots, y_t)\in\REAL^{t}$, we denote by $my$ the tuple $m_1,\cdots,m_s,y_1,\cdots,y_t$. Let $h=(1,r-c,h_0\cdots h_j)\in \REAL^{d(j+1)+2}$, and $q=(1,p p')\in \REAL^{2d+1}$ where $p'=b(p)$.
Hence, we have
\begin{equation}
\label{eq:sum1}
\dist^2(p,x)-\dist^2(p',x)-(r-c)^2=\sum_{i_0,i_1\in [2d+1], i_2,i_3\in [d(j+1)+2]} c_{i_0,i_1,i_2,i_3}q_{i_0} q_{i_1}h_{i_2} h_{i_3},
\end{equation}
where $c_{i_0,i_1,i_2,i_3}$ is a constant that depends only on $i_0,\ldots, i_3$, and equals to zero for all except $d_1=O(d(j+1))$ terms of the summation.
\mrx{This is hard to follow ... It seems that the expression above includes terms $h_ih_jh_k$ (namely a power of three) ... Also, I do not follow the size of $d_1$ ... Maybe expand this part accordingly.}
Equation~\eqref{eq:sum1} implies that there are two $d_1$-dimensional vectors $u_1=u_1(p)$ and $v_1=v_1(x,r,c)$, such that \begin{equation}
\label{eq:utv3}
u_1^Tv_1>0 \Leftrightarrow \dist^2(p,x)-\dist^2(p',x)-(r-c)^2>0.
\end{equation}

Similarly,
\[
\begin{split}
&\big(\dist^2(p,x)-\dist^2(p',x)-(r-c)^2\big)^2- \big(2(r-c)\dist(p',x)\big)^2\\
&=\sum_{i_0,\ldots,i_3\in [2d+1], i_4,\cdots,i_7\in [d(j+1)+2]} c'_{i_0,\ldots,i_7}q_{i_0}\cdots q_{i_3}h_{i_4}\cdots h_{i_7},
\end{split}\]
where $c'_{i_0,\ldots,i_7}$ is a constant that depends only on $i_0,\ldots, i_7$ and equals to zero for all except $d_2=O(d(j+1))$ terms.
Hence, there are two $d_2$-dimensional vectors, $u_2=u_2(p)$ and $v_2=v_2(x,r)$, such that
\begin{equation}
\label{eq:utv2}
u_2^Tv_2>0 \Leftrightarrow \big(\dist^2(p,x)-\dist^2(p',x)-(r-c)^2\big)^2- \big(2(r-c)\dist(p',x)\big)^2 >0.
\end{equation}

Let $o_1=(0,\cdots,0)\in\REAL^{d_1}$, $o_2=(0,\cdots,0)\in\REAL^{d_2}$, $u=u(p)=(u_1u_2), v=v(x,r)=(v_1 o_2)$, and  $z=z(x,r)=(o_1v_2)\in\REAL^{d_1+d_2}$. By~\eqref{eq:utv3} we have
\begin{equation}
\label{eq:utv1}
u^Tv> 0 \Leftrightarrow \dist^2(p,x)-\dist^2(p',x)-(r-c)^2>0,
\end{equation}
and by~\eqref{eq:utv2}
\begin{equation}
\label{eq:utv5}
u^Tz> 0 \Leftrightarrow \big(\dist^2(p,x)-\dist^2(p',x)-(r-c)^2)^2- (2(r-c)\dist(p',x)\big)^2> 0.
\end{equation}

Suppose that $\range(x,r)\in R^{+}$. We now prove that
\begin{equation}\label{ps3}
p\in \range(x,r) \Leftrightarrow \big(u^Tv\leq 0 \text{ or } u^Tz\leq 0\big).
\end{equation}
Indeed, since $r-c\geq 0$,
\begin{equation}
\label{eq:px1}
\begin{split}
p\in \range(x,r) &\Leftrightarrow
\dist(p,x)-\dist(p',x)\leq r-c \\
&\Leftrightarrow
\dist(p,x)\leq r-c+\dist(p',x)\\
&\Leftrightarrow \dist^2(p,x)\leq (r-c)^2+\dist^2(p',x)+2(r-c)\dist(p',x)\\
&\Leftrightarrow \dist^2(p,x)-\dist^2(p',x)-(r-c)^2\leq 2(r-c)\dist(p',x).
\end{split}
\end{equation}
By~\eqref{eq:utv1},
\[
\begin{split}
&\big(u^Tv>0 \text{ and }\dist^2(p,x)-\dist^2(p',x)-(r-c)^2\leq 2(r-c)\dist(p',x)\big)\\
&\Leftrightarrow \big(u^Tv>0 \text{ and } (\dist^2(p,x)-\dist^2(p',x)-(r-c)^2)^2\leq (2(r-c)\dist(p',x))^2\big)\\
&\Leftrightarrow \big(u^Tv>0 \text{ and } u^Tz\leq 0 \big),
\end{split}
\]
where the last deviation is by~\eqref{eq:utv5}.
By the last equation and~\eqref{eq:px1},
\begin{equation}\label{pq22}
\big(u^Tv>0 \text{ and } p\in \range(x,r)\big) \Leftrightarrow \big(u^Tv>0 \text{ and } u^Tz\leq 0\big).
\end{equation}

We have by~\eqref{eq:utv1}
\[
\begin{split}
&u^Tv\leq 0 \Rightarrow \\
&\dist^2(p,x)-\dist^2(p',x)-(r-c)^2\leq 0 \leq 2(r-c)\dist(p',x).
\end{split}
\]
Combining this with~\eqref{eq:px1} yields $u^Tv\leq 0 \Rightarrow p\in \range(x,r)$. Using the last equation with~\eqref{pq22} proves~\eqref{ps3}.

Let $U=\br{u(p)\mid p\in S}\subseteq \REAL^{d_1+d_2}$. For every $v,z\in\REAL^{d_1+d_2}$, let \[\range'(v,z)=\br{u\in U\mid u^Tv\leq 0 \text{ or } u^Tz\leq 0}.\]
By~\eqref{ps3}, $\range(x,r)=\range'(v(x,r),z(x,r))$.
Hence,
\[
|R^+|=|\br{\range(x,r)\mid x\in X(j,1), r-c\geq 0}|
\leq \big|\br{\range'(v,z)\mid v,z\in\REAL^{d_1+d_2}}\big|
\]
It is not hard to verify that \[
|\br{\range'(v,z)\mid v,z\in\REAL^{d_1+d_2}}|\leq |U|^{O(d_1+d_2)}=|S|^{O(d(j+1))}.
\]
Combining the last two equations yields
\begin{equation}
\label{eq:S}
|R^+|\leq |S|^{O(d(j+1))}.
\end{equation}

We now bound $|R^-|$ in a similar way. We have
\begin{equation}
\label{rminus}
\begin{split}
|R^-|
&=|\br{\range(x,r)\mid x\in X(j,1), r-c<0}|\\
&=|\br{\br{p\in S\mid \dist(p,x)-\dist(p',x)\leq r-c} \mid x\in X(j,1), r-c<0}|\\
&=|\br{\br{p\in S\mid \dist(p',x)-\dist(p,x)\geq |r-c|} \mid x\in X(j,1), r-c<0}|\\
&=|\br{\br{p\in S\mid \dist(p',x)-\dist(p,x)\geq r-c} \mid x\in X(j,1), r-c>0}|\\
%&\leq |\br{\br{p\in S\mid \dist(p',x)-\dist(p,x)\geq r-c} \mid x\in X, r-c\geq 0}|.
\end{split}
\end{equation}
For every $r\in\REAL$ and a set $Q=\br{p\in S\mid \dist(p',x)-\dist(p,x)\geq r-c}$
there is a corresponding distinct set $S\setminus Q=\br{p\in S\mid \dist(p',x)-\dist(p,x)< r-c}$.
Hence,
\begin{equation}
\label{eq:beginn}
\begin{split}
&|\br{\br{p\in S\mid \dist(p',x)-\dist(p,x)\geq r-c} \mid x\in X(j,1), r-c> 0}|\\
&\leq |\br{\br{p\in S\mid \dist(p',x)-\dist(p,x)< r-c} \mid x\in X(j,1), r-c> 0}|\\
&\leq |\br{\br{p\in S\mid \dist(p',x)-\dist(p,x)\leq  r-c} \mid x\in X(j,1), r-c\geq 0}|.
\end{split}
\end{equation}
By replacing $p$ with $p'$ and $\range(x,r)$ with $\br{p\in S\mid \dist(p',x)-\dist(p,x)\leq r-c}$ in the proof of~\eqref{ps3}, we can bound the last term of~\eqref{eq:beginn} by $|S|^{O(d(j+1))}$. Together with~\eqref{rminus}, we obtain $|R^-|\leq |S|^{O(dj)}$.

Plugging the last equation and~\eqref{eq:S} in~\eqref{rangebound} yields
\[
|\br{\range(x,r)\mid x\in X, r\geq 0}|
\leq |R^+|+|R^-|\leq |S|^{O(d(j+1))}.
\]
Since the last inequality holds for any $S\subseteq P$, the dimension of $\br{f_p| p\in P}$ is $O(d(j+1))$.
\end{proof}

The following lemma follows from the fact that every cell in an arrangement of balls \mbx{in $\REAL^d$} corresponds to a different intersection of at most $O(d)$ balls; see~\cite{SA}.
\begin{lemma}\label{arr}
Let $A$ be the arrangement of a set of $n$ open ball in $\REAL^d$. There is a set $V\subset\REAL^d$, $|V|\leq n^{O(d)}$, that intersects every vertex, edge, face and cell of $A$.
\end{lemma}

Recall that $X(j,k)$ was defined in Section~\ref{notation} to be all the possible $k$-tuples of $j$-flats in $\REAL^d$.

\newcommand{\nrange}{\overline{\range}}
\begin{lemma}\label{kballs}
Let $P$ be a set of points in $\REAL^d$, and $k,j\geq 1$. For every $p\in P$, let $s_p,c_p,z_p\geq 0$ and define $g_p:X(j,k)\rightarrow [0,\infty)$ as
\[g_p=\begin{cases}
c_p\dist(p,x) & z_p < \dist(p,x)< s_p\\
0 & \text{otherwise},
\end{cases}\enspace
\]
and let $G=\br{g_p\mid p\in P}$. Then $\dim(G)=O(djk)$.
\end{lemma}
\begin{proof}
We prove the lemma for the case $k=1$. The case $k\geq 1$ then follows from Lemma~\ref{clustering}.
Put $S\subseteq P$. For every $x\in X(j,1)$ and $r\geq 0$, let
\begin{equation}\label{ee}
\begin{split}
\range(x,r)
&=\br{p\in S\mid g_p(x)\leq r }\\
&=\br{p\in S \mid \dist^2(p,x)- s_p^2\geq 0 \text{ or } \dist^2(p,x)-z_p^2\leq 0  \text{ or }  \dist(p,x)^2-r^2/c_p^2\leq 0}.
\end{split}
\end{equation}

Let $r\geq 0$, $p\in S$, and $x\in X(j,1)$.
Since $x$ is a $j$-flat, there is a tuple of $j+1$ vectors $h_0,\cdots, h_{j}\in\REAL^d$ such that
$x=\br{h_0+\sum_{i=1}^j a_ih_j\mid a_1,\ldots,a_j\in\REAL}$, and
\[
\begin{split}
\dist^2(p,x)- r^2/c_p^2&=\norm{p-h_0}_2^2-\sum_{i=1}^{j} \big((p-h_0)^T h_i\big)^2- r^2/c_p^2\\
&=\norm{p-h_0}_2^2-\sum_{i=1}^{j} (p^Th_i-h_0^T h_i)^2- r^2/c_p^2.
\end{split}
\]
For two vectors $(m_1,\cdots, m_s)\in\REAL^{s}$ and $(y_1,\cdots, y_t)\in\REAL^{t}$, we denote by $my$ the tuple $m_1,\cdots,m_s,y_1,\cdots,y_t$. Let $h'=(1,r^2/c_p^2,h_0\cdots h_j)\in \REAL^{d(j+1)+2}$, and $p'=(1,p)\in \REAL^{d+1}$.
Hence,
\begin{equation}
\label{eq:sum}
\dist^2(p,x)- r^2/c_p^2=\sum_{i_0,i_1\in [d+1], i_2,i_3\in [d(j+1)+2]} c_{i_0,i_1,i_2,i_3}p'_{i_0} p'_{i_1}h'_{i_2} h'_{i_3},
\end{equation}
where $c_{i_0,i_1,i_2,i_3}$ is a constant that depends only on $i_0,\ldots, i_3$, and equals to zero for all except $d_1=O(d(j+1))$ terms of the summation.

Equation~\eqref{eq:sum} implies that there are two $d_1$-dimensional vectors $u_1=u_1(p)$ and $v_1=v_1(x,r^2/c_p^2)$, such that \begin{equation}
\label{eq:utv31}
u_1^Tv_1\leq 0 \Leftrightarrow  \dist(p,x)^2-r^2/c_p^2\leq 0.
\end{equation}
Similarly, we can prove that there are two $d_1$-dimensional vectors $u_2=u_2(p)$ and $v_2=v_2(x,z_p^2)$, such that \begin{equation}
\label{eq:utv32}
u_2^Tv_2\leq 0 \Leftrightarrow  \dist(p,x)^2-z_p^2\leq 0,
\end{equation}
and that there are two $d_1$-dimensional vectors $u_3=u_3(p)$ and $v_3=v_3(x,s_p^2)$.
\begin{equation}
\label{eq:utv33}
u_3^Tv_3\geq 0 \Leftrightarrow  \dist(p,x)^2-s_p^2\geq 0,
\end{equation}

Let $o=(0,\cdots,0)\in\REAL^{d_1}$. Let $u=u(p)=(u_1u_2u_3), z_1=z_1(x,r)=(v_1 o \,o)$,
$z_2=z_2(x,r)=(o \,v_2 \,o)$,
$z_3=z_3(x,r)=(o\, o \,v_3)$ be vectors in $\REAL^{3d_1}$.
By~\eqref{ee},~\eqref{eq:utv31},~\eqref{eq:utv32} and~\eqref{eq:utv33}
\begin{equation}\label{ps31}
p\in \range(x,r) \Leftrightarrow \big(u^Tz_1\leq 0 \text{ or } u^Tz_2\leq 0  \text{ or } u^Tz_3\geq 0 \big).
\end{equation}

Let $U=\br{u(p)\mid p\in S}$. For every $z_1,z_2,z_3\in\REAL^{3d_1}$ let
\[
\range'(z_1,z_2,z_3)=
\br{u\in U\mid u^Tz_1\leq 0 \text{ or } u^Tz_2\leq 0 \text{ or }u^Tz_3\geq 0}.
\]
It is not hard to verify that
\begin{equation}
\label{eq:zz1}
|\br{\range'(z_1,z_2,z_3)\mid z_1,z_2,z_3\in\REAL^{3d_1}}|\leq |U|^{O(3d_1)}=|S|^{O(d(j+1))}.
\end{equation}
By~\eqref{ps31}, $\range(x,r)=\range'(z_1,z_2,z_3)$. Hence,
\[
|\br{\range(x,r)\mid x\in X(j,1), r\geq 0}|
\leq |\br{\range(z_1,z_2,z_3)\mid z_1,z_2,z_3\in \REAL^{3d_1}}|
\]

Using the last equation with~\eqref{eq:zz1} yields
\[
|\br{\range(x,r)\mid x\in X, r\geq 0}|\leq |S|^{O(d(j+1))}.
\]
Since the last inequality holds for any $S\subseteq P$, the dimension of $\br{f_p| p\in P}$ is $O(d(j+1))$.
\end{proof}

\begin{theorem}[strong coresets for $k$-median in $\REAL^d$]
Let $P$ be a set of $n$ points in $\REAL^d$. Let $k\geq 1$ be an integer, $0<\eps,\delta<1/2$, and $c$ be a sufficiently large constant.
Then, a set $\DD\subseteq P$ and a function $w:D\rightarrow (-\infty,\infty)$, can be computed
 such that, with probability at least $1-\delta$,
\[
\forall x\in (\REAL^d)^k: \left|\sum_{p\in P}\dist(p,x)-\sum_{p\in D}w(p)\dist(p,x)\right|\leq \eps \sum_{p\in P}\dist(p,x).
\]
The construction time of $\DD$ is \mbx{$O(ndk+\log^2(1/\delta)\log^2 n+\DD)$}, where either one of the following holds:
\begin{enumerate}
\item The size of $\DD$ is
\[
\frac{c}{\eps^2}\cdot \big(k+\log(1/\delta)\big)
\]
and $w(p)$ may be negative for some $p\in \DD$.
\item The size of $\DD$ is
\[
\frac{c}{\eps^2}\cdot \big(k\min\br{d,\log k}+\log(1/\delta)\big),
\]
and $w(p)> 0$ for every $p\in \DD$.
\end{enumerate}
\end{theorem}

\mrx{Need to verify proof in detail.}

\begin{proof}
\begin{enumerate}
\item The proof is the same as the proof of Theorem~\ref{metrickmed}, except for the computation of $\dim(\KK(P))$.
In this case, we have $\dim(\KK(P))=O(kd)$ instead of $\dim(\KK(P))=O(k\log n)$, as proved in Lemma~\ref{kballs}.
\item Lemma~\ref{corkmed} requires that $t\geq k\log k$ for fixed $\eps$ and $\delta$, and together with the bound on $\dim(\KK(P))$ we need $t\geq k\min\br{\log k,d}$.
    \end{enumerate}
\end{proof}

\section{$k$-Line Median}
\label{sec:kline}

\mrx{Need to add introductory text.}

\begin{theorem}[Strong coreset for $k$-lines in $\REAL^d$]
Let $P\subseteq \REAL^d$, $k\geq 1$, $0<\eps,\delta \leq 1/2$,  $r=k+\log(1/\delta)$ and
\[
t\geq \frac{c}{\eps^2} \left(dk+\log\frac{1}{\delta}\right),
\]
for a sufficiently large constant $c$. A set $D$ of $O(\t)+((1/\eps)\log n)^{O(k)}$ points and a weight function $w:D\rightarrow [-\infty,\infty)$ can be computed in
\mbx{$O(ndk)+O(dt^2)+t^{O(k)}\log^2 n$} time, such that, with probability at least $1-\delta$, for every set $x$ of $k$ lines in $\REAL^d$,
\[
\left|\sum_{p\in P}\dist(p,x)-\sum_{p\in D}w(p)\dist(p,x)\right|\leq \eps \sum_{p\in P}\dist(p,x).
\]
\end{theorem}
\begin{proof}
Let $r=k+\log\frac{1}{\delta}$. By Theorem~\ref{smalljk}, a set $B$ of $O(k\log n$) lines that satisfies
\begin{equation}\label{BBC}
\cost(P,B)\leq O(1)\min_{x^*\in X(2,k)}\cost(P,x^*)
\end{equation}
can be computed, with probability at least $1-\delta$, in time
\[
\bicriteriatime = O(ndk)+O(dr^2)+r^{O(k)}\log^2 n.
\]
%\mr{What is $r$? There is a clash here between $\eps^2$ and $\eps^4$.}
Assume that this event indeed occurs.

Let $(D',\T,w')$ be the output of a call to the algorithm~{\dimred$(P,B,\t,\eps/c)$}.
For every $S\subseteq \KK(P)$, let $\XX(S)=X(2,k)$ denote all the possible lines in $\REAL^d$.
By Lemma~\ref{kballs}, we have that $\dim(\KK(P),\XX)=O(dk)$.
By Theorem~\ref{mainsub3}, with probability at least $1-\delta$,
\begin{equation}\label{SD3}
\begin{split}
&\forall x\in X(2,k):\\
&\left|\cost(P,x)-\left(\cost(\proj(P,B),x)+\sum_{p\in \T}w'(p)\dist(p,x)-\sum_{p\in \T}w'(p) \dist(\proj(p,B),x)\right) \right|\\
&\leq \eps \cost(P,B).
\end{split}
\end{equation}

Using the result from~\cite{FFS06}, a set $C$, $|C|=|B|\cdot ((1/\eps)\log n)^{O(k)}$, with a weight function $u:C\rightarrow [0,\infty)$ can be constructed in $O(ndk)$ time such that
\[
\forall x\in X(2,k)\ \ |\cost(\proj(P,B),x)-\sum_{p\in C}u(p)\dist(p,x)|\leq \eps\cost(\proj(P,B),x).
\]
We have $\dist(\proj(p,B),x)\leq \dist(\proj(p,B),p)+\dist(p,x)=\dist(p,B)+\dist(p,x)$ for every $p\in P$. Summing over every $p\in P$, yields
$\cost(\proj(P,B),x)\leq \cost(P,B)+\cost(P,x)$.
Hence,
\begin{equation}
\label{eq:begg}
\forall x\in X(2,k)\ \ |\cost(\proj(P,B),x)-\sum_{p\in C}u(p)\dist(p,x)|\leq O(\eps)\cost(P,B)+O(\eps)\cost(P,x).
\end{equation}

%\mr{Need to read new version.}

Let $D=C\cup \T\cup \proj(\T,B)$ and
\[
w(p)=\begin{cases}
u(p)  & p\in C\\
w'(p) & p\in \T\\
-w'(p) & p\in \proj(\T,B)\\
\end{cases}.
\]
Using the triangle inequality,
\[
\begin{split}
&\forall x\in X(2,k):
\left|\sum_{p\in P}\dist(p,x)-\sum_{p\in D}w(p)\dist(p,x)\right|\\
&=\left|\cost(P,x)-
\left(\sum_{p\in C}u(p)\dist(p,x)+\sum_{p\in \T}w'(p)\dist(p,x)-\sum_{p\in \proj(\T,B)}w'(p)\dist(p,x)  \right)\right|\\
&\leq \left|\cost(P,x)-
\left(\cost(\proj(P,B),x)+\sum_{p\in \T}w'(p)\dist(p,x)-\sum_{p\in \T}w'(p) \dist(\proj(p,B),x)\right) \right|\\
&\quad+\left|\cost(\proj(P,B),x)-\sum_{p\in C}u(p)\dist(p,x)
\right|\\
\end{split}
\]
Together with~\eqref{BBC},~\eqref{SD3} and~\eqref{eq:begg} this proves the theorem as
\[
\begin{split}
&\forall x\in X(2,k):
\left|\sum_{p\in P}\dist(p,x)-\sum_{p\in D}w(p)\dist(p,x)\right|\\
&\leq \eps\cost(P,B)
+ O(\eps)\cost(P,B)+O(\eps)\cost(P,x)
\\& \leq O(\eps)\cost(P,x).
\end{split}
\]
\end{proof}

%\section{Projective Clustering (todo)}
%\begin{corollary}[strong coreset for projective clustering]\label{corrweak}
%Let $P\subseteq \REAL^d$, $j,k\geq 1$, $0<\eps,\delta \leq 1/2$, and
%\[
%t\geq \frac{c}{\eps^4} \left(\frac{j^2k\log(1/\eps)}{\eps}+\log\frac{1}{\delta}\right),
%\]
%for some sufficiently large constant $c$.
%Then a set $\DD\subseteq P$, $|\DD|=t$, with a weight function $w:D\rightarrow [0,\infty)$ can be computed
% such that, with probability at least $1-\delta$, for every set $x$ of $k$ $j$-flats we have
%\[
%\left|\sum_{p\in P}\dist(p,x)-\sum_{p\in D}w(p)\dist(p,x)\right|\leq \eps \sum_{p\in P}\dist(p,x).
%\]
%The running time is $..$.
%\end{corollary}

\section{$B$-Coresets for Projective Clustering}\label{sec:weak}
\mrx{Need to add introductory text}

\begin{definition}\label{ZZ}
For a set $P$ of points in $\REAL^d$ and an integer $j\geq 1$, we define $\ZZ(P,j)$ to be the set of all the possible $j$-flats that are spanned by at most $10j\log(1/\eps)/\eps$ points from $P$. For an integer $k\geq1$, we define $\ZZ(P,j,k)=(\ZZ(P,j))^k$.
\end{definition}

\begin{lemma}\label{klemma}
Let $k$,$j$, $G$,  $z_p$, $s_p$ and $c_p$ be defined as in Lemma~\ref{kballs}.
%\mr{[To define $G$ we also need to define $z_p,s_p,c_p$ (if we are using Lemma 16.3). Danny: fixed]}
For every set $S\subseteq G$ and its corresponding set $\T\subseteq P$, let $\XX(S)=\ZZ(\T,j,k)$. Then $\dim(G,\XX)=O(kj^2\log(1/\eps)/\eps)$.
\end{lemma}
\begin{proof}
Follows from the proof of Lemma~\ref{highd} with $m=10j\log(1/\eps)/\eps$, where the usage of Lemma~\ref{centfms}(i) is replaced by Lemma~\ref{kballs}.
Notice that replaceing Lemma~\ref{centfms}(i) by Lemma~\ref{kballs} adds a multiplicative factor of $j$ to the asserted dimension.
\end{proof}

\mrx{Add introductory text}

\mrx{Mike: need to re-read lemma below - did not add remarks.}
\begin{lemma}\label{mylem}
Let $P, B\subseteq \REAL^d$, $j\geq 1$, $0<\eps\leq 1/10$, and
\[
t\geq \frac{c}{\eps^4} \left(\frac{j^2k\log(1/\eps)}{\eps}+\log\frac{1}{\delta}\right),
\]
for some sufficiently large constant $c$.
Suppose that $\cost(P,B)>0$, and let $(D,\T,w)$ be the output of \mbx{$\dimred(P,B,t,\eps)$}.
\mr{Mike: we need to pass $z$ to $\dimred$ also, maybe just say $z=1$.}
Let $x^*=(x_1,\cdots,x_k)\in X(j,k)$ and $(P_1,\cdots,P_k)$ be a partition of $P$ such that $P_i=\br{p\in P\mid \dist(p,x_i)=\dist(p,x^*)}$ for $1\leq i\leq k$. Then, with probability at least $1-\delta$, for every $x\in \ZZ(D,j,k)$ that satisfies
\begin{equation}\label{costpt2}
\cost(P,x)> (1+2\eps)\cost(P,x^*)+8\eps \cost(P,B),
\end{equation}
there is $i\in [k]$ and $p\in (\T\cap P_i)\cup \proj(P_i,B)$ such that \[\dist(p,x)>(1+\eps)\dist(p,x^*)+\eps\cdot\frac{\cost(P_i,x^*)+\cost(P_i,B)}{|P_i|}.\]
\end{lemma}
\begin{proof}
Put $c_1=10$ and $c_2=2$. For each $i\in [k]$, let
\[
e_i:=c_2\eps\cdot\frac{\cost(P_i,x^*)+\cost(P_i,B)}{|P_i|},
\]
and for each $p\in P_i$, let $f_p:X(j,k)\rightarrow [0,\infty)$ be defined as:
\[
f_p(x)=\begin{cases}
\dist(p,x) & \dist(p,x)> (1+c_1\eps)\dist(p,x^*)+e_i\\
0 & \text{otherwise}.
\end{cases}\enspace
\]
and $f'_p:X(j,k)\rightarrow [0,\infty)$ be defined as:
\[f'_p(x)=\begin{cases}
\dist(\proj(p,B),x) & \dist(\proj(p,B),x)> (1+\eps)\dist(\proj(p,B),x^*)+e_i/c_2\\
0 & \text{otherwise}.
\end{cases}\enspace
\]
Let $s_{f_p}(x)=\dist(p,B)$, $F=\br{f_p\mid p\in P}$, $F'=\br{f'_p\mid p\in P}$ and
\[
m_f=m_p=\left\lceil \frac{|P|\dist(p,B)}{\cost(P,B)}\right\rceil +1.
\]

Fix $x\in \XX(D)$ such that~\eqref{costpt2} holds, $i\in[k]$, $p\in P_i$, and let $M(x)=\br{f_p\in F: f_p(x)\leq s_{f_p}(x)}$.
We now prove that
\begin{equation}\label{i}
f_p\in F\setminus M(x) \quad \Longrightarrow \quad \quad |f_p(x)-f'_p(x)|< \eps f_p(x).
\end{equation}
Indeed, if $f_p\in F\setminus M(x)$ then $f_p(x)>s_f(x)\geq 0$, so
\begin{equation}
\label{fii}
\dist(p,x)=f_p(x)>(1+c_1\eps)\dist(p,x^*)+e_i,
\end{equation}
and
\begin{equation}
\label{mme}
\dist(p,B)=\eps s_{f_p}(x)<\eps f_p(x)=\eps\dist(p,x).
\end{equation}
%By~\eqref{fii},~\eqref{fii2} and the assumption $\eps\leq 1/c_1$, we have
By the last two inequalities and the assumption $\eps\leq 1/c_1$, we have
\[
\begin{split}
\dist(p,x)
&> (1+c_1\eps)\dist(p,x^*)+e_i\\
&\geq (1+c_1\eps)(\dist(\proj(p,B),x^*)-\dist(p,B))+e_i\\
&\geq (1+c_1\eps)\big(\dist(\proj(p,B),x^*)-\eps\dist(p,x)\big)+e_i\\
\end{split}
\]
That is,
\begin{equation*}
\label{fii3}
(1+\eps+c_1\eps^2)\dist(p,x)\geq (1+c_1\eps)\dist(\proj(p,B),x^*)+e_i.
\end{equation*}
By the triangle inequality and~\eqref{mme}
\[
\dist(\proj(p,B),x)\geq \dist(p,x)-\dist(p,B)
> (1-\eps)\dist(p,x).
\]
Together with the previous inequality and the assumptions $\eps\leq 1/10$, $c_1=10$, and $c_2=2$, we obtain
\[
\begin{split}
\dist(\proj(p,B),x)
&\geq \frac{1-\eps}{1+\eps+c_1\eps^2}\left((1+c_1\eps)\dist(\proj(p,B),x^*)+e_i\right)\\
&\geq (1+\eps)\dist(\proj(p,B),x^*)+e_i/c_2.
\end{split}
\]
Hence, $f'_p(x)=\dist(\proj(p,B),x)$.
By~\eqref{fii}, we also have $\dist(p,x)=f_p(x)$, which proves~\eqref{i} as
\[
\begin{split}
|f_p(x)-f'_p(x)|
&=|\dist(p,x)-\dist(\proj(p,B),x)|\\
&\leq \dist(p,B)=\eps s_{f_p}(x)<\eps f_p(x),
\end{split}
\]

For every $f=f_p\in F$, let $g_f=g_{f_p}$ be defined as in Line~\ref{def:G_f} of a call to~$\coresetA(F_{\ind \XX(S)},F'_{\ind \XX(S)},s,m,\eps^2)$; see Fig~\ref{fig:algA}. Let $G=\br{g_{f_p}\mid f_p\in F}$,  $S=\br{g_{f_p} \mid p\in \T}$, and $\XX(D)=\ZZ(D)$. By applying Lemma~\ref{klemma}, we have $\dim(G,\XX)=O(kj^2\log(1/\eps)/\eps)$.
%\mr{[Why are we taking both $P$ and $ \proj(P,B)$ in the above? Danny: fixed]}
By its construction, $S$ is a random sample of $t$ i.i.d function from $G$. By Theorem~\ref{functionspacerand}, with probability at least $1-\delta$, $S$ is thus an $\eps^2$-approximation of \mr{$G_{\ind \XX(S)}$}.
Assume that this event indeed occurs, and let $C$ be the output of a call to $\coresetA(F_{\ind \XX(S)},F'_{\ind \XX(S)},s,m,\eps^2)$ using $S$ as an $\eps$-approximation for $G$ in Line~\ref{constructS}.
By Theorem~\ref{the:coresetA} we obtain
\mr{
\[
\begin{split}
|\cost(F,x)-\cost(C,x)|
\leq &\sum_{f\in F\setminus M(x)}\big|f(x)-f'(x)\big|+\eps^2\max_{f\in M(x)}\frac{s_f(x)}{m_f}\sum_{f\in F}m_f.
\end{split}
\]
}
We have
\[
\frac{s_f(x)}{m_f}=\frac{\dist(p,B)}{\eps m_f}\leq \frac{\cost(P,B)}{|P|\eps}.
\]
Combining the last two inequalities and~\eqref{i} yields
\[
\begin{split}
|\cost(F,x)-\cost(C,x)|
&\leq \sum_{f\in F}\eps  f(x)+\eps^2\cdot \frac{\cost(P,B)}{|P|\eps}\sum_{f\in F} \left(\frac{|P|f(B)}{\cost(P,B)}+2\right)\\
&\leq \eps\cost(F,x)+3\eps\cost(P,B).
\end{split}
\]
Hence,
\begin{equation}\label{ccx}
\cost(C,x)\geq (1-\eps)\cost(F, x)-3\eps\cost(P,B).
\end{equation}

We now prove that the right hand side of the last inequality is positive.
By letting
\[\Pbad=\bigcup_{i=1}^k \br{p\in P_i \mid \dist(p,x)> (1+c_1\eps)\dist(p,x^*)+e_i},\]
we obtain
%\mr{[In last equality below there seems to be an additive factor of $c_1\eps$ missing. Danny: right]}
\begin{equation}\label{bbad2}
\begin{split}
\cost(P\setminus \Pbad, x)
&=\sum_{p\in P\setminus \Pbad}\dist(p,x)
\leq (1+c_1\eps)\sum_{p\in P\setminus \Pbad}\dist(p,x^*)+\sum_{i\in [k]}e_i\cdot |P_i|\\
&= (1+c_1\eps +c_2\eps)\cost(P,x^*)+c_2\eps \cost(P,B).
\end{split}
\end{equation}
Using~\eqref{costpt2}, and the assumption $\eps\leq 1/10$ of the lemma, we have
\[
\begin{split}
\cost(P,x)
&>(1+c_2\eps)\cost(P,x^*)+(6+c_2)\eps\cost(P,B)\\
&\geq (1+c_2\eps)\cost(P,x^*)+3(1+2\eps)\eps\cost(P,B)+c_2\eps\cost(P,B).
\end{split}
\]
Combining the last inequality with~\eqref{bbad2} yields
\begin{equation}\label{FFx}
\begin{split}
\cost(F,x)&=\cost(\Pbad,x)
=\cost(P,x)-\cost(P\setminus \Pbad,x)\\
&\geq \cost(P,x)-(1+c_2\eps)\cost(P,x^*)-c_2\eps\cost(P,B)\\
&> 3(1+2\eps)\eps\cost(P,B)
> \frac{3\eps\cost(P,B)}{1-\eps}.	
\end{split}
\end{equation}
By this and~\eqref{ccx},
\[
\cost(C,x) \geq (1-\eps)\cost(F, x)-3\eps\cost(P,B)>0.
\]
By construction of $C$, we have either (i) $f'_p(x)>0$ for some $f_p\in F$, or (ii) $g_{f_p}(x)>0$ for some $g_{f_p}\in S$; see Fig.~\ref{fig:algA}. Let $i\in [k]$ such that $p\in P_i$. In case (i), we have
\[
\dist(\proj(p,B),x)>(1+\eps)\dist(\proj(p,B),x^*)+\eps\cdot\frac{\cost(P_i,x^*)+\cost(P_i,B)}{|P_i|}.
\]
In case (ii), $g_{f_p}(x)=f_p(x)/m_f>0$ for some $p\in \T$. Hence, $f_p(x)>0$, and
\[
\dist(p,x)>(1+c_1\eps)\dist(p,x^*)+e_i
>(1+\eps)\dist(p,x^*)+\eps\cdot\frac{\cost(P_i,x^*)+\cost(P_i,B)}{|P_i|}.
\]
We conclude that the lemma holds for both cases.
\end{proof}

\begin{theorem}\label{weaktheorem}
Let $P$ be a finite set of points in $	\REAL^d$, $0<\eps<1/2$, $B\subseteq \REAL^d$, $j,k\geq 1$ and \[
t\geq \frac{c}{\eps^4} \left(\frac{j^2k\log(1/\eps)}{\eps}+\log\frac{1}{\delta}\right),
\] for sufficiently large constant $c$. Let $(D,\T,w)$ be the output of the algorithm $\dimred(P,B,t,\eps/c)$ \mr{Mike: same remark regarding $z$}.
Let \mbx{$y^*$} be a $k$-set that minimizes $\sum_{p\in D}w(p,x)\dist(p,x)$ over every $x\in \ZZ(D,j,k)$ up to a multiplicative factor of $(1+\eps/c )$.
\mr{Mike: I am not sure I understand the last sentence.}
Then, with probability at least $1-\delta$,
\mbx{
\[
\cost(P,y^*)\leq (1+\eps)\min_{x^*\in X(j,k)}\cost(P,x^*)+\eps \cost(P,B).
\]
}
\end{theorem}
\begin{proof}
We prove the case where $x$ is a set of $k$ points (that is, $j=1$). The case $j\geq 2$ is similar, using the observations from~\cite{DesVar07,ShyVar07}.
\mrx{Need to elaborate on $j \geq 2$.}
Let $x^*=(x_1,\cdots, x_k)$ be a $k$-tuple of points that minimizes $\cost(P,x)$ over every $x\in X(1,k)$.
%Put $i$, $1\leq i\leq k$, and let $P_i$ denote the set of points whose closest center in $x^*$ is $x_i$.
By Lemma~\ref{mylem} (using the notation introduced in its statement and proof), we infer that, with probability at least $1-\delta$,
for every $k$-tuple $x\in \ZZ(D,1,k)$ that satisfies
\begin{equation}\label{holds}
\cost(P,x)>(1+c_2\eps)\cost(P,x^*)+(6+c_2)\eps\cost(P,B),
\end{equation}
there is $p\in D$ such that $\dist(p,x)>\mbx{(1+\eps)\dist(p,x^*)}+\eps\cost(P_i,x^*)/|P_i|$.
Assume that~\eqref{holds} holds, which happens with probability at least $1-\delta$.

\mbx{Our proof contains two conceptual steps.
In the first step, we use Lemma~\ref{mylem} to iteratively prove the existence of a point $x' \in \ZZ(D,1,k)$ for which
\[
\cost(P,x') \leq  (1+c_2\eps)\cost(P,x^*)+(6+c_2)\eps\cost(P,B).
\]
Combining the properties of $x'$, with the fact that $D$ is a coreset (via Theorem~\ref{mainsub}), will consist of the second part of our proof.}
%\mr{danny: D is not a strong coreset.}

\mbx{Our starting point for the first step of our proof, is the set of points $y^0$ defined as follows.}
For every $1\leq i\leq k$, let $y^0_i$ denote the closest point to $x_i$ in $\proj(P,B)$. That is, for every $p_b\in \proj(P,B)$,
\begin{equation}\label{yz}
\norm{y^0_i-x_i}\leq \norm{p_b-x_i}
\end{equation}
\mbx{Notice that $y^0 \in \ZZ(D,1,k)$.}
\mbx{If
\[
\cost(P,y^0) \leq  (1+c_2\eps)\cost(P,x^*)+(6+c_2)\eps \cost(P,B)
\]
then we are done, and have completed the first step of our proof (we set $x'=y^0$).
%Otherwise, it holds that
%\begin{equation}\label{cpv}
%\cost(P,y^0)>(1+c_2\eps)\cost(P,x^*)+(6+c_2)\eps \cost(P,B).
%\end{equation}

Otherwise, we now present a procedure $\textsc{Improve}$, that for any integer $v \geq 0$, receives $y^v=(y^v_1,\cdots,y^v_k)\in X(1,k)$ such that
\begin{equation}\label{witness2}
\cost(P,y^v)> (1+c_2\eps)\cost(P,x^*)+(6+c_2)\eps\cost(P,B),
\end{equation}
and outputs $y^{v+1} \in \ZZ(D,1,k)$.
We show that iteratively applying $\textsc{Improve}$ will result in the desired $x'$.
}
%By substituting $P'=P_i$, $x^*=x_i$ and $x=y^v_i$ in Lemma~\ref{mylem}, we infer that there is $p\in T\cup \proj(P,B)$ such that $\dist(p,y^v_i)>(1+\eps)\dist(p,x_i)$.

%We now define the output $y^{v+1}$ of the procedure $\textsc{Improve}(y^{v})$.
%By.. there is a ``witness" $i\in [k]$ such that
%\begin{equation}\label{witness}
%\\cost(P_i,y_i^v)> (1+\eps)\cost(P_i,x_i)+c_1\eps\cost(P,B)/k,
%\end{equation}
%There is a ``witness" $i\in [k]$ such that
%\begin{equation}\label{witness}
%\dist(p,y^v)\geq (1+\eps)\dist(p,x_i)+\eps\norm{y_i^0-x_i}.
%\end{equation}

By substituting $x=y^v$ in Lemma~\ref{mylem}, we infer that there is a ``witness" $i\in [k]$ and $p\in (\T\cap P_i)\cup \proj(P_i,B)$ such that
\begin{equation}\label{witness}
\dist(p,y_i^v)>(1+\eps)\dist(p,x_i)+\eps\cdot\frac{\cost(P_i,x^*)+\cost(P_i,B)}{|P_i|}.
\end{equation}
Using the last inequality, it is not hard to prove (see, for example,~\cite[Lemma 2.2 ]{ShyVar07}), that there is a point $y^{v+1}_{i}\in \Span{p\cup y^v_i}$ such that
\begin{equation}\label{normv}
\norm{y^{v+1}_i-x_i}\leq (1-\eps/2)\norm{y^{v}_i-x_i}.
\end{equation}
The procedure $\textsc{Improve}$ returns $y^{v+1}$ which is the $k$-tuple $y^v$ after replacing $y^v_i$ with $y^{v+1}_{i}$.
\mbx{Notice that $y^{v+1} \in \ZZ(D,1,k)$.}

Suppose that we call to the procedure $\textsc{Improve}(y^{v})$ for $v=0,1,\ldots$ until~\eqref{witness2} does not hold.
Fix $i\in [k]$ and $m=10\log(1/\eps)/\eps$. We now prove that \mbx{in at most $m$ calls of $\textsc{Improve}$ the index $i$ was a ``witness" that govern the construction of $y^{v+1}$}.
Indeed, by contradiction assume that~\eqref{witness} holds for $i\in[k]$ for the $v$th time, $v>m$. Applying~\eqref{normv} $v$ times yields
\[
\label{cpi}
 \norm{y^{v}_i-x_i}\leq (1-\eps/2)^{m}\norm{y^{0}_i-x_i}<\eps\norm{y^{0}_i-x_i}.
\]
For every $p\in P_i$, let $p_b=\proj(p,B)$ denote its closest center in $B$.
By~\eqref{yz}, $\norm{y^0_i-x_i}\leq \norm{p_b-x_i}$. By the triangle inequality,
$\norm{p_b-x_i} \leq \dist(p_b,p)+\dist(p,x_i)$. Combining the last two inequalities yields $\norm{y^0_i-x_i}\leq\dist(p_b,p)+\dist(p,x_i)$.
Hence,
\[
|P_i|\cdot\norm{y^0_i-x_i}
\leq \sum_{p\in P_i}\big(\dist(p_b,p)+\dist(p,x_i)\big)
= \cost(P_i,B)+\cost(P_i,x^*).
\]
For every $p\in P_i$, we thus have
\[
\begin{split}
\dist(p,y_i^v)
&\leq \dist(p,x_i)+\norm{y^v_i-x_i}\\
&\leq \dist(p,x_i)+\eps\norm{y^{0}_i-x_i}\\
&\leq \dist(p,x_i)+\eps\cdot\frac{\cost(P_i,B)+\cost(P_i,x^*)}{|P_i|}.
\end{split}
\]
which contradicts the assumption that~\eqref{witness} holds.

Let $x'=y^v$ be the output of the last call to~$\textsc{Improve}$. Hence,~\eqref{holds} does not hold for \mbx{$x'$}, i.e,
\begin{equation}\label{almostthere2}
\cost(P,x')\leq(1+c_2\eps)\cost(P,x^*)+(6+c_2)\eps \cost(P,B).
\end{equation}
By construction, every point in \mbx{$x'$} is spanned by at most $m$ points from $D$. That is, $x\in \ZZ(D,1,k)$.
\mbx{This concludes the first part of our proof.}

By Theorem~\ref{mainsub}, with probability at least $1-\delta$ we have
\begin{equation}
\label{eq18:delta}
\begin{split}
&\forall x\in \ZZ(D,j,k):\\
& \left|  \cost(P,x)-\sum_{p\in D}w(p,x)\dist(p,x) \right|\leq \eps\cost(P,B)+\eps \cost(P,x).
\end{split}
\end{equation}
Using this inequality, we now claim that
\[
\cost(P,y^*)\leq (1+O(\eps))\cost(P,x')+O(\eps)\cost(P,B),
\]
where $y^*$ minimizes $\sum_{p\in D}w(p,x)\dist(p,x)$ over $\ZZ(D,j,k)$ up to a multiplicative factor of $(1+\eps)$.
This follows as \eqref{eq18:delta} implies that $\cost(P,y^*)\leq \sum_{p\in D}w(p,y^*)\dist(p,y^*) + \eps\cost(P,B)+\eps \cost(P,y^*)$.
Since $x'\in \ZZ(D,j,k)$ we have $\sum_{p\in D}w(p,y^*)\dist(p,y^*) \leq (1+\eps)\sum_{p\in D}w(p,x')\dist(p,x')$.
Moreover, \eqref{eq18:delta} also implies
$\sum_{p\in D}w(p,x')\dist(p,x') \leq \cost(P,x') + \eps\cost(P,B)+\eps \cost(P,x')$.
Combining all these inequalities yields
\[
\cost(P,y^*)\leq (1+O(\eps))\cost(P,x')+O(\eps)\cost(P,B).
\]

Combining this with~\eqref{almostthere2} yields
\[
\begin{split}
\cost(P,y^*)
&\leq (1+O(\eps))\cost(P,x^*)+O(\eps) \cost(P,B),
\end{split}
\]
which proves the theorem for a call to $\dimred(P,B,t,\eps/c)$ and a sufficiently large $c$.
\end{proof}

%\begin{lemma}\label{weak}
%Let $F$ be a set of functions from a set $X$ to $[0,\infty)$, and let $x'\in X$.
%Let $\Delta>0$, $U:X\rightarrow (-\infty,\infty)$, and $X_U\subseteq X$ such that $x'\in X_U$. Suppose that
%\begin{equation}\label{XU}
%\forall x\in X_U:|\cost(F,x)-U(x)|\leq \Delta.
%\end{equation}
%
%Let $y^*$ be the item that minimizes $|U(y)|$ over every $y\in X_U$ up to a multiplicative factor of $\alpha$.
%Then
%\[
%\cost(F,y^*)\leq \alpha\cost(F,x')+2\alpha\Delta.
%\]
%\end{lemma}
%\begin{proof}
%By substituting $x=y^*$ in~\eqref{XU}, we obtain $\cost(F,y^*)\leq |U(y^*)|+\Delta$.
%Since $x'\in X_U$, we have $|U(y^*)|\leq \alpha|U(x')|$. Using~\eqref{XU} again yields
%$|U(x')|\leq \cost(F,x')+\Delta$. Combining all these inequalities yields
%\[
%\cost(F,y^*)\leq |U(y^*)|+\Delta\leq \alpha|U(x')|+\alpha\Delta\leq \alpha\cost(F,x')+2\alpha\Delta.
%\]
%\end{proof}

\subsection{Weak coreset and PTAS for $k$-median}

\mr{Mike: Up to here}

\mrx{Add introductory text.}

\mrx{There may be a problem with $1/\eps^2$ vs. $1/\eps^4$ below.}

\begin{theorem}[weak coresets for $k$-median in $\REAL^d$]\label{weakmedian}
Let $P$ be a set of $n$ points in $\REAL^d$. Let $k\geq 1$ be an integer, $0<\eps,\delta<1/2$, and
\[
\t=\frac{c}{\eps^4}\cdot \left(\frac{k\log(1/\eps)}{\eps}+\log(1/\delta)\right),
\]
where $c$ is a sufficiently large constant.
Then, a set $\DD\subseteq P$ of size $|\DD|=t$, with a weight function $w:D\rightarrow [0,\infty)$, can be computed
 such that, with probability at least $1-\delta$,
\begin{equation}\label{costDY}
\sum_{p\in P}\dist(p,y)\leq (1+\eps )\min_{x\in (\REAL^d)^k}\sum_{p\in P}\dist(p,x),
\end{equation}
where $y$ is any center that minimizes $\cost(D,y^*)$ over \mbx{$y\in \ZZ(D,1,k)=\ZZ(D)$} up to a multiplicative factor of $(1+\eps)$.
The construction time of $\DD$ is $O(ndk)+O(1)\cdot\log^2(1/\delta)\log^2 n+O(k^2)+O(\t\log n)$.
\end{theorem}
\begin{proof}
By Theorem~\ref{constfactor}, a set $B\subseteq P$ of $k$ points can be computed in $O(ndk)+(k+\log(2/\delta)\log n)^{2}$ time such that, with probability at least $1-\delta$,
\begin{equation}\label{PBMedian3}
\cost(P,B) \leq O(1)\min_{x\in P^k}\cost(P,x).
\end{equation}
Assume that~\eqref{PBMedian3} indeed holds.
Let $(D,\T,w)$ be the output of a call to the algorithm~{\rkmedian$(P,B,t,\eps)$}

Consider the set of functions $\KK(P)$; see Definition~\ref{GGdef3}.
For every $S\subseteq \KK(P)$, let $\XX(S)=\ZZ(S,1,k)$.
Using Lemma~\ref{klemma}, we have $\dim(\KK(P),\XX)=O(kj^2\log(1/\eps)/\eps)$.
Similarly to the proof of Theorem~\ref{metrickmed}, using the above definition of $\dim(\KK(P),\XX)$, we have with probability at least $1-\delta$,
\[
\forall x\in \ZZ(S,1,k): \left|\sum_{p\in P}\dist(p,x)-\sum_{p\in D}w(p)\dist(p,x)\right|\leq \eps \sum_{p\in P}\dist(p,x).
\]
The running time is $O(ndk)+O(1)\cdot\log^2(1/\delta)\log^2 n+O(k^2)+O(\t\log n)$.
\mrx{Why is there a $d$ in the running time above ... it does not appear in Theorem~\ref{metrickmed} ... maybe we need to add a parameter $\tt{time}$ that holds the complexity of finding the distance between two points in a given metric space. Danny: I used $d$}
Let $y\in \ZZ(S,1,k)$ be a tuple of $k$ points that satisfies
\begin{equation*}
\cost(D,y)\leq (1+O(\eps))\min_{y^*\in \ZZ(S,1,k)}\sum_{p\in D}w(p)\dist(p,x).
\end{equation*}

\mr{What are we quoting below - currently it is the theorem we are proving ... notice that in all lemmas in this section we use $1/\eps^4$ and not $1/\eps^2$. Danny: I think that now it's ok.}

By Theorem~\ref{weaktheorem},
\[
\cost(P,y)\leq (1+O(\eps ))\min_{x^*\in X(1,k)}\cost(P,x^*),
\]
as desired, for choosing a sufficiently large $c$.
\end{proof}

\begin{theorem}[PTAS for $k$-median in $\REAL^d$]
Let $P$ be a set of $n$ points in $\REAL^d$. Let $k\geq 1$ be an integer, $0<\eps,\delta<1/2$, and
\[
\mbx{\t=\frac{c}{\eps^4}\cdot \left(\frac{k\log(1/\eps)}{\eps}+\log(1/\delta)\right),}
\]
where $c$ is a sufficiently large constant.
Then, a tuple $y$ of $k$ points can be computed in
\[
O(ndk)+O(1)\cdot\log^2(1/\delta)\log^2 n+O(k^2)+O(\t\log n)+d\cdot \t+\mbx{d\log(1/\delta)\cdot 2^{\poly(k,1/\eps)}}
\]
 time such that, with probability at least $1-\delta$,
\[
\sum_{p\in P}\dist(p,y)\leq (1+\eps)\min_{x^*\in  (\REAL^d)^k}\sum_{p\in P}\dist(p,x^*)
\]
\end{theorem}

\begin{proof}
Using the result of Theorem~\ref{weakmedian}, we only need to compute $y$ that satisfies~\eqref{costDY}.
To this end, we can simply compute $\cost(D,x)$ over every $x\in \ZZ(S,1,k)$. This will take time
$|\ZZ(S,1,k)|\cdot d|S|=dt^{O(k^2\log(1/\eps)/\eps)}$.
A little faster option is to compute $y'$ such that
\[
\cost(D,y')\leq (1+\eps)\min_{y^*\in X(1,k)}\sum_{p\in D}w(p)\dist(p,x).
\]
This takes time \[O(d\cdot \t\cdot 2^{\poly(k,1/\eps)})=d\log(1/\delta)\cdot 2^{\poly(k,1/\eps)};\] see, for example,~\cite{AYS10}.
The overall running time is therefore $O(ndk)+O(1)\cdot\log^2(1/\delta)\log^2 n+O(k^2)+O(\t\log n)+d\cdot \t + \mbx{d\log(1/\delta)\cdot 2^{\poly(k,1/\eps)}}$.
\end{proof}

\mrx{Need to add explicit running times for approximation when $j \ne 1$.}

\section{Subspace Approximation}
%For a pair of positive integers $n$ and $d$, we denote the set of all $n\times d$ matrices by $\REAL^{n\times d}$.
%We denote $\norm{A}_F$ as the sum of squared entries of the matrix $A$.
\subsection{Sum of squared distances ($\ell_2$ error)}
\mrx{Did not read.}
Let $P$ be a set of $n$ points in $\REAL^d$. Let $UDV^T$ denote the svd of the matrix whose rows are the points of $P$. Since the columns of $U$ are orthogonal, the rows of the matrix $DV^T$ corresponds to $d$ points in $\REAL^d$ such that the sum of squared distances from the points of $D$ to any subspace $x\in X(j,1)$ equals to the sum of squared distances from the points of $P$ to $x$; see details in the proof of Theorem~\eqref{jj}. The construction of $D$ takes $O(nd^2)$ time and $O(nd)$ space. In the next two theorems we prove that an approximation to the optimal subspace $x\in X(j,1)$ of $P$ can be computed faster and in the streaming model.
% next: we can compute optimal $j$-subspace in space $O(j^2\log n)$. streaming: Let $P$ be a set. Let $Q\subseteq P$. WE compute a coreset that approximates the optimal $j$-subspace of every subset of $P$.
\begin{theorem}\label{jj}
Let $P$ be a set of $n$ points in $\REAL^d$. Let $j\geq 1$ be an integer and $\eps,\delta \geq 0$. Let $c\geq 1$ be a sufficiently large universal constant, and $n\geq1$ be sufficiently large. Then, a $j$-subspace $x$ that minimizes $\sum_{p\in P}\dist^2(p,x)$ over every $x\in X(j,1)$ up to a multiplicative factor of $(1+\eps$) can be computed, with probability at least $1-\delta$,
in time
\[
O(nd)\cdot \min\br{j, \log (n)}+O(nd)\log\left(\frac{1}{\delta}\right).
\]
\end{theorem}
\begin{proof}
Let $T$ be a random sample of
\[
r=c\left(j+\log\frac{1}{\delta}\right).
\]
points from $P$. By applying Lemma~\ref{jk4} with $k=1$, $\eps=1/10$, and $\gamma=3/4$, the span of $T$ contains, with probability at least $1-\delta$, a $(\gamma,\eps,1+\eps,\infty)$-median for $F(P,j)$. If $P=T$ then the span of $T$ trivially contains a $(1,0,1)$-median for $F(P,j)$. Let $y\in X(r,1)$ an $r$-dimensional subspace, and let $A$ be an $d\times r$ matrix whose columns are mutually orthogonal unit vectors that span $y$. The squared distance from a point $p\in P$ to $y$ is then
\[
\norm{p-p^TAA^T}^2=\norm{p}^2-\norm{p^TAA^T}^2=\norm{p}^2-\norm{p^TA}^2.
\]
The construction of $A$ from the set $T$ that spans $y$ takes $O(dr^2)$ time via SVD~\cite{golub1996matrix}.

Using the observations from the previous paragraph we apply Theorem~\ref{73} with $\beta=1$, $\eps=1/10$, and $\alpha=1$ to obtain a set $Z=\br{Z_1,Z_2,\cdots, }$, $|Z|\leq \log_2 n$ of $O(r)$-dimensional subspaces and a partition $(P_1,\cdots, P_{|Z|})$ of $P$ such that, with probability at least $1-\delta/10$,
\begin{equation}
\label{ss1}
\sum_{i=1}^{|Z|}\cost(P_i,Z_i)\leq 2\min_{x\in X(j,1)} \cost(P,x).
\end{equation}
This takes time
\begin{equation*}
\bicriteriatime' = O(ndr)+O(dr^2\log^2n).
\end{equation*}
Since the last term is the bottleneck of our construction, we now suggest a construction which is faster for large values of $r$.

Let $V$ denote a $d\times (d-r)$ matrix whose columns are mutually orthogonal unit vectors that span the $(d-j)$-subspace that is orthogonal to $y$. Hence, the distance from $p\in P$ to $y$ is $\norm{p^TV}^2$. Let $B$ be a $(d-r)\times (c\log (n/\delta))$ matrix whose entries are Gaussian unit vectors. Using the Johnson-Lindenstrauss lemma~\cite{GD02}, we have, with probability at least $1-\delta$ \danny{(??)}
\begin{equation}\label{jleq}
\frac{\norm{p^TVB}^2}{2}\leq \norm{p^TV}^2\leq 2\norm{p^TVB}^2.
\end{equation}
Let $t=O(n)$, and let $Q$ be the points of $P$ that are closest to $y$, i.e, the points $p\in P$ with the smallest values $\norm{p^TB}$. Let $\tilde{Q}$ be the $t$ points $p\in P$ with the smallest values $\norm{p^TVB}$. By~\eqref{jleq},
\begin{equation}\label{44}
\begin{split}
\sum_{p\in \tilde{Q}}\dist^2(p,y)
=\sum_{p\in \tilde{Q}}\norm{p^TV}^2
&\leq 2\sum_{p\in \tilde{Q}}\norm{p^TVB}^2\\
&\leq 2\sum_{p\in Q}\norm{p^TVB}^2
\leq 4\sum_{p\in Q}\norm{p^TV}^2
= 4\sum_{p\in Q}\dist^2(p,y).
\end{split}
\end{equation}
Using this construction of $\tilde{Q}$ in order to compute an approximation to $G_i$ in Line 4 of the algorithm \bicriteria, would yield a bicriteria approximation with $\alpha=4$; see Fig.~\ref{fig:alg_bi2}. This is because, using~\eqref{44}, the term $\cost(G_i,Y_i)$ in~\eqref{costgi} is increased by a factor of at most $\alpha=4$ when we replace the desired set $G_i=Q$ with $\tilde{Q}$.

The matrix $V$ can be computed from $T$ in $O(d^2r)$ time. The matrix $B$ can be computed in $O(d\log (n/\delta))$ time. Multiplying $VB$ takes $O(d^2\log (n/\delta))$ time, computing $\norm{p^TVB}$ using $VB$ takes $O(d\log (n/\delta))$, and computing $\tilde{Q}$ from $VB$ takes $O(nd\log (n/\delta))$ time, using order statistics. The overall construction time of $Z$ is then
\begin{equation}\label{timeb}
\bicriteriatime = O(nd\log (n/\delta))+O(d^2\log^2n(r+\log (n/\delta))).
\end{equation}

For every $p\in P_i$ let $p'=\proj(p,Z_i)$ denote its projection on $Z_i$.
Let $A$ denote a $d\times (d-j)$ matrix whose columns are mutually orthogonal unit vectors that span the orthogonal subspace to $x\in X(j,1)$. Hence, $\norm{p^TA}=\dist(p,x)$, where we consider $p\in\REAL^d$ as a column vector.
We have
\begin{equation}\label{sumeq}
\begin{split}
\dist^2(p,x)
&=\norm{p^TA}^2
=\norm{(p-p'+p')^TA}^2\\
&=\norm{(p-p')^TA+p'^TA}^2\\
&=\norm{(p-p')^TA}^2+\norm{p'^TA}^2+A^Tp'(p-p')^TA\\
&=\norm{(p-p')^TA}^2+\norm{p'^TA}^2\\
&=\dist^2((p-p'),x)+\dist^2(p',x).
\end{split}
\end{equation}
For every $p\in P$ and $x\in X(j,1)$, we define $f_p(x)=\dist^2(p-p',x)$ and $F=\br{f_p\mid p\in P}$. For every $f_p\in F$, let $s_f=f$, $b=(1+\eps)/n$ and
\begin{equation}\label{mmf}
m(f_p):=\left\lceil \frac{n\norm{p-p'}^2}{\sum_{i=1}^{|Z|}\cost(P_i,Z_i)}\right\rceil
 \geq \frac{n\dist^2(p-p',x)}{(1+\eps)\cost(P,x)}
= \frac{s_f(x)}{b\cost(F,x)},
\end{equation}
where the first inequality is by~\eqref{ss1} and the fact that any subspace contains the origin.

Pick a random sample $\T$ of
\[
s=\frac{c}{\eps^2}\left(\frac{j\log(1/\eps)}{\eps}+\log\frac{1}{\delta}\right),
\]
i.i.d. points from $P$, where the probability that a point in $\T$ equals $p\in P$ is $m(f_p)/\sum_{p\in P}m(f_p)$. Let $\XX(S)=\ZZ(S,j)$ denote the set of all the possible $j$-flats that are spanned by at most $10j\log(1/\eps)/\eps$ points from $\T$, as in Definition~\ref{ZZ}.
Let $x^*_P$ denote the $j$-subspace that minimizes $\cost(P,x)$ over every $x\in X(j,1)$. Let $\XX^{+}(S)=\XX(S)\cup \br{x^*_P}$.

For every $f=f_p\in F$, let $g_f=g_{f_p}$ be defined as in Line~\ref{def:G_f} of a call to~$\coresetA(F_{\ind \XX^+(S)}, F_{\ind \XX^{+}(S)}, s, m,\eps)$; see Fig~\ref{fig:algA}. Let $G=\br{g_{f_p}\mid f_p\in F}$, and $S=\br{g_{f_p} \mid p\in \T}$.
Note that $(G,\XX^+)$ is a generalized range space; see Definition~\ref{dimfunctionsHigh2}.
By Theorem~\ref{1e}(i), we have $\dim(G,\XX)=O(j\log(1/\eps)/\eps)$.
The number of ranges in $\XX^+(S)$ is larger by at most $|S|$ than the number of ranges in $\XX(S)$. Hence, $\dim(G,\XX^+)\leq \dim(G,\XX)+1$. See the proof of a similar argument in Lemma~\ref{functions}.

By its construction, $S$ is a random sample of $c\eps^{-2}(\dim(G,\XX^+)+\log(1/\delta))$ i.i.d functions from $G$. By Theorem~\ref{functionspacerand}, with probability at least $1-\delta/10$, $S$ is thus an $\eps$-approximation of $G_{\ind \XX^+(S)}$.
Assume that this event indeed occurs, and let $C$ be the output of such a call to $\coresetA(F_{\ind \XX^+(S)}, F_{\ind \XX^+(S)}, s, m, \eps)$ using $S$ as an $\eps$-approximation for $(G_{\ind \XX^+(S)})$ in Line~\ref{constructS}.

Put $x\in X^+(S)$. By Corollary~\ref{cor:mean1} and~\eqref{mmf},
\[
\begin{split}
|\cost(F,x)-\cost(C,x)|&\leq \eps b\cost(F,x)\left(1+2\sum_{f\in F}m_f\right)\\
&\leq
\eps b\cost(F,x)\left(1+2n+2\sum_{f\in F}\frac{n\norm{p-p'}^2}{\sum_{i=1}^|Z|\cost(P_i,Z_i)}\right)\\
&=\frac{(1+\eps)\eps \cost(F,x)(1+4n)}{n}
\leq10\eps\cost(F,x).
\end{split}
\]
Let
\[
D':=\br{(p-p')\sqrt{\frac{|G|}{|\T|\cdot m(f_p)}}\mid p\in \T}.
 \]
By the previous inequality and the construction of $C$, we have
\begin{equation}\label{pps}
\begin{split}
\left|\sum_{p\in P}\dist^2(p-p',x)-\sum_{p\in D'}\dist^2(p,x)\right|
&=\left|\sum_{p\in P}\dist^2(p-p',x)-\sum_{p\in \T}\frac{|G|}{|S|}\cdot\frac{\dist^2(p-p',x)}{m(f_p)}\right|\\
&=\left|\cost(F,x)-\cost(C,x)\right|\\
&\leq  10\eps\cost(F,x)=10\eps\sum_{p\in P}\dist^2(p-p',x).
\end{split}
\end{equation}

For every $i$, $1\leq i\leq \log_2n$, let $P'_i$ denote an $n_i\times d$ matrix whose set of rows is $\br{p'\mid p\in P_i}$. The matrix $P'_i$ can be constructed from $P_i$ and $Z_i$ in $O(n_idr)$ time. Since $P'_i$ has rank $O(r)$, there is a decomposition $P'_i=Q_iR_i$ such that $Q_i$ is an $n_i\times O(r)$ matrix whose columns are mutually orthogonal unit vectors, and $R_i$ is an $O(r)\times d$ matrix. $Q_i$ and $R_i$ can be computed using the QR or SVD decomposition of $P'_i$ in $n_i\cdot O(r^2)$ time. Hence, the overall time over all $1\leq i\leq |Z|$ is $O(ndr+nr^2)=O(ndr)$.

By denoting $\norm{\cdot}_F$ as the Frobenius norm, we obtain
\begin{equation*}
\sum_{p\in P'_i}\dist^2(p,x)=\norm{P'_iA}_F=\norm{Q_iR_iA}_F=\norm{R_iA}_F.
\end{equation*}
Let $R$ be an $n\times O(r)$ matrix whose rows are the union of rows in the matrices $R_1,\ldots,R_{|Z|}$. Hence,
\begin{equation}
\label{eq:ppp}
\sum_{p\in P'}\dist^2(p,x)=\sum_{i=1}^{|Z|}\norm{R_iA}_F=\norm{RA}_F.
\end{equation}

Let $D_1$ be the union of $D'$ with the set of points which consists of the $O(r)$ rows of $R$. The size of $D_1$ is
 \begin{equation}\label{sized1}
|D_1|=O(|\T|+r|Z|).
\end{equation}
Plugging~\eqref{pps} and~\eqref{eq:ppp} in~\eqref{sumeq} yields that for every $x\in X^+(S$) we have
\begin{equation}\label{costd1}
\begin{split}
&\left|\sum_{p\in P}\dist^2(p,x)-\sum_{p\in D_1}\dist^2(p,x)\right|\\
&=\left|\sum_{p\in P}\dist^2(p-p',x)+\sum_{p\in P'}\dist^2(p,x)
-\left(\sum_{p\in D'}\dist^2(p,x)+\norm{RA}_F\right)\right|\\
&\leq 10\eps\sum_{p\in P}\dist^2(p-p',x)\\
&\leq 10\eps\sum_{p\in P}\dist^2(p-p',x)+10\eps\sum_{p\in P'}\dist^2(p,x)\\
&=10\eps\sum_{p\in P}\dist^2(p,x).
\end{split}
\end{equation}
The constructing time of $D_1$ is dominated by~\eqref{timeb}.

Next, we construct from $D_1$ a smaller coreset $D$ of size only $O(s)$ as follows.
We compute $D_1$ as described above using $\eps=1/10$. We then compute the optimal $j$-subspace $Z_1$ of $D_1$, i.e, $Z_1$ that minimizes $\sum_{p\in D_1}\dist^2(p,y)$ over every $y\in X(j,1)$. This takes time $O(d|D_1|^2)$ time using SVD. By~\eqref{costd1}, $Z_1$ is a $(1,0,O(1))$-median (i.e, constant factor approximation) for $F(P,j)$. We now construct $D$ similarly to the way that $D_1$ was constructed, but using $Z_1$ instead of $Z$ in the beginning of the construction. Replacing $r$ by $j$, and $|Z|$ by $1$ in~\eqref{sized1} yields a set $D$ of size $O(s+j)=O(s)$.
Once we have a small coreset $D$ for $P$, we can use it to compute an approximation to the optimal solution as follows. Compute the optimal $j$-subspace $x^*_D$ of $D$ using SVD in $O(ds\min\br{s, d})$ time. Applying Lemma~\ref{shyvar}(ii) with $x^*_D$ yields a $j$-subspace $\tilde{x}_D\in X^{+}(S)$ such that
\begin{equation}\label{costd}
 \cost(D,\tilde{x}_D)\leq (1+\eps)\cost(D,x^*_D).
\end{equation}

All together we have,
\begin{eqnarray}
\cost(P,\tilde{x}_D)
\label{aa1}&\leq (1+10\eps)\cost(D,\tilde{x}_D)\\
\label{aa2}&\leq (1+10\eps)(1+\eps)\cost(D, x^*_D)\\
\label{aa3}&\leq (1+10\eps)(1+\eps)\cost(D, x^*_P)\\
\label{aa5}&\leq (1+10\eps)^2(1+\eps)\cost(P,x^*_P) \\
\nonumber &\leq (1+c\eps)\cost(P,x^*_P).
\end{eqnarray}
where~\eqref{aa1} and~\eqref{aa5} holds by~\eqref{costd1}, inequality~\eqref{aa2} is by~\eqref{costd}, and inequality~\eqref{aa3} is by the definition of $x^*_D$.
The overall running time is
\[
O(\min\br{\bicriteriatime,\bicriteriatime'} +d|D_1|^2+d|D|\min\br{|D|, d})
=O(a+s^2+ds\min\br{s, d}),
\]
where
\[
a=d\cdot \min\br{r(n+r\log^2n), n\log (n/\delta)+dr\log^2n}.
\]
\end{proof}

\end{document}